\documentclass[preprint,9pt]{elsarticle}

\usepackage{longtable}
\usepackage{pdflscape}
\usepackage{float}
\usepackage{multirow}
\usepackage{amssymb}
\usepackage{xcolor}
\usepackage{tabularx}
\usepackage{makecell}
\usepackage{geometry}
\usepackage{graphicx}
\usepackage{subcaption}
\usepackage{hyperref}
\usepackage{amsmath}
\usepackage{geometry}
\usepackage{booktabs} 
\usepackage{array} 
\usepackage{tikz}
\usepackage{hyperref}
\usepackage{adjustbox}
\usepackage{ragged2e}
\usepackage{amsthm}

\begin{document}

\begin{frontmatter}

\title{Self-Supervised Learning for Graph-Structured Data in Healthcare Applications: A Comprehensive Review}

\author[inst1,inst2]{Safa Ben Atitallah}

\affiliation[inst1]{organization={Robotics and Internet of Things Laboratory, Prince Sultan University},
            city={Riyadh },
            postcode={12435}, 
            country={Saudi Arabia}}
\affiliation[inst2]{organization={RIADI Laboratory, National School of Computer Science, University of Manouba},
            city={Manouba },
            postcode={2010}, 
            country={Tunisia}}

\author[inst2]{Chaima Ben Rabah}

\author[inst1,inst2]{Maha Driss}

\author[inst1,inst2]{Wadii Boulila}

\author[inst1]{Anis Koubaa}

\begin{abstract}
The abundance of complex and interconnected healthcare data offers numerous opportunities to improve prediction, diagnosis, and treatment. Graph-structured data, which includes entities and their relationships, is well-suited for capturing complex connections. Effectively utilizing this data often requires strong and efficient learning algorithms, especially when dealing with limited labeled data. It is increasingly important for downstream tasks in various domains to utilize self-supervised learning (SSL) as a paradigm for learning and optimizing effective representations from unlabeled data. In this paper, we thoroughly review SSL approaches specifically designed for graph-structured data in healthcare applications. We explore the challenges and opportunities associated with healthcare data and assess the effectiveness of SSL techniques in real-world healthcare applications. Our discussion encompasses various healthcare settings, such as disease prediction, medical image analysis, and drug discovery. We critically evaluate the performance of different SSL methods across these tasks, highlighting their strengths, limitations, and potential future research directions. Ultimately, this review aims to be a valuable resource for both researchers and practitioners looking to utilize SSL for graph-structured data in healthcare, paving the way for improved outcomes and insights in this critical field. 
To the best of our knowledge, this work represents the first comprehensive review of the literature on SSL applied to graph data in healthcare.
\end{abstract}


\end{frontmatter}


\section{Introduction}

Integrating Artificial Intelligence (AI) into healthcare represents a pivotal evolution in developing smart healthcare services, fueled by the rapid rise of the Internet of Things (IoT) and big data analytic \cite{aung2021promise,atitallah2024enhancing}. 
These technologies have attracted enormous worldwide attention. They promote the foundation of smart services by using advanced data collection, processing, communication, networking, and computing technologies. This development improves existing healthcare systems and paves the path for innovative, more efficient, and robust healthcare solutions.

Building on this technological foundation, researchers increasingly focus on the untapped potential of graph-structured data within the healthcare sector \cite{calazans2024machine}. Graphs represent various interconnected data types prevalent in healthcare, such as genes, molecules, neurons, and patient records. By applying Deep Learning (DL) techniques to these complex datasets, they can train predictive models capable of uncovering groundbreaking insights that conventional data structures may fail to reveal. This advancement in utilizing graph-based analytics underscores a significant trend where AI not only enhances current systems but also innovates new approaches to solve complex medical challenges.

\begin{figure}[h]
    \centering
    \includegraphics[scale=0.8]{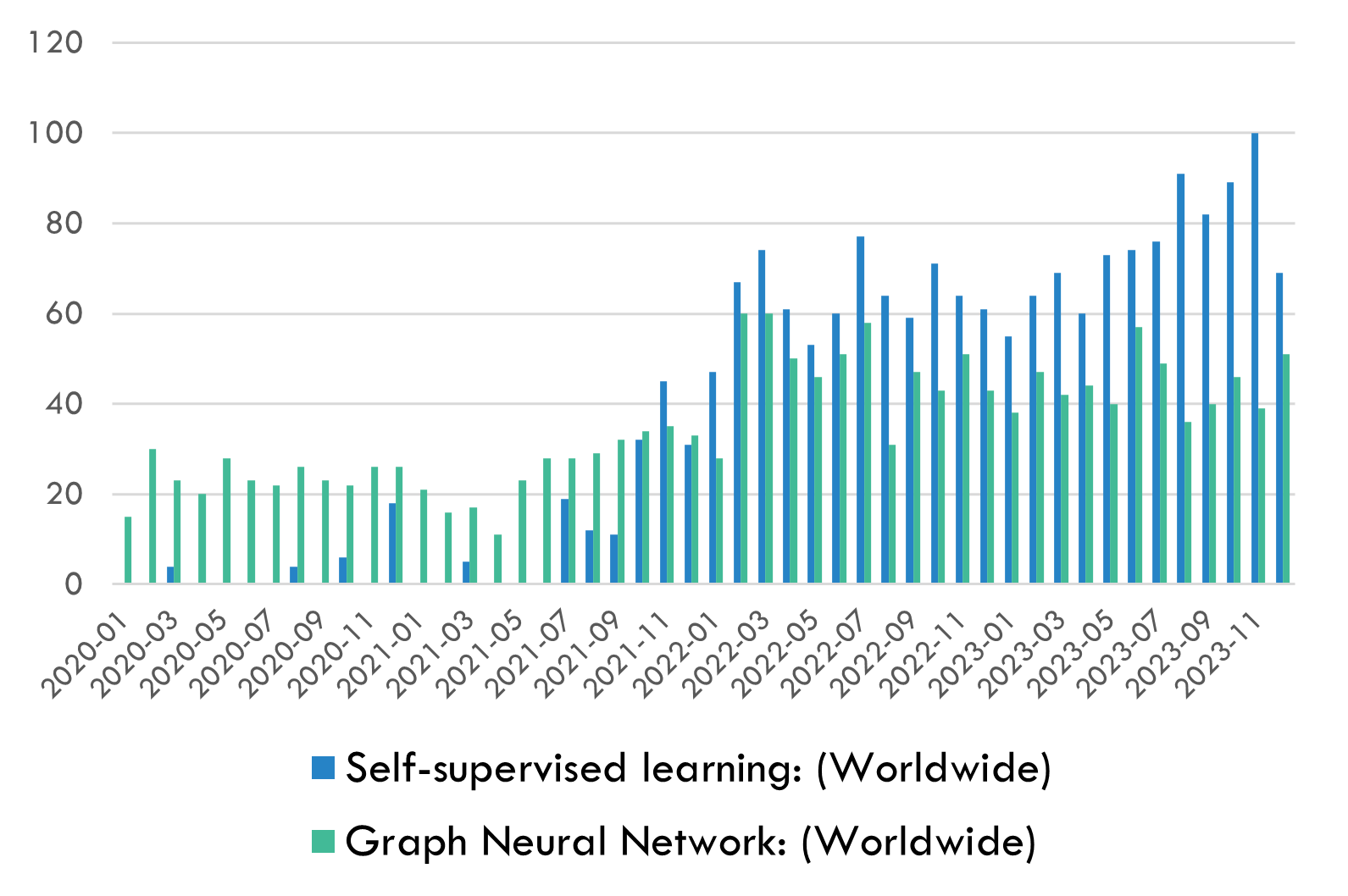}
    \caption{The number of Google searches for the terms Graph Learning and SSL from 2020 to 2024, according to Google trends.}
    \label{fig:trends}
\end{figure}

Expanding upon these innovations, the combination of Self-Supervised Learning (SSL) and graph learning has become a significant breakthrough in developing smart healthcare. SSL, a learning paradigm that gets insights from data without requiring explicit labeling, adds to the huge unstructured datasets that are common in healthcare. Graph learning, specifically Graph Neural Networks (GNNs), excels in modeling the complex connections and interdependencies inherent in healthcare data, such as patient histories, illness trends, and treatment results \cite{xia2021graph}. SSL and graph learning, when combined, open up new directions in healthcare analytics, improving the capacity to draw relevant insights and enhance decision-making in smart healthcare environments. These models can extract useful insights by anticipating missing parts or features inside a healthcare graph, assisting in patient similarity analysis, illness progression modeling, and medication development. These techniques not only exploit the complexity of healthcare data but also mitigate the challenges associated with the medical field's scarcity of labeled datasets. 
Data privacy is a significant challenge in this sector, which limits the availability of comprehensive datasets for training purposes. SSL and GNNs provide privacy-preserving mechanisms by reducing dependence on labeled data. This approach minimizes the need for sensitive patient information to be exposed. SSL enables pre-training on unlabeled data, allowing anonymized datasets to be used effectively. In this context, Figure \ref{fig:trends} showcases the growing attention towards graph learning and SSL. This increasing attention highlights the significance of these methodologies, which can transform data privacy challenges into opportunities for innovation and enhanced predictive accuracy.

In this comprehensive review, we examine the SSL and graph learning approaches in depth, particularly focusing on their applications and advancements in smart healthcare. We explore how these technologies are changing healthcare by boosting data analysis, increasing diagnostic accuracy, and customizing patient care. 
Our objective is to comprehensively understand these technologies, explaining their groundbreaking impact on smart healthcare and highlighting their potential for future advancements in this rapidly expanding field.

\subsection{Related Works} 
SSL has attracted much attention across various domains, particularly in the field of healthcare. Many researchers have been motivated by this rise in interest to carry out in-depth studies that review how SSL can improve healthcare systems. These surveys examine different aspects of SSL implementation, providing details about its theoretical foundations and real-world applications. This section investigates relevant surveys covering SSL use in healthcare and others covering Graph SSL. By thoroughly comparing these existing reviews and our work, we aim to clearly underscore the novel insights and significant contributions our survey introduces to the ongoing research about SSL and graph-structured data in healthcare applications. 
Table \ref{tab:surveys} provides an overview of all identified surveys published from 2020 to 2024. These include comprehensive surveys, which provide an in-depth overview of a specific area, and systematic surveys, which follow a structured approach to analyze studies using clearly defined criteria.  

Most recently, in 2024, several pivotal studies have highlighted the advancements and potential of SSL in healthcare \cite{c1,c2,c3,pani2024examining}. These works emphasize SSL as a medical data analysis tool covering the whole spectrum. As well as shedding light on how SSL research is currently being conducted, they discuss primary methodologies, applications, and challenges that are being encountered. They also highlight future directions and unresolved concerns. 

Liu \textit{et al.} \cite{c2} describe the fundamental principles of SSL and summarized their applications in cardiac, abdomen, and brain magnetic resonance imaging (MRI) segmentation. An extensive list of publicly available MRI segmentation datasets, along with the available online algorithms, is also presented in their review.
A range of recent studies spanning modalities, datasets, and methods was reviewed in \cite{c3} that explored self-supervised pre-training's impact on radiological imaging diagnostic tasks, including  X-ray, CT, MRI, and ultrasound. The surveys offer valuable insights. However, they do not explore the integration of graph-based methodologies with SSL to address challenges such as patient similarity analysis or drug discovery.

Works in \cite{huang2023self, krishnan2022self,shurrab2022self} present a deep examination of the use of SSL in medical applications. They focus on several SSL methodologies used in medical imaging and outline the advancements in SSL techniques to enhance medical smart services without needing extensively labeled data. 

Shurrab \textit{et al.} \cite{shurrab2022self} review the state-of-the-art SSL approaches that are adapted and implemented in medical imaging. They categorize SSL approaches into predictive, generative, and contrastive methods, detailing how each method works and its application for medical imaging. They also summarize the case studies implemented in real-world imaging tasks and focus on their effectiveness in enhancing the performance of medical analysis systems. 

Additionally, in an earlier work, Chowdhury \textit{et al.} \cite{chowdhury2021applying} offer a comprehensive review of SSL methodologies and their specific applications in the medical field. They categorize and explain different SSL approaches and report the medical applications covering the period from 2014 to 2020. All these surveys focus on SSL applications in medical imaging, such as classification and segmentation tasks. However, they do not address the use of graph-structured data or the potential of graph-based SSL methodologies in healthcare.

Expanding the focus to graph-based applications, Lu \textit{et al.} \cite{lu2023disease} provides a detailed exploration of graph machine learning (ML) methods applied to predicting diseases through the use of electronic health data (EHD). They categorize and explain different graph ML approaches and discuss their current challenges and opportunities. However, this work does not delve into the nuanced interplay between graph ML and the emerging paradigm of SSL. The discussion focuses more on graph-based methodologies, and it doesn't go into how SSL could make graph ML better at solving certain healthcare problems. 

Exploring the role of graph-based methodologies for SSL, the surveys in \cite{liu2022graph,xie2022self} provide a systematic review for a comprehensive understanding of the existent approaches in graph SSL. The papers introduce various SSL techniques specifically designed for graph data and categorize them into four main groups: generation, auxiliary property, contrast, and hybrid-based methods. 
Moreover, Liu \textit{et al.} \cite{liu2021self} explore different SSL methods, particularly generative and contrastive approaches. They discuss their applications, advantages, and challenges in different ML domains. They highlight the SSL evolution and its importance as an alternative to supervised learning due to its efficiency in using unlabeled data. They also discuss open problems in SSL, such as the design of effective pretext tasks, and outline future research directions to enhance the applicability of SSL methods. 
Lastly, Jaiswal \textit{et al.} \cite{jaiswal2020survey} present a comprehensive overview of contrastive SSL techniques and their effectiveness across various domains. They mainly focus on their applications in computer vision and NLP, highlighting how these methods use unlabeled data to extract useful representations and reduce the reliance on large labeled datasets.  These works categorize SSL techniques for graph data into various approaches but do not discuss their potential in solving healthcare problems. Conversely, the survey of Wang \textit{et al.} \cite{wang2023review} has centered on medical imaging, specifically examining the adaptation of cutting-edge contrastive SSL algorithms, initially conceived for natural images. They conclude by delving into recent developments, existing limitations, and potential avenues for future research in the application of contrastive SSL to the medical field.

The reviewed literature in the previous subsection showcases the importance of SSL as an innovative approach for improving data analytics across various domains, particularly in medical imaging and graph-based applications. The surveys on SSL in healthcare underline the importance of the transition from traditional supervised learning to SSL due to the privacy, high cost, and effort of labeling large datasets. They highlight SSL's ability to employ unlabeled data, which is plentiful in healthcare, for training models that achieve the same level of accuracy as those trained with supervised methods. Specific reviews focus on medical imaging, a critical area within healthcare AI, to underscore SSL's effective use in overcoming data annotation challenges. On the other hand, graph SSL surveys explore various methodologies, such as generation-based, contrast-based, and hybrid approaches, tailored specifically for graph data. They offer comprehensive insights into how these methods can be adapted to different applications, providing a deeper understanding of data. 
Although some research investigates the use of SSL in healthcare, there is a gap in the thorough examination of SSL methods within graph-structured data frameworks. This gap includes a lack of detailed exploration into their practical application, key benefits, and possible constraints within varied use cases. Our study makes a significant contribution to the field by introducing a comprehensive review that addresses these shortcomings. It presents and elaborates on the concept of SSL for graph-structured data in healthcare contexts. Differing from other studies, this review encompasses a broad spectrum of research, provides new insights, and highlights the transformative impact of SSL in this critical area.

\subsection{Purpose and Contribution of the Review}

This work aims to comprehensively assess and analyze previously published research on SSL applications in graph-structured data, particularly in healthcare. As SSL and GNN evolve within the healthcare domain, a comprehensive repository that compiles the work from these diverse aspects is necessary. The significant contributions of the presented review are illustrated in the following:
\begin{itemize}
\item As far as we know, this study is the first survey to thoroughly compare graph-based SSL methods used in the healthcare sector.

\item{An outline of the graph-based models in healthcare, their types, applicability, and performances is provided. }

\item {A review of the most commonly used GNN architectures for healthcare in the literature is presented, mathematically defined, and systematically compared.}

\item {The graph-based SSL methods are outlined with a comprehensive overview, detailing their categories and strategies of training.}

\item The various applications of SSL in graphs for healthcare are thoroughly reviewed, compared, and analyzed to derive insights and identify trends.

\item{The publicly available datasets and the used evaluation metrics for graph-structured data applied for healthcare applications are highlighted.}

\item{The challenges, limitations, and future research directions for graph-based SSL healthcare architectures are discussed.}

\end{itemize}

\begin{table}[H]
\centering
\caption{Comparative analysis of previous surveys}
\label{tab:surveys}
\resizebox{\textwidth}{!}{%
\begin{tabular}{p{2.2cm}p{2.5cm}p{2.5cm}p{1.5cm}p{6.5cm}p{5cm}}
\hline
 \textbf{Survey}  & \textbf{Journal} & \textbf{Methodology} & \textbf{Domain}     & \textbf{Main purpose} & \textbf{Limitations} \\
\hline

 \begin{tabular}[c]{@{}l@{}}Rani \textit{et al.} \\ \cite{c1} 2024\end{tabular} & \begin{tabular}[c]{@{}l@{}}Evolving \\Systems \end{tabular} & Comprehensive  & \begin{tabular}[c]{@{}l@{}}SSL for\\ Medical \\ imaging \end{tabular} & \begin{tabular}[c]{p{6.5cm}} Analyze recent research for medical image diagnosis tasks, focusing on studies comparing SSL pre-training to fully supervised learning \end{tabular} & \begin{tabular}[c]{p{5cm}} Identify only articles providing evidence of SSL pre-training impact on radiology diagnosis. \end{tabular} \\
\hline
 \begin{tabular}[c]{@{}l@{}}Liu \textit{et al.} \\ \cite{c2} 2024\end{tabular} & \begin{tabular}[c]{@{}l@{}}NMR in\\ Biomedicine \end{tabular} & Comprehensive  & \begin{tabular}[c]{@{}l@{}}SSL for\\ MRI \end{tabular} & \begin{tabular}[c]{p{6.5cm}} Examine existent approaches for MRI segmentation with focus on unlabeled data \end{tabular} & \begin{tabular}[c]{p{5cm}} Limited to MRI; Do not focus on SSL \end{tabular} \\
\hline
 \begin{tabular}[c]{@{}l@{}}VenBerlo \textit{et al.} \\ \cite{c3} 2024\end{tabular} & \begin{tabular}[c]{@{}l@{}}BMC\\ Medical\\ Imaging \end{tabular} & Comprehensive  & \begin{tabular}[c]{@{}l@{}}SSL for\\ medical \\ imaging \end{tabular} & \begin{tabular}[c]{p{6.5cm}} Cover many aspects of SSL in medical image analysis \end{tabular} & \begin{tabular}[c]{p{5cm}} Do not mention graph-structured data \end{tabular} \\
\hline
\begin{tabular}[c]{@{}l@{}}Pani \textit{et al.} \\ \cite{pani2024examining} 2024\end{tabular} & \begin{tabular}[c]{@{}l@{}}Multimedia\\ tools and\\ applications\end{tabular} & Comprehensive  & \begin{tabular}[c]{@{}l@{}}SSL for\\ Medical \\ imaging \end{tabular} & \begin{tabular}[c]{p{6.5cm}} Examine the application of SSL-based computational techniques in medical image analysis and explore the advantages and limitations of these methods 
\end{tabular} & \begin{tabular}[c]{p{5cm}} Limited to image analysis; Do not mention graph-structured data \end{tabular} \\
\hline
 \begin{tabular}[c]{@{}l@{}}Huang \textit{et al.} \\ \cite{huang2023self} 2023\end{tabular} & \begin{tabular}[c]{@{}l@{}}NPJ Digital \\ Medicine\end{tabular} & Systematic  & \begin{tabular}[c]{@{}l@{}}SSL for\\ medical \\ imaging\end{tabular} & \begin{tabular}[c]{p{6.5cm}} Provide a systematic analysis of prior research that apply SSL to medical imaging classification \end{tabular} & \begin{tabular}[c]{p{5cm}}Consider papers published after 2012, limiting to medical image classification and excluding graph data. \end{tabular} \\
\hline
 \begin{tabular}[c]{@{}l@{}}Krishan \textit{et al.}\\ \cite{krishnan2022self} 2022\end{tabular} & \begin{tabular}[c]{@{}l@{}}Nature \\Biomedical\\ Engineering\end{tabular} & Comprehensive & \begin{tabular}[c]{@{}l@{}}SSL in\\ Medicine \\and\\ healthcare\end{tabular} & \begin{tabular}[c]{p{6.5cm}} Review SSL strategies and demonstrate their role in enhancing medical services \end{tabular} & \begin{tabular}[c]{p{5cm}} Do not cover the use of graphs; Do not compare to previous studies \end{tabular} \\
\hline
 \begin{tabular}[c]{@{}l@{}}Shurrab \textit{et al.}\\\cite{shurrab2022self}2022\end{tabular}  &  PeerJ Computer Science & Comprehensive & SSL in medical imaging analysis & Review state-of-the-art SSL approaches in medical imaging analysis, categorizing them into predictive, generative, and contrastive methods. & Focus on computer vision methods; Underrepresentation of relational learning\\ \hline
 
 \begin{tabular}[c]{@{}l@{}}Chowdhury\\ \textit{et al.}\\ \cite{chowdhury2021applying} 2021\end{tabular} & Informatics & Systematic & \begin{tabular}[c]{@{}l@{}}SSL in\\ Medicine\end{tabular} & \begin{tabular}[c]{p{6.5cm}} Review and highlight the ability of SSL to leverage the abundant unlabeled data in medicine, therefore reducing the reliance on costly labeled datasets \end{tabular} & \begin{tabular}[c]{p{5cm}} Consider studies from 2014 to 2020; Did not cover the use of graph data \end{tabular} \\
\hline
 \begin{tabular}[c]{@{}l@{}}Lu \textit{et al.}\\ \cite{lu2023disease} 2023\end{tabular} & \begin{tabular}[c]{@{}l@{}} MDPI \end{tabular} & Comprehensive & \begin{tabular}[c]{@{}l@{}}Disease\\ prediction\end{tabular} & \begin{tabular}[c]{p{6.5cm}} Present a thorough literature review of studies that utilized graph ML for disease prediction using EHD. \end{tabular} & \begin{tabular}[c]{p{5cm}} Consider studies from 2015 to 2022; Do not cover the use of SSL; Limited to disease prediction using EHD \end{tabular} \\
\hline
 \begin{tabular}[c]{@{}l@{}}Liu \textit{et al.}\\ \cite{liu2022graph} 2022\end{tabular} & \begin{tabular}[c]{@{}l@{}}IEEE\\ Transactions on\\ Knowledge \\and Data\\ Engineering\end{tabular} & Comprehensive & Graph SSL & \begin{tabular}[c]{p{6.5cm}} Examine SSL techniques applied to graphs and organize them by creating a classification system with 4 groups: generation, auxiliary property, contrast, and hybrid base approaches \end{tabular} & \begin{tabular}[c]{p{5cm}} Do not mention healthcare applications \end{tabular} \\
\hline
\begin{tabular}[c]{@{}l@{}}Yaochen \textit{et al.}\\ \cite{xie2022self} 2022 \end{tabular} & \begin{tabular}[c]{@{}l@{}}IEEE\\ Transactions on\\ Pattern analysis \\and Machine\\ Intelligence\end{tabular} & Comprehensive & \begin{tabular}[c]{@{}l@{}}SSL \\ on graphs\end{tabular} & \begin{tabular}[c]{p{6.5cm}} Propose a unified framework for analyzing SSL methods on graphs by categorizing these methods into two main approaches: contrastive and predictive, then compare them. \end{tabular} & \begin{tabular}[c]{p{5cm}} Do not mention healthcare applications \end{tabular} \\
\hline
 \begin{tabular}[c]{@{}l@{}}Liu \textit{et al.}\\ \cite{liu2021self} 2021\end{tabular} & \begin{tabular}[c]{@{}l@{}}IEEE \\Transactions \\on Knowledge\\ and Data\\ Engineering\end{tabular} & Comprehensive & SSL & \begin{tabular}[c]{p{6.5cm}} Review the current state of SSL, compare generative and contrastive methods, and identify effective strategies for implementing SSL in different domains \end{tabular} & \begin{tabular}[c]{p{5cm}} Do not mention healthcare applications; Do not focus on graph-structured data \end{tabular} \\
\hline
\begin{tabular}[c]{@{}l@{}}Jaiswal \textit{et al.}\\ \cite{jaiswal2020survey} 2020\end{tabular} & Technologies & Comprehensive & SSL & \begin{tabular}[c]{p{6.5cm}} Provide a review of contrastive SSL methods and its applications in computer vision and natural language processing \end{tabular} & \begin{tabular}[c]{p{5cm}} Focus only on contrastive SSL; Did not mention healthcare applications; Did not focus on graph structure \end{tabular} \\
\hline
 \begin{tabular}[c]{@{}l@{}}Wang \textit{et al.}\\ \cite{wang2023review} 2023 \end{tabular} & \begin{tabular}[c]{@{}l@{}}Machine\\ Intelligence \\Research \end{tabular} & Comprehensive & \begin{tabular}[c]{@{}l@{}}SSL in \\ Medicine\end{tabular} & \begin{tabular}[c]{p{6.5cm}} Investigate the applicability of advanced contrastive SSL algorithms, successful with natural images, to the domain of medical imaging \end{tabular} & \begin{tabular}[c]{p{5cm}} Focus only on contrastive SSL; Do not consider graph data \end{tabular} \\
\hline

\end{tabular}%
}
\end{table}

\subsection{Structure of the Paper}

\begin{figure}[h!]
    \centering
    \includegraphics[scale=0.285]{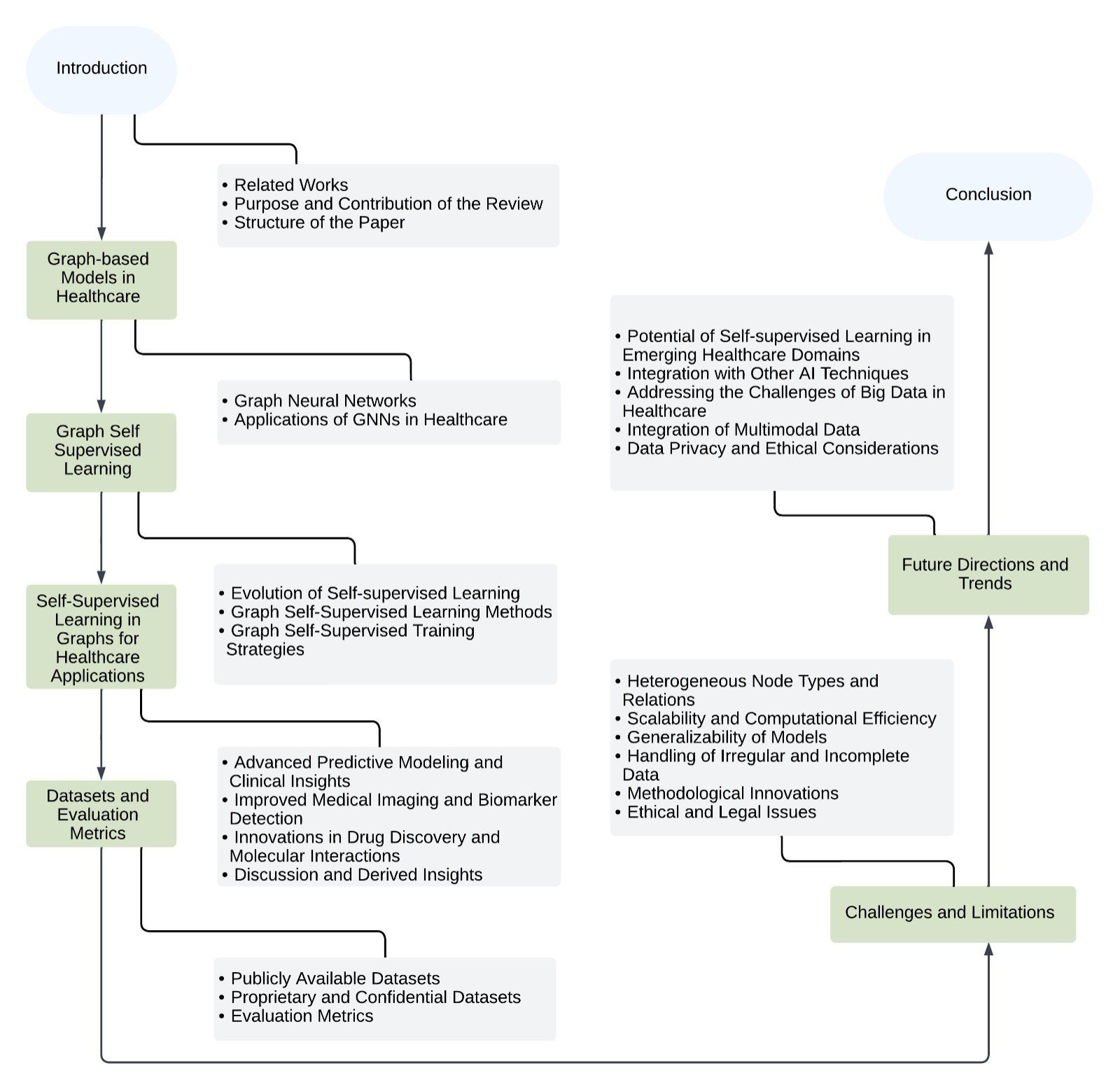}
    \caption{Visual overview of the paper structure depicting key sections and subsections.}
    \label{fig:diag}
\end{figure}

The rest of this survey is structured as follows. Section \ref{sec2} covers the main graph-based models in healthcare. Section \ref{sec3} gives an overview of graph SSL. Section \ref{sec4} discusses the applications of graph-based SSL in the healthcare domain. Section \ref{sec5} classifies the used medical benchmark datasets and presents the metrics used to evaluate and compare the existing approaches. Sections \ref{sec7} and \ref{sec8} provide the open issues and future directions. Finally, Section \ref{sec9} wraps up the paper. A diagram of the survey structure is illustrated in Figure \ref{fig:diag}. 
\section{Graph-based Models in Healthcare}
\label{sec2}
The graph-based models have been proposed as an effective analysis and prediction tool for enhancing healthcare applications. They have a strong ability to capture complex dependencies and interactions within medical data by modeling relationships between data entities as graph edges. This section will review the foundational GNN architecture and its variants proposed in the literature, including GCNs, GraphSAGE, GATs, and GAEs. Then we will review the applications of these models in healthcare.

\subsection{Graph Neural Networks (GNNs)}
\begin{figure}[h]
    \centering
    \includegraphics[scale=0.58]{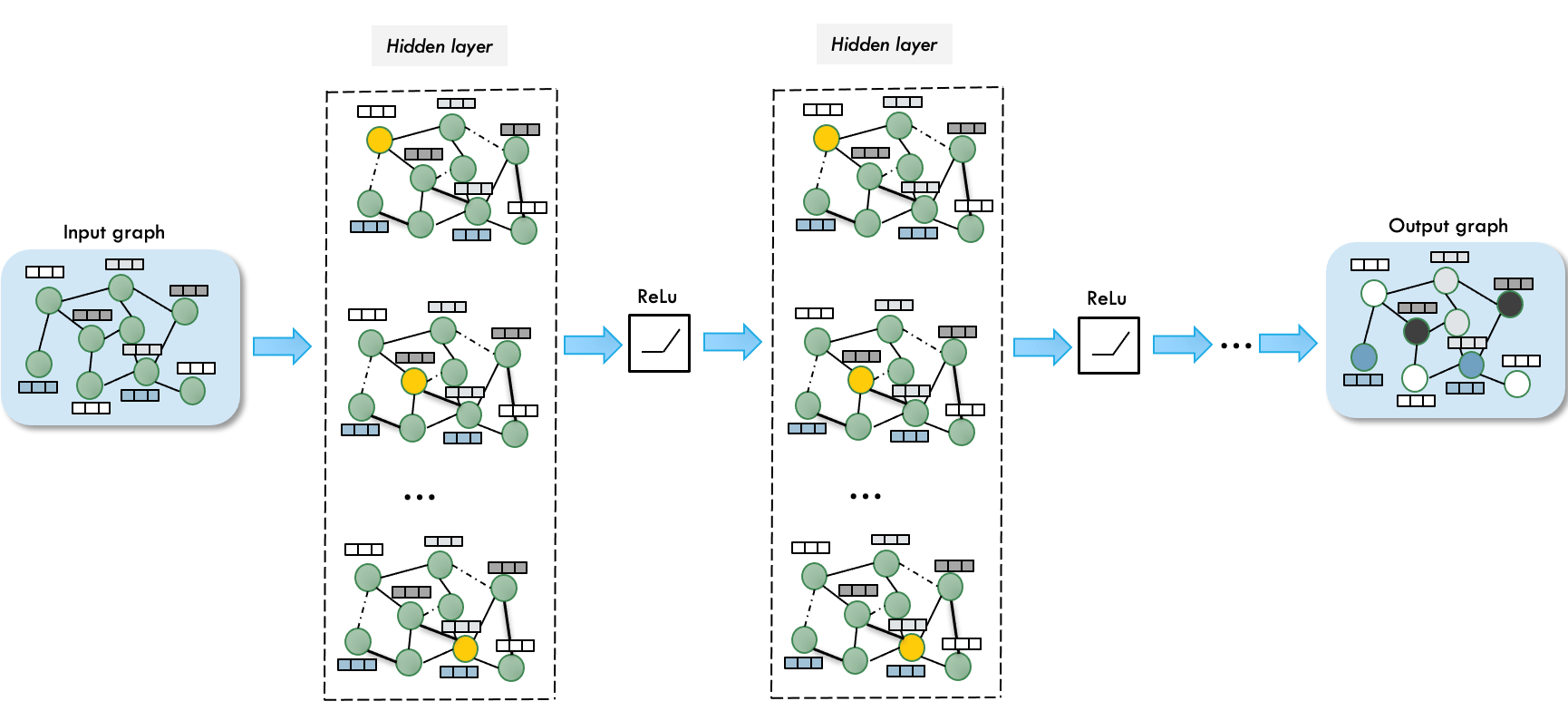}
    \caption{A general GNN architecture.}
    \label{fig:genr}
\end{figure}

GNNs, introduced by Gori \textit{et al.} \cite{gori2005new}, present an important advancement for neural networks, specifically tailored to process graph-structured data. They have lately garnered significant attention for their potential to revolutionize healthcare \cite{lu2023disease, paul2024systematic}. 
Graphs are mathematical structures that include nodes and edges, representing connections, to model relationships in different domains. GNNs extend traditional neural networks by incorporating information about the complex connections between nodes which makes them highly adept at tasks involving relational data. In the field of healthcare research, where data is inherently interconnected and relational, GNNs offer a powerful tool for extracting meaningful insights. One of the main strengths is their ability to model complex relationships. They are well-suited for applications in healthcare analytics, drug discovery, and personalized medicine.
The typical GNN architecture is depicted in Figure \ref{fig:genr}.

While GNNs have shown good performance in dealing with graph-structured data, they have several drawbacks \cite{wu2020comprehensive}. A major drawback is the computational cost involved with the model's hierarchical feature-extraction technique. In each iteration of this approach, the same parameters are used as the network transfers information from neighboring nodes through a neural network, eventually reaching a stable fixed state to acquire the node's representation. This iterative procedure, especially for big networks, can be computationally intensive. Furthermore, the GNN model fails to describe important features on the graph's edges. The inability to collect edge features correctly restricts the model's to extract complicated relationships from data.

Different GNN architectures have been developed to address the limitations of conventional GNNs. Each architecture introduces special mechanisms to enhance the model's performance in specific aspects. The following review the main GNN architectures adapted for healthcare applications. A comparative overview of different GNN architectures is depicted in Table \ref{tab:GNN}.

\subsubsection{Graph Convolutional Networks (GCNs)}
\label{subsec2}
GCNs are a class of GNN, introduced by Kipf \textit{et al.} \cite{kipf2016semi}, and designed to learn from graph-structured data. 
The fundamental idea of GCNs is to extend the concept of convolutional processes from regular grids to irregular graph structures. In standard Convolutional Neural Networks (CNNs), filters slide over a regular grid of pixels to capture local patterns and hierarchical features. However, graph-structured data lacks a fixed grid which makes it challenging to apply convolutional operations directly. GCNs address this challenge by defining a localized convolutional operation for each node in the graph. The convolutional operation aggregates information from a node's neighbors, allowing the model to capture the structural relationships within the graph. The propagation rule in a GCN layer can be expressed mathematically as:
\\
\\
\begin{equation}
    h_i^{(l+1)} = \sigma\left(\sum_{j\in \mathcal{N}(i)} \frac{1}{c_{ij}} W^{(l)}h_j^{(l)}\right)
\end{equation}

where $i$ represents the target node, $N_{i}$ denotes the set of neighbors of node $i$, $h_{i}^{l}$ is the node $i$ representation at layer $l$, $W^{l}$ is the layer weight matrix, and $c_{ij}$ is a normalization factor for the varying numbers of neighbors.

The graph convolution operation is carried iteratively across different layers, enabling the model to capture increasingly complex relationships and higher-order dependencies in the graph.

Therefore, GCNs address the limitations of conventional GNNs by introducing localized feature learning, which focuses on aggregating information from neighboring nodes to enable efficient feature extraction. Additionally, they simplify training by using shared weight matrices across layers. Despite these advantages, GCNs face challenges such as limited receptive fields, over-smoothing in deeper layers, and an inability to explicitly model edge features.

\subsubsection{Graph Sample and Aggregation (GraphSAGE)}

GraphSAGE is a GNN architecture proposed by Hamilton \textit{et al.} in \cite{hamilton2017inductive}, designed to be scalable for large graphs.
It is intended to learn node embeddings by sampling and aggregating information from a graph's local neighbors. The main concept is to use inductive learning to allow the model to be generalized to nodes not included during training.
GraphSage is made up of two components: the embedding generation procedure (forward propagation) and the parameter learning process. 
The GraphSAGE embedding generation method iteratively traverses several search depths, allowing nodes to progressively gather information from their local surroundings. The representation update for node \(v\) at each depth \(k\) in the GraphSAGE model is captured by the following equation:

\begin{equation}
h_v^{(k)} = \sigma\left(W_k \cdot {CONCAT}\left(h_v^{(k-1)}, AGGREGATE_k\left(\{h_u^{(k-1)}, \forall u \in N(v)\}\right)\right)\right)
\end{equation}

where \(h_v^{(k)}\) represents the node \(v\) at depth \(k\), incorporating information from the previous representation and the aggregated neighborhood information,
\(\sigma\) is the non-linear activation function that introduces non-linearity to the model, \(W_k\) presents the learnable weight matrix associated with depth \(k\), {CONCAT} is the concatenation operation, combining the previous node representation \(h_v^{(k-1)}\) with the aggregated neighborhood information, \(N(v)\) is the set of neighbors of the node \(v\), and \(AGGREGATE_k\), is the differentiable aggregator function, responsible for combining information from neighboring nodes. the aggregation function can be either mean,  long short-term memory (LSTM), or pooling aggregator.

The GraphSAGE embedding generation equation encapsulates the iterative process through which node representations evolve. This allows the model to capture complex relationships and dependencies within the graph. The dynamic updating mechanism enables GraphSAGE to generate expressive embeddings for nodes in a scalable and inductive manner.

\subsubsection{Graph Attention Networks(GATs)}
In 2018, GATs have been proposed by Velickovic \textit{et al.} \cite{velickovic2017graph} They are a class of GNN that handle some of the shortcomings of classic GNNs, especially in the capture of complicated connections and the management of irregular graph topologies.
The use of attention mechanisms to selectively aggregate input from nearby nodes, allowing nodes to focus on more relevant information throughout the learning process, is the main innovation of GATs. This attention method allows nodes to contribute various weights to their neighbors, stressing the significance of specific nodes during the aggregation process. The GAT layer can be defined as follows:
\begin{equation}
    h_i^{(l+1)} = \sigma\left(\sum_{j\in \mathcal{N}(i)} \alpha_{ij} W^{(l)}h_j^{(l)}\right)
\end{equation}
where $i$ represents the target node, $N_{(i)}$ is the group of neighbors of node $i$, $h_{i}^{(l)}$  is the representation of node $i$ at layer $l$, $W^{(l)}$ is a shared weight matrix for layer $l$, $\alpha_{ij}$ is the attention weight assigned to the edge between nodes $i$ and $j$, computed using a learnable attention mechanism.

GATs are highly adaptable and expressive to capture complex relationships, making them suitable for tasks like personalized treatment planning and disease prediction. However, they face significant challenges, particularly high computational costs and difficulties in scaling to large graphs. Despite these issues, their ability to dynamically weight neighbors and enhance feature learning makes them a valuable tool for graph-based tasks.


\subsubsection{Graph Auto-Encoders (GAEs)}
In 2016, graph auto-encoders \cite{kipf2016variational} were proposed to address scenarios involving graph-based data. In a GAE, the graph structure is encoded into a low-dimensional latent space, preserving structural information while reducing dimensionality. The GAE architecture is composed of two main components; the encoder and the decoder.

The encoder takes the graph structure as input and produces a low-dimensional representation (embedding) of the graph. Let $h_i^l$ denote the representation (or embedding) of node $i$ in layer $l$. The encoder operation can be defined as:
    
\begin{equation}
h_i^{l+1} = \sigma \left( \sum_{j \in \mathcal{N}(i)} W^l h_j^l \right)
\end{equation}

where, \(\mathbf{\mathcal{N}}(i) \) is the set of neighboring nodes of node $i$, $W^l$ is the weight matrix for layer $l$, and $\sigma$ is the activation function, such as ReLU or sigmoid. The input graph is represented as an adjacency matrix $A$, where $A_{ij} = 1$ if there is an edge between nodes $i$ and $j$, and $0$ otherwise.

The decoder takes the low-dimensional representation generated by the encoder and reconstructs the original graph structure. 
     Let $\hat{A}$ be the reconstructed adjacency matrix. The decoder operation can be formulated as:

\begin{equation}
\hat{A} = g_{\text{decoder}}(\mathbf{Z})
\end{equation}

where $g_{\text{decoder}}$ is the decoder function, which may involve fully connected layers followed by appropriate activation functions to generate the reconstructed adjacency matrix.

GAEs are widely applied in link prediction, anomaly detection, and graph generation. They have a strong ability to efficiently compress graph structures into informative representations. However, challenges include scalability to large graphs and the difficulty of modeling dynamic or heterogeneous graph structures.


\begin{table}[H]
    \centering
    \caption{Comparative overview of Graph Neural Network Architectures in healthcare.}
    \label{tab:GNN}
     \resizebox{\textwidth}{!}{
    \begin{tabular}{llll}
    \hline
        \textbf{GNN Model }&	\textbf{Key Feature}	& \textbf{Main Strengths}	&\begin{tabular}[c]{@{}l@{}} \textbf{Primary Applications}\\ \textbf{in healthcare}  \end{tabular} \\ \hline
        \vspace{0.5cm}
        \begin{tabular}[c]{@{}l@{}}GCN \end{tabular} &\begin{tabular}[c]{@{}l@{}} Localized convolution \\operation for graphs\end{tabular} &\begin{tabular}[c]{@{}l@{}} Efficient in learning spatial hierarchies and \\dependencies by leveraging node connectivity. \end{tabular} & \begin{tabular}[c]{@{}l@{}}Drug discovery \cite{sun2020graph}\\ Protein interaction networks \cite{jha2022prediction} \end{tabular} \\ 
        \vspace{0.5cm}

        \begin{tabular}[c]{@{}l@{}}GraphSAGE \end{tabular} & \begin{tabular}[c]{@{}l@{}} Sampling and aggregating \\neighbor node information\end{tabular} &\begin{tabular}[c]{@{}l@{}} Scalable to large graphs due to inductive \\learning and neighborhood sampling.\end{tabular} &\begin{tabular}[c]{@{}l@{}} Patient similarity networks\\Predictive healthcare models \cite{hamilton2017inductive} \end{tabular} \\ \vspace{0.5cm}

        \begin{tabular}[c]{@{}l@{}} GAT \end{tabular} &\begin{tabular}[c]{@{}l@{}} Attention mechanism to \\weigh neighbor contributions \end{tabular} &\begin{tabular}[c]{@{}l@{}} The ability to concentrate on the most\\ pertinent aspects of the graph,\\ improving the learning of features. \end{tabular} & \begin{tabular}[c]{@{}l@{}} Personalized treatment plans\\ Disease prediction \cite{bian2021gatcda,ji2021predicting} \end{tabular}\\ 
        \vspace{0.5cm}
        
        \begin{tabular}[c]{@{}l@{}} GAE \end{tabular} &\begin{tabular}[c]{@{}l@{}}Based on an encoding-decoding \\mechanism to learn latent \\representations \end{tabular} &\begin{tabular}[c]{@{}l@{}} Effective in learning compact and continuous\\ node representations, very useful for \\(un/self)-supervised learning.\end{tabular} & \begin{tabular}[c]{@{}l@{}} Synthetic electronic health records \\generation \cite{nikolentzos2023synthetic}\end{tabular}\\ 

        \hline
         
    \end{tabular}}
   
\end{table}

\subsection{Applications of GNNs in Healthcare}
GNNs have shown considerable potential in healthcare applications, owing to their capacity to represent complex relations and dependencies in structured data. In the following, we provide an overview of different applications of GNNs in healthcare, including disease prediction and diagnosis, medical image analysis, and drug discovery. The reviewed works are outlined in Table \ref{tab:gnn_healthcare}.

\subsubsection{Disease Prediction and Diagnosis}
GNNs are increasingly employed for disease prediction and diagnosis by analyzing patient health records and modeling relationships between various medical entities \cite{lu2023disease}. These capabilities make GNNs well-suited for tasks like disease risk prediction, early diagnosis, and prognosis, contributing to advancements in personalized medicine. For example, Lu \textit{et al.} \cite{lu2021weighted} have used the GNN to handle high-dimensional data problems and predict chronic diseases. 
Firstly, they prepare the health data and create a bipartite graph to get a weighted patient network. Then, they employ different GNN models to generate robust patient representations for predicting chronic diseases.  Recent studies \cite{sun2020disease, zheng2022multi} have introduced new methodologies to enhance the use of GNNs in disease prediction. In \cite{sun2020disease}, the authors integrate external knowledge bases in the model to improve the limited electronic medical record data and enable the construction of medical concept graphs. These graphs help learn high-representative node embeddings for patients, diseases, and symptoms. Similarly, Zheng \textit{el al.}  \cite{zheng2022multi} use modality-aware representation learning and adaptive graph learning to discern latent graph structures automatically. This approach optimizes the use of available data and ensures that the models can adapt to unseen data effectively.

GNNs significantly enhance disease prediction and diagnosis capabilities, especially for rare diseases. They effectively leverage sparse and limited data. However, in disease diagnosis, the interpretability of GNNs remains a key focus area. Researchers are actively exploring ways to make GNN predictions more transparent and understandable \cite{zheng2024bpi,zheng2024ci}, which is crucial in clinical settings. Enhanced interpretability fosters trust among healthcare professionals and ensures that GNN-based solutions are both reliable and actionable.

\subsubsection{Medical Image Analysis}
GNNs have demonstrated great performance when applied to analyze medical figures, including X-rays and MRIs \cite{zhang2023graph}. Enhancing image segmentation, object detection, and disease classification are their potential applications.
Different works have been proposed to detect serious diseases, including pneumonia, COVID-19, Autism spectrum disorder, and Alzheimer's disease. Works \cite{lee2022chexgat,bagwan2023precise,song2023covid} used Chest X-ray images with different architectures of GNN to detect Covid-19. Authors in \cite{gaggion2022improving} introduced a novel encoder-decoder neural architecture that integrates standard convolutions for image feature encoding with GCN for decoding anatomical structures. It is specifically designed to address topological errors and anatomical inconsistencies in medical image segmentation.

In \cite{kumar2022sars}, Kumar \textit{et al.} present SARS-Net, a system combining GCN and CNN to detect abnormalities in Chest X-ray images for COVID-19 diagnosis. 

The use of GNNs in medical imaging has been widely recognized for their high accuracy and sensitivity in identifying anomalies and delivering precise medical diagnoses. They excel at leveraging relational data, making them particularly efficient for tasks like segmentation. GNNs also demonstrate scalability to handle large datasets effectively, and their adaptability makes them suitable for a variety of diagnostic tasks. However, these advantages are accompanied by challenges. GNNs demand significant computational resources for training and inference, which pose restrictions especially in resource-limited healthcare settings. Additionally, the effectiveness of GNNs depends on the availability of high-quality annotated datasets, which are scarce and costly to collect in the medical field. Addressing these challenges is essential to fully recognize the potential of GNNs in medical imaging and diagnostics.
 
\subsubsection{Drug Discovery and Repurposing}
Drug discovery and repurposing are critical areas in healthcare where GNNs have demonstrated transformative potential. By analyzing molecular graphs, GNNs can predict chemical properties, identify promising drug candidates, and suggest novel therapeutic uses for existing drugs \cite{gaudelet2021utilizing,han2021reliable}. 

These capabilities make GNNs valuable for uncovering drug-target interactions and hidden relationships between drugs and diseases. In this context,
Cheung \textit{et al.} \cite{cheung2020graph} present an approach based on GNNs for identifying potential treatments for SARS-CoV-2, the virus responsible for COVID-19. Their approach highlights the ability of GNNs to predict key molecular features crucial for drug discovery.
Li \textit{et al.} \cite{li2020learn} develop a molecular pre-training graph-based framework to address the challenges of generating expressive molecular representations in AI-driven drug discovery. The model is designed to effectively learn complex molecular structures. It can also generate interpretable representations of molecules to help understand chemical insights.
Bongini \textit{et al.} \cite{bongini2021molecular} proposed a novel approach to drug discovery through GNN, specifically focusing on generating molecular graphs. This method effectively creates potential new drug molecules, representing a significant advancement in computational drug design.
Moreover, Wang \textit{et al.} \cite{wang2024accurate} have used the GNNs to predict drug-drug interactions, a significant issue in clinical treatments and drug development. They leverage the rich neighborhood information available in biomedical knowledge graphs. Their approach focuses on learning a knowledge subgraph for each drug pair and using connection strengths to make accurate predictions.
Luo \textit{et al.} \cite{luo2024prediction} combine GCNs with reinforcement symmetric metric learning to predict potential drug-disease associations.
Han \textit{et al.} \cite{han2021reliable} introduced a distance-aware GNN model to improve reliability in drug discovery. They proposed the GNN-SNGP architecture. Their model demonstrated enhanced robustness and reliability.

In the field of drug discovery, GNNs excel at modeling complex and irregular data structures, such as those found in biomolecular data. This capability enables detailed and accurate representations of drugs, diseases, and their interactions. Their ability to capture complex relationships allows them to outperform standard ML methods in tasks like drug-disease interaction prediction and drug repositioning. GNNs are particularly effective in applications that rely on understanding relational data. They address challenges like data sparsity by generalizing effectively from known to unknown data. This capability is important in drug discovery, where new compounds and disease targets are constantly emerging. However, GNNs face significant challenges, including high computational demands and the need for high-quality annotated datasets, which are often scarce or expensive. Overcoming these limitations requires strategies such as developing more efficient architectures and leveraging SSL in limited labeled data scenarios.

\begin{table}[]
\centering
\caption{Applications of GNNs in healthcare }
\label{tab:gnn_healthcare}
\begin{tabular}{lllp{8cm}}
\hline
\textbf{Reference}                                                                    & \textbf{Application}                              & \textbf{Model}   & \textbf{Details}                                                                                                     \\ \hline
Lu \textit{et al.} \cite{lu2021weighted}            & \multirow{3}{*}{\begin{tabular}[c]{@{}l@{}} Disease Prediction \\and Diagnosis\end{tabular}} & GNN              & Handles high-dimensional data for predicting chronic diseases.                                                       \\ 
Sun \textit{et al.} \cite{sun2020disease}           &                                                   & GNN              & Integrates external knowledge bases to construct medical concept graphs for learning representative node embeddings. \\ 
Zheng \textit{et al.} \cite{zheng2022multi}         &                                                   & GNN              & Uses modality-aware representation learning and adaptive graph learning to discern latent graph structures.          \\ \hline
Lee \textit{et al.} \cite{lee2022chexgat}           & \multirow{5}{*}{\begin{tabular}[c]{@{}l@{}}Medical Image \\Analysis \end{tabular}}           & Hybrid (CNN-GNN) & Proposes a hybrid DL framework combining convolutional and GNNs for detecting COVID-19, and leverages implicit disease correlations for improved diagnostic performance.                                                                     \\
Bagwan \textit{et al.} \cite{bagwan2023precise}     &                                                   & IsoCovNet        &  Converts medical imaging data into graph representations, and employs GNN architectures for COVID-19 detection.                                                                    \\ 
Song \textit{et al.} \cite{song2023covid}           &                                                   & GNN              & Utilizes GNN architectures to predict the influence of COVID-19-infected individuals on future infections by incorporating interaction data and individual properties.                                                   \\ 
Gaggion \textit{et al.} \cite{gaggion2022improving} &                                                   & HybridGNet       & Proposes a novel encoder-decoder architecture combining convolutions and GCN for medical image segmentation.         \\
Kumar \textit{et al.} \cite{kumar2022sars}          &                                                   & SARS-Net         & Combines GCN and CNN to detect abnormalities in chest X-rays for COVID-19 diagnosis.                                 \\ \hline
Han \textit{et al.} \cite{han2021reliable}          & \multirow{6}{*}{ \begin{tabular}[c]{@{}l@{}} Drug Discovery \\and Repurposing \end{tabular}}   & GNN-SNGP         & Introduces a new GNN architecture and a benchmark dataset for drug discovery.                                        \\ 
Cheung \textit{et al.} \cite{cheung2020graph}       &                                                   & GNN              & Predicts key molecular features for drugs that can inhibit SARS-CoV-2 (COVID-19).                                    \\ 
Li \textit{et al.} \cite{li2020learn}               &                                                   & MolGNet          & Develops a pre-training graph-based deep learning framework for generating expressive molecular representations.     \\ 
Bongini \textit{et al.} \cite{bongini2021molecular} &                                                   & MG\(^2\)N\(^2\)  & Generates potential new drug molecules through molecular graphs.                                                     \\ 
Wang \textit{et al.} \cite{wang2024accurate}        &                                                   & KnowDDI          & Predicts drug-drug interactions by learning a knowledge subgraph for each drug pair.                                 \\ 
Luo \textit{et al.} \cite{luo2024prediction}        &                                                   & RSML-GCN         & Combines GCNs with reinforcement symmetric metric learning.                                                          \\ \hline 
\end{tabular}
\end{table}

\section{Graph Self-Supervised Learning}
\label{sec3}
SSL is presented as an advanced ML approach that falls under the general category of unsupervised learning. Without any explicit external labels, the system learns to predict and classify using input data itself \cite{jaiswal2020survey}. This approach is especially beneficial when labeled data is rare or expensive to get. While unsupervised learning focuses on the model behavior with data, the SSL uses the data itself to guide the learning process. Usually, SSL is based on pretext tasks, designed to extract meaningful representations from unlabeled data. These tasks are developed such that the labels are created automatically from the data, eliminating the need for human annotation. 
In the following, we will discuss SSL's evolution, provide an overview of SSL approaches and models, and review SSL for graph representation learning.

\subsection{Evolution of Graph Self-supervised Learning}
\begin{figure}[h]
    \centering
    \includegraphics[scale=0.8]{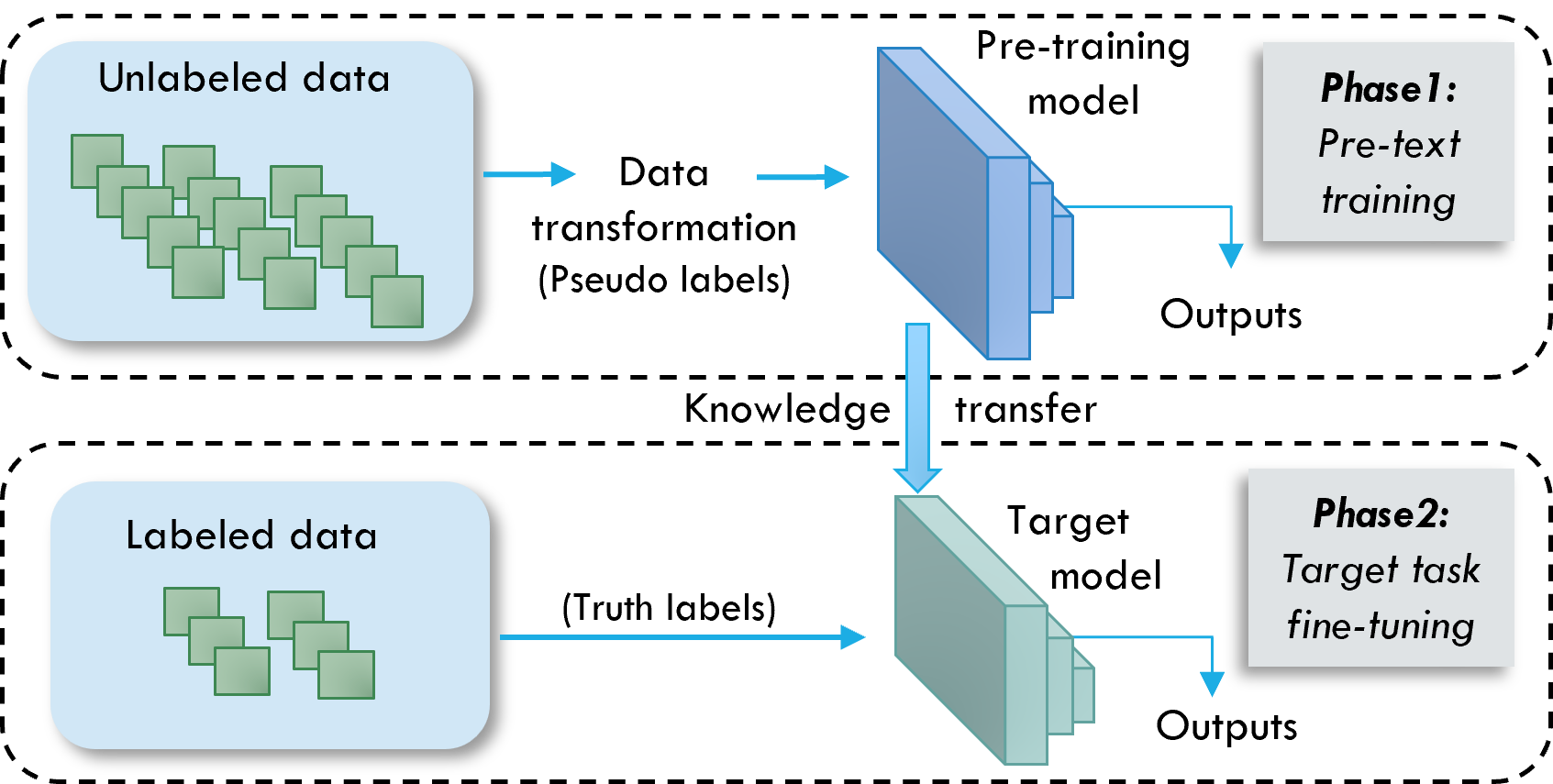}
    \caption{The standard SSL framework.}
    \label{fig:SSLframework}
\end{figure}

SSL has evolved significantly in the ML paradigm \cite{rani2023self}. This strategy changes the focus from mainly depending on large labeled datasets to exploiting unlabeled data. It addresses challenges, including the high cost of obtaining high-quality labeled data. Over the past decade, a large part of research interests has been focused on using SSL in different fields. Developing an SSL model has two primary phases: the pretext, and the downstream tasks \cite{zhang2023dive}. Figure \ref{fig:SSLframework} illustrates the standard SSL framework.

The model is trained on a self-supervised task with unlabeled data during the initial phase. The main goal in this phase is to get useful representations and features from data without the need for explicit labels. Pretext tasks can include predicting missing parts of the input, generating data augmentations, and solving puzzles created from the input data. The pretext task should help the model learn meaningful and important features for the target task.

Once the model has been trained on the pretext task, it is ready to be fine-tuned on a specific downstream task that requires some labeled data. The learned representations from the pretext task are used as a starting point. Then, the model is adapted to the specific task, such as image classification and object detection. Fine-tuning the downstream task helps the model to specialize and improve its performance on the target task.

Representation learning in graphs is represented as an important aspect of graph-based ML, where the goal is to learn meaningful and informative representations of nodes, edges, and graphs \cite{khoshraftar2024survey}. Graph representation learning aims to capture the structural and relational information present in the graph to enable better downstream tasks such as node classification, edge prediction, and graph classification.
With the increasing availability of unlabeled graph data, SSL has emerged as a powerful technique for graph representation learning, particularly in domains like healthcare.

In the context of graphs, SSL techniques are used to learn meaningful graph representations from unlabeled graph data \cite{wang2022graph}.
These techniques leverage the abundance of unlabeled graph data, which is often easier to obtain than labeled data. It also help in learning more robust and generalizable graph representations, as the model learns to capture the underlying graph structure. The pretext tasks for SSL in graphs are different from the tasks used in computer vision and include predicting missing nodes or edges, reconstructing the graph from a subgraph, and predicting the presence of an edge between two nodes.

\subsection{Graph Self-Supervised Learning Methods}
\begin{figure}[ht]
    \centering
    
    \begin{subfigure}[b]{\textwidth}
        \centering
        \includegraphics[scale=0.8]{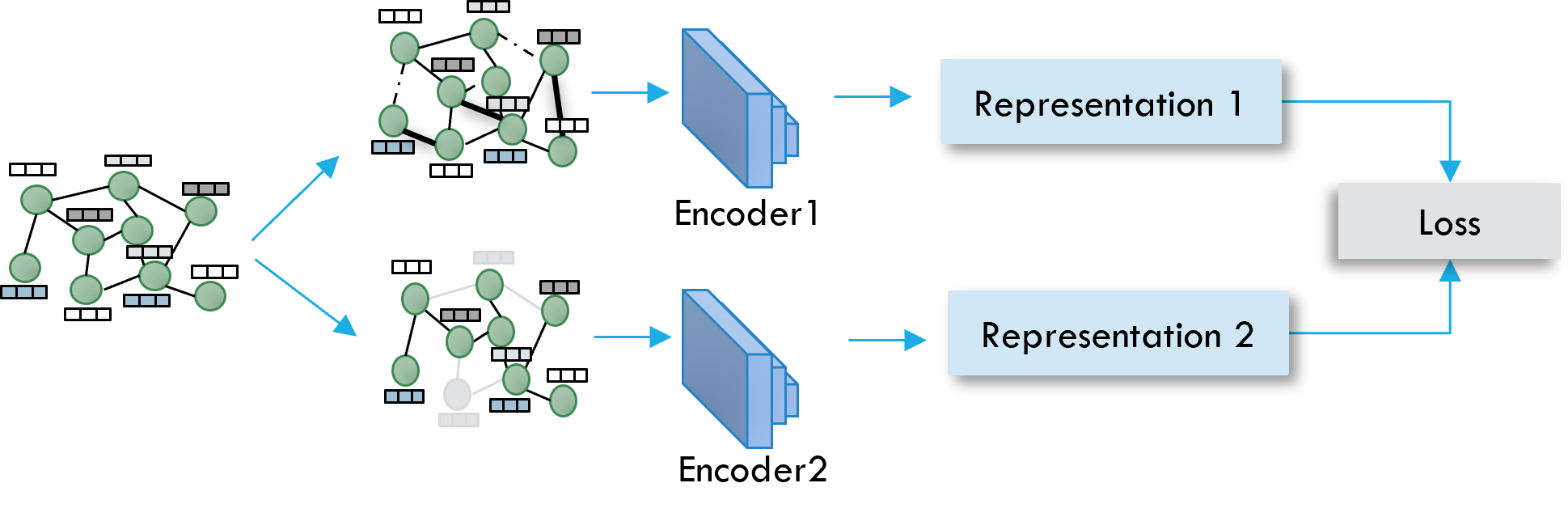}
        \caption{Contrastive learning.}
        \label{fig:subfig-a2}
    \end{subfigure}

    \begin{subfigure}[b]{\textwidth}
        \centering
        \includegraphics[scale=0.8]{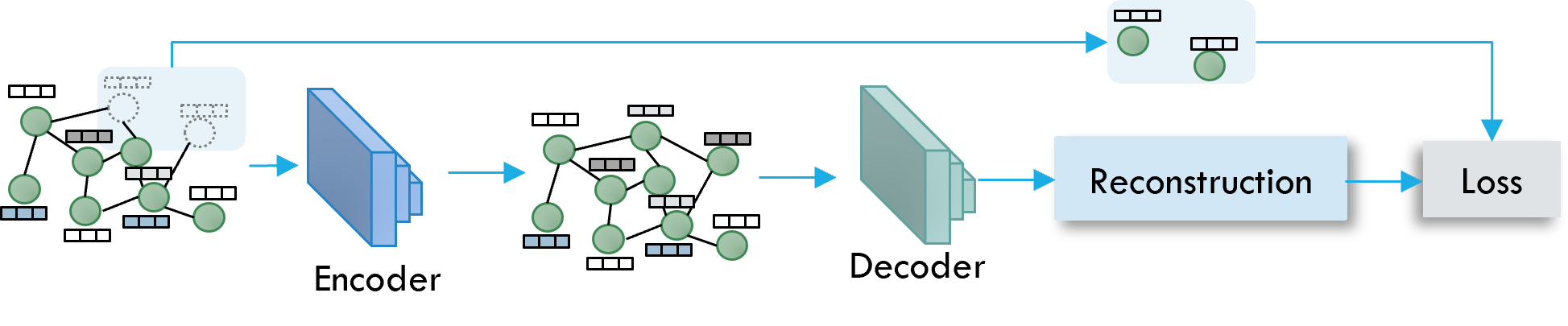}
        \caption{Generative learning.}
        \label{fig:subfig-b2}
    \end{subfigure}
    \vspace{1cm}
    \begin{subfigure}[b]{\textwidth}
        \centering
        \includegraphics[scale=0.8]{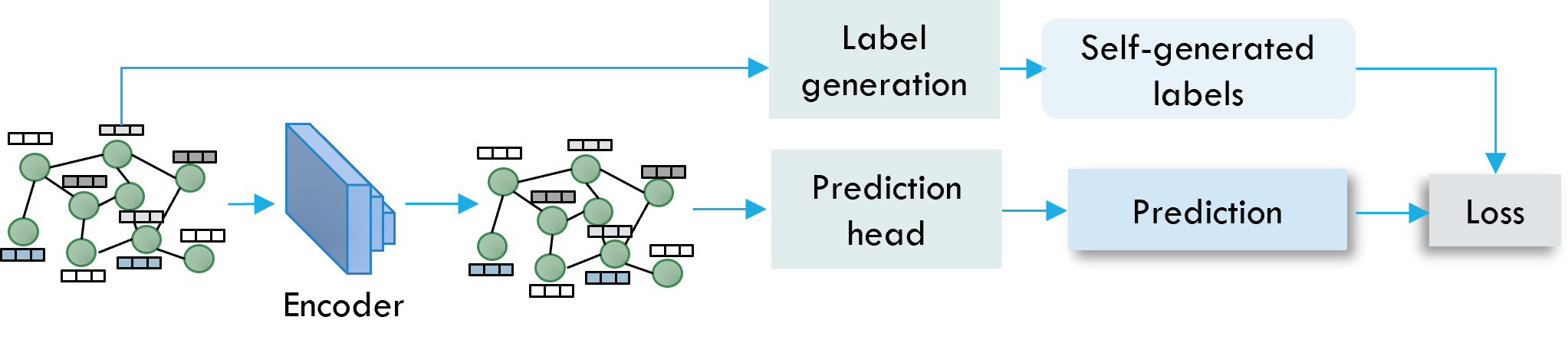}
        \caption{Predictive learning.}
        \label{fig:subfig-c2}
    \end{subfigure}
    \vspace{-1.5cm}
    \caption{The graph SSL categories.}
    \label{fig:SSLcategories}
\end{figure}

In this subsection, we will delve into the different methodologies presented for graph-based SSL. We categorize this approach into three groups: contrastive, generative, and predictive, as depicted in Figure \ref{fig:SSLcategories}. A comparative overview of these methods is provided in Table \ref{tab:SSLcompar}.

\subsubsection{Contrastive Learning}
The contrastive learning method involves models that learn by comparing samples to distinguish between similar and dissimilar instances effectively \cite{jaiswal2020survey,wu2021self}. It is particularly effective in teaching models to identify similarities and differences that define different classes in datasets. This method enhances the model's ability to see patterns in both the similarities and contrasts across data sets. The primary objective is to train the model to link similar samples while separating distinct ones. The fundamental concept of this process consists of providing pairs of data samples to the model and asking it to identify which pairs (positive pairs) are part of the same group and which pairs (negative pairs) are not.
Following that, the model gains the strength to extract relevant variables through continuous classification prediction and repeated exposure to several pairings.
The model then moves on to explicit training on labeled data. At this stage, the focus is on target tasks like detection or classification. By utilizing the knowledge gained from contrastive learning, this transition enables the model to perform with good precision and effectiveness.

In this learning method, we will delve into and discuss the essential steps that contribute to its effectiveness. These include the augmentation techniques for graph transformation, various methods for structuring pretext tasks, and the development of the contrastive objective.

\begin{enumerate}
    \item \textbf{Graph Transformation: Augmentation Techniques}\\
Graph data requires specialized augmentation techniques to enhance contrastive learning effectively. Unlike images, graphs are structured in non-Euclidean data, making the use of image augmentation methods like cropping and cutouts unsuitable \cite{yu2022graph}. 
The primary concept here is to introduce alterations within graphs to find positive pairs. Different augmentation techniques are used for graphs,  some of them concentrate on modifying the graph's structure, while others keep the graph structure and perturb node features to ensure the representations are invariant to the initial node attributes. Below, we outline some augmentation techniques used for generating positive graph pairs, which are depicted in Figure \ref{fig:augmentations}. \\
  \\
\textbf{Node Feature Perturbation:}
\begin{itemize}
    \item \textit{Node Feature Masking:} 
    In this augmentation method, certain features of a specific set of nodes are masked to generate an augmented graph \cite{hu2019strategies, cui2023smg}. 
    This masking involves sampling each feature entry from a Gaussian distribution with predetermined mean and variance. The goal is to train the model using representations with different node features but keep the same graph structure.

    \item \textit{Node Feature Shuffling:} 
    The node feature shuffling involves randomly arranging the features of nodes within the graph \cite{zhu2021graph, ali2024features}. This technique perturbs the node features while keeping the graph structure intact. By shuffling the attributes, the goal is often to reduce the dependency of the learned representations on specific node features and encourage the model to focus more on the graph's structural properties.   
\end{itemize}

\textbf{Sub-Graph Sampling:} 
\begin{itemize}
    \item \textit{Random Walk Sampling:}
    In this method, a random walk is conducted on the graph, continuously adding nodes until a predetermined fixed number of nodes is reached \cite{jiao2020sub}. A sub-graph is then formed from these nodes. Traversal involves moving from the current node to a randomly selected neighboring node along an edge during the random walk.
    \item \textit{Knowledge Sampling:} In this method, experts or domain-specific individuals are consulted to select a subset of nodes or edges from the graph based on their expertise or knowledge \cite{subramonian2021motif}. Therefore, the sampled representation will collect the more important and relevant portions of the graph for a specific task or analysis.
\end{itemize}

\textbf{Graph Structure Alteration:}
\begin{itemize}
    \item  \textit{Edge Perturbation:} In this augmentation method, the edges are randomly added or removed from an existing graph to generate new graphs \cite{suresh2021adversarial,yu2024sparse}. Usually, a maximum fraction of edges is identified for perturbation to ensure that the fundamental structure of the graph remains unchanged.

    \item \textit{Node Dropping:} 
    In this augmentation technique, a small fraction of nodes are randomly dropped to create new graphs \cite{ding2022data,yu2024sparse,jin2024sadr}. Following that, all edges connected to the dropped nodes are also removed.

    \item \textit{Diffusion:} In diffusion-based augmentation, the focus is on transforming the graph's adjacency matrix into a diffusion matrix using a heat kernel \cite{ding2022data}. This process provides a global graph view by capturing relationships and interactions between nodes beyond immediate neighbors. The resulting diffusion matrix represents the spread of influence or information across the entire graph.
\end{itemize}

\begin{figure}[h]
    \centering
    \includegraphics[scale=0.8]{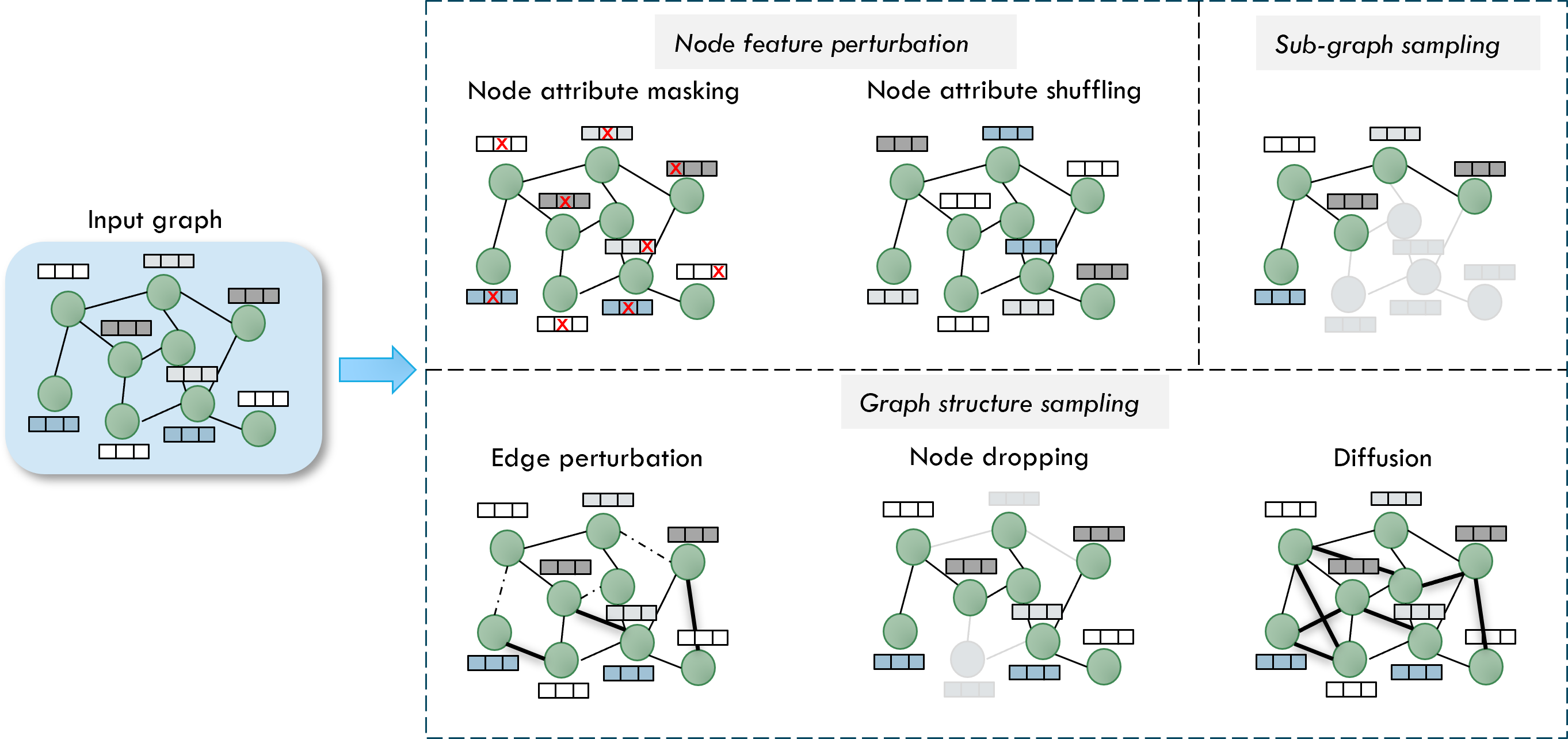}
    \caption{The commonly used graph augmentation techniques.}
    \label{fig:augmentations}
\end{figure}

    \item \textbf{Graph Contrastive Learning: Pretext Tasks}\\

In the context of contrastive learning, the goal is to maximize the mutual information between positive pairs of instances, which are usually obtained from comparable semantic content \cite{zeng2021contrastive}. By maximizing mutual information, contrastive learning encourages the model to capture and encode meaningful relationships and patterns present in the data. This process facilitates the extraction of informative representations that will be used for downstream tasks. Moreover, maximizing mutual information enables the model to discern subtle similarities and differences between samples and enhances its ability to discriminate between semantically similar and dissimilar samples.

In the following, we explore how different models define and address these pretext tasks:

\textbf{Graph-Level Pretext Tasks:}
\begin{itemize}
    \item \textit{Graph Clustering:} The objective here is to cluster graphs with similar semantic structures and use these clusters as labels for contrastive learning. 
    \item \textit{Graph Partitioning:} The graph is divided into sub-graphs based on structural or domain knowledge. Then, these sub-graphs are contrasted with the original graph as positive pairs. 
\end{itemize}

\textbf{Node-Level Pretext Tasks}
\begin{itemize}
    \item \textit{Neighbor Prediction:} The aim of this method is to predict whether two nodes are neighbors in the graph. The positive pairs are created from adjacent nodes while the negative pairs are from non-adjacent nodes.
    \item \textit{Node Context Prediction:} Based on random walks, node sequences are created, and then these nodes are contrasted based on their shared or distinct contextual embeddings.
\end{itemize}

\textbf{Edge-Level Pretext Tasks}
\begin{itemize}
    \item \textit{Link Prediction:} The goal is to predict whether an edge exists between two nodes in a graph. Positive pairs represent real edges, while negative pairs are non-existent edges.
    \item \textit{Edge Attribute Matching:} In this method, the edges are contrasted based on their attributes or weights. 
\end{itemize}

Graph-specific contrastive learning pretext tasks effectively address the unique challenges of non-Euclidean graph data. They enable models to learn robust and generalized representations. These tasks empower the model to learn informative and transferable features for downstream tasks.

    \item \textbf{Contrastive Objective: Mutual Information Evaluation}\\
Building the contrastive objective serves as the SSL model's main framework, enabling it to distinguish and integrate various structural insights from the data. 

The contrastive objective within the field of contrast-based methods is evaluating the mutual information between instances. This approximation is the foundation for training models to identify similarities and differences efficiently.
The idea behind this is simple: representations of positively linked instances are brought closer to each other, while instances that are negatively linked are pushed away from one another.

Given a pair of instances, $(x_{i}, x_{j})$ sampled from the positive pairs, their respective representations are $(r_{i}, r_{j})$. The MI between these representations, denoted by  $MI(r_i, r_j)$ is a pivotal metric for guiding the training process. It is calculated using the Kullback-Leibler (KL) divergence as follows:

\begin{equation}
    MI(r_i,r_j) = KL (P(r_i, r_j)||P(r_i)P(r_j))
    =E_{p_{(r_i,r_j)}}[ log \frac{P(r_i, r_j)}{P(r_i) P({r_j)}}]
\end{equation}

where $ P (r_i, r_h) $ represents the joint density of the representations $ r_i$ and $ r_j$, and $ P (r_i) $ and $ P (r_i) $ are their marginal densities. The objective is to equip the encoder with discriminative capabilities and enable it to distinguish between instances sampled from the joint density $P(r_i, r_j)$ and those from the marginal densities $P(h_i)$ and $P(h_j)$. 
 
\end{enumerate}

Contrastive learning is the most used technique in SSL due to its strong performance. However, its scalability and efficiency are constrained when used with very large graphs, posing a challenge in real-world applications.

\subsubsection{Generative Learning}
The generative learning method trains models to reconstruct data based on unlabeled inputs. This method is based on the idea of autoencoders, which compress and then reconstruct data \cite{hinton2006reducing,baldi2012autoencoders}. In contrast to conventional data formats, graphs require different techniques and specific management for linked components. Rebuilding particular graph elements, such as graph structure or node features, is a common task for generative-based SSL methods proposed for graphs. 
In the following, we provide a summary of the graph generative methods related to node feature and graph structure generation.

\textbf{Node Feature Generation}
\\
In SSL, Node feature generation methods focus on reconstructing the feature information from the original or the augmented graph structures. These methods are formalized through an optimization problem that aims to minimize the Mean Squared Error loss function $L_{mse}$, using Equation \ref{eq:7}. They also use a feature regression decoder $d_{\phi}(.)$, which can be a fully connected network, to reconstruct features based on the learned representations.

\begin{equation}
\theta^*, \phi^* = \arg \min_{\theta, \phi} L_{\text{mse}} \left( d_{\phi} \left( f_{\theta} ({G}) \right), \hat{X} \right)
\label{eq:7}
\end{equation}

where $G$ is the considered graph data, and $\hat{X}$ represents the feature matrices, which can be, as an example, the node feature matrix.

Some methods used for feature generation are:
\begin{itemize}
    \item \textit{Masked Feature Regression:} In this strategy, some node or edge features are masked during preprocessing and then regenerated using information from unmasked components \cite{liu2024developing}. 
    \item \textit{Graph Completion:} This method predicts the features of masked nodes using the features of neighboring nodes \cite{liu2023self,you2020does}. 

    \item \textit{Attribute Masking:} In this approach, the model is trained to reconstruct masked node features. The encoder’s parameters, once trained on this task, are used to initialize the encoder for downstream tasks \cite{kim2024hypeboy}.
\end{itemize}

\textbf{Graph Structure Generation }\\
Graph structure generation methods focus on reconstructing the structural information of graphs, primarily through the recovery of the adjacency matrix, which concisely represents the graph's topological structure \cite{zang2023hierarchical}. The graph structure generation methods are developed following equation \ref{eq:sg}.

\begin{equation}
\label{eq:sg}
\theta^*, \phi^* = \arg \min_{\theta, \phi} L_{\text{ssl}} \left( d_{\phi} \left( f_{\theta} ({G}) \right), AM \right)
\end{equation}

where, $p_{\phi}(.)$ is the structure decoder and $AM$ represents the adjacency matrix.

Some methods used for structure generation methods are:
\begin{itemize}
    \item \textit{Adjacency Matrix Reconstruction:} This method directly reconstructs the adjacency matrix of the graph. The adjacency reconstruction task ensures the learned embeddings capture graph topology.
    \item \textit{Edge Prediction:} Instead of reconstructing the entire adjacency matrix, this method focuses on predicting the existence of edges between nodes.

    \item \textit{Random Walk-based Generation:} This method rebuilds the graph structure by learning the patterns in how nodes interact within the sampled subgraphs. It uses random walks to create subgraphs and sequences to sample.

    \item \textit{Diffusion Matrix Reconstruction:} This approach provides a more comprehensive view of network structure by capturing global interactions and impacts between nodes through the reconstruction of a diffusion matrix generated from the graph's adjacency matrix.

\end{itemize}

\subsubsection{Predictive Learning}

The predictive learning method focuses on predicting missing parts of the input data to aid the model in learning significant temporal and contextual patterns.
Predictive learning in graph-based models uses self-generated labels for supervision, which focuses on leveraging relationships between data and labels. This approach differs from contrastive methods emphasizing data-data pair interactions and generative methods focusing on intra-data information. Predictive learning methods are particularly effective for deriving meaningful insights from graphs without requiring extensive manual labeling \cite{xu2021predictive,tang2021self}.  

This type of learning focuses on extracting robust representations from unlabeled data by converting the pretext task into a classification or regression problem. Depending on the selected pretext task, a pseudo label is assigned to each unlabeled sample based on the data itself. Some widely used predictive learning tasks include:

\begin{itemize}
    \item \textit{Node Property Prediction:} This task involves predicting node-specific attributes based on the graph structure and other node features.
    \item \textit{Edge Property Prediction:} This task focuses on predicting attributes associated with edges, such as edge weights, types, or existence.
    \item \textit{Edge Prediction:} This is a common predictive task where the model predicts whether an edge exists between two nodes.
    \item \textit{Mask Prediction:} The model predicts masked features of nodes, edges, or subgraphs using the information from neighboring nodes and edges.    
     
\end{itemize}

Predictive learning excels in healthcare graph-based applications. It can address a wide range of topics, including molecular graphs, by translating pretext problems into pseudo-label predictions. This approach enhances representations by encoding rich temporal and structural context, which improves downstream task performance.

\begin{table}[H]
    \centering
    \caption{Comparative overview of contrastive, predictive, and generative Self-Supervised Learning methods for graph data.}
    \resizebox{\textwidth}{!}{
    \begin{tabular}{cllll}
    \hline
     \textbf{Aspect }   & \textbf{Primary Goal} & \textbf{Core Mechanism} & \textbf{Loss of objective} & \textbf{Key challenges}\\ \hline
      \begin{tabular}[c]{@{}l@{}}Contrastive \\graph SSL \end{tabular}  & \begin{tabular}[c]{@{}l@{}} Maximize the similarity\\ between representations \\of similar pairs  and \\distinguish from \\dissimilar pairs. \end{tabular}  & \begin{tabular}[c]{@{}l@{}} Employs contrasting pairs \\to learn robust representations\\ by emphasizing differences\\ and similarities. \end{tabular}  & \begin{tabular}[c]{@{}l@{}} Mutual Information between\\ representations of positive\\ and negative instance pairs. \end{tabular}  & \begin{tabular}[c]{@{}l@{}} Scalability issues with \\large graphs,\\ maintaining balance between\\positive and negative\\ pair selection.\end{tabular} \\
      
      \begin{tabular}[c]{@{}l@{}} Generative \\graph SSL \end{tabular}  &  \begin{tabular}[c]{@{}l@{}}  Learn to reconstruct \\input data from \\unlabeled examples. \end{tabular}  & \begin{tabular}[c]{@{}l@{}}  Employs architectures like\\ Autoencoders that attempt \\to regenerate the original \\graph data or its attributes. \end{tabular}  &   \begin{tabular}[c]{@{}l@{}}MSE in feature \\reconstruction, fidelity of \\regenerated structural \\features.  \end{tabular}  & \begin{tabular}[c]{@{}l@{}} Balancing the fidelity\\ of reconstruction with \\the generalization capability \\of the model to handle\\ unseen graph variations. \end{tabular}  \\
      
      \begin{tabular}[c]{@{}l@{}} Predictive \\graph SSL \end{tabular}  &  \begin{tabular}[c]{@{}l@{}} Predict missing parts \\of data to learn \\significant patterns and \\relationships within\\ the graph. \end{tabular}  & \begin{tabular}[c]{@{}l@{}} Uses auto-generated labels\\ for supervision based \\on the data itself, \\often through the prediction \\of missing node properties \\or relationships. \end{tabular}  &  \begin{tabular}[c]{@{}l@{}} Accuracy of predicted \\pseudo-labels, efficacy\\ in capturing complex \\graph relationships. \end{tabular}  & \begin{tabular}[c]{@{}l@{}} Dependence on the quality\\ and relevance of self-\\generated labels for \\effective learning.\end{tabular} \\
      
      \hline

    \end{tabular}}
    
    \label{tab:SSLcompar}
\end{table}

\subsection{Graph Self-Supervised Training Strategies}
\begin{figure}[h!]
    \centering
    
    \begin{subfigure}[b]{\textwidth}
        \centering
        \includegraphics[scale=0.7]{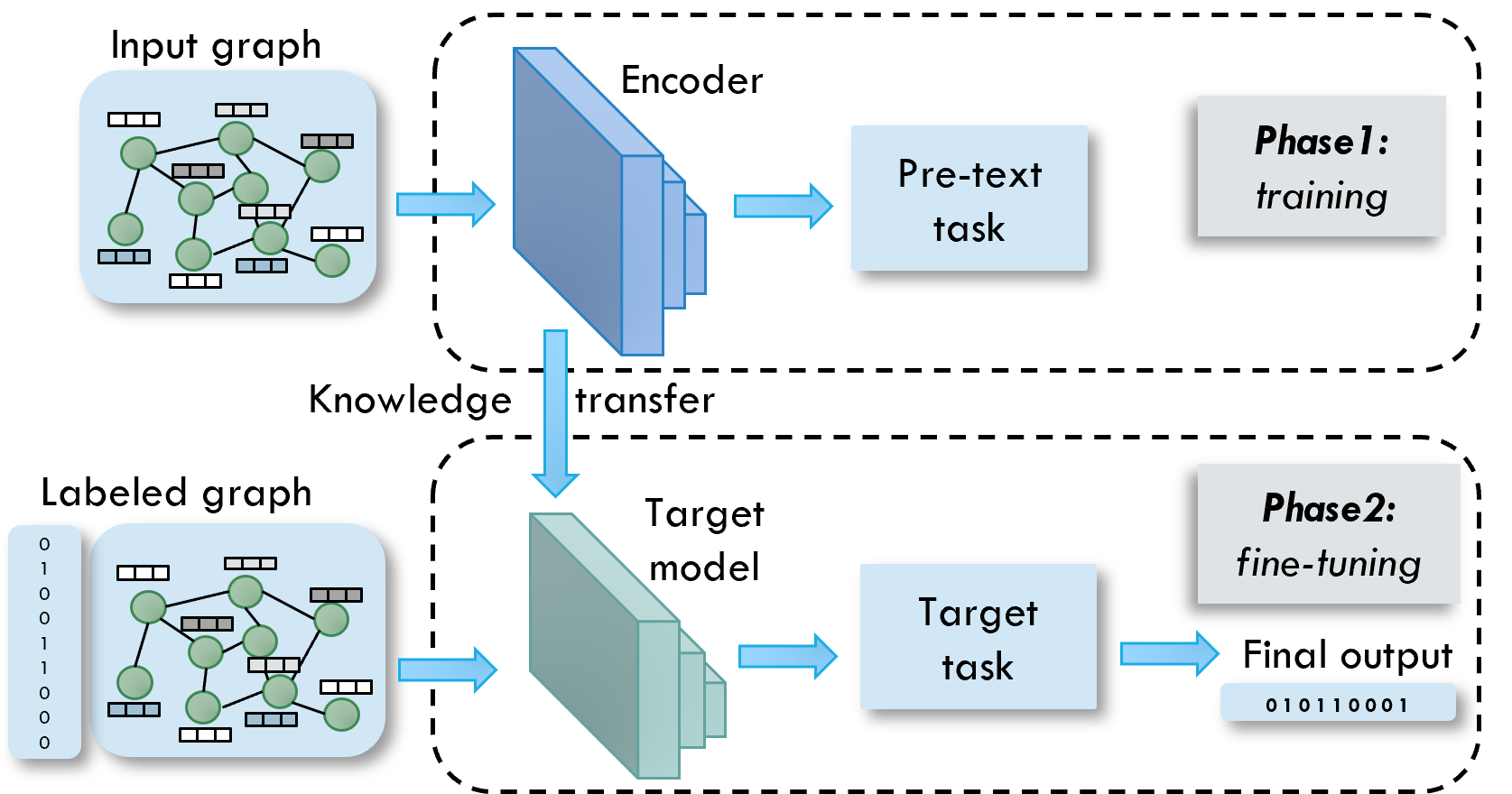}
        \caption{Pre-training and fine-tuning learning strategy.}
        \label{fig:subfig-a}
    \end{subfigure}

    \begin{subfigure}[b]{\textwidth}
        \centering
        \includegraphics[scale=0.7]{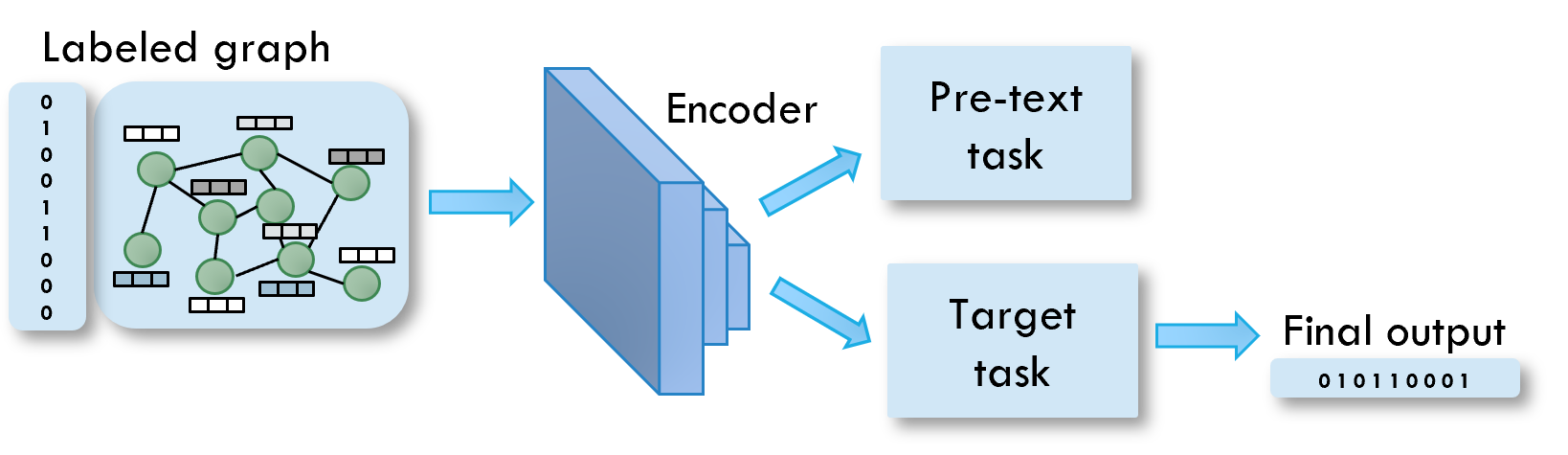}
        \caption{Joint training strategy.}
        \label{fig:subfig-b}
    \end{subfigure}
    \vspace{1cm}
    \begin{subfigure}[b]{\textwidth}
        \centering
        \includegraphics[scale=0.7]{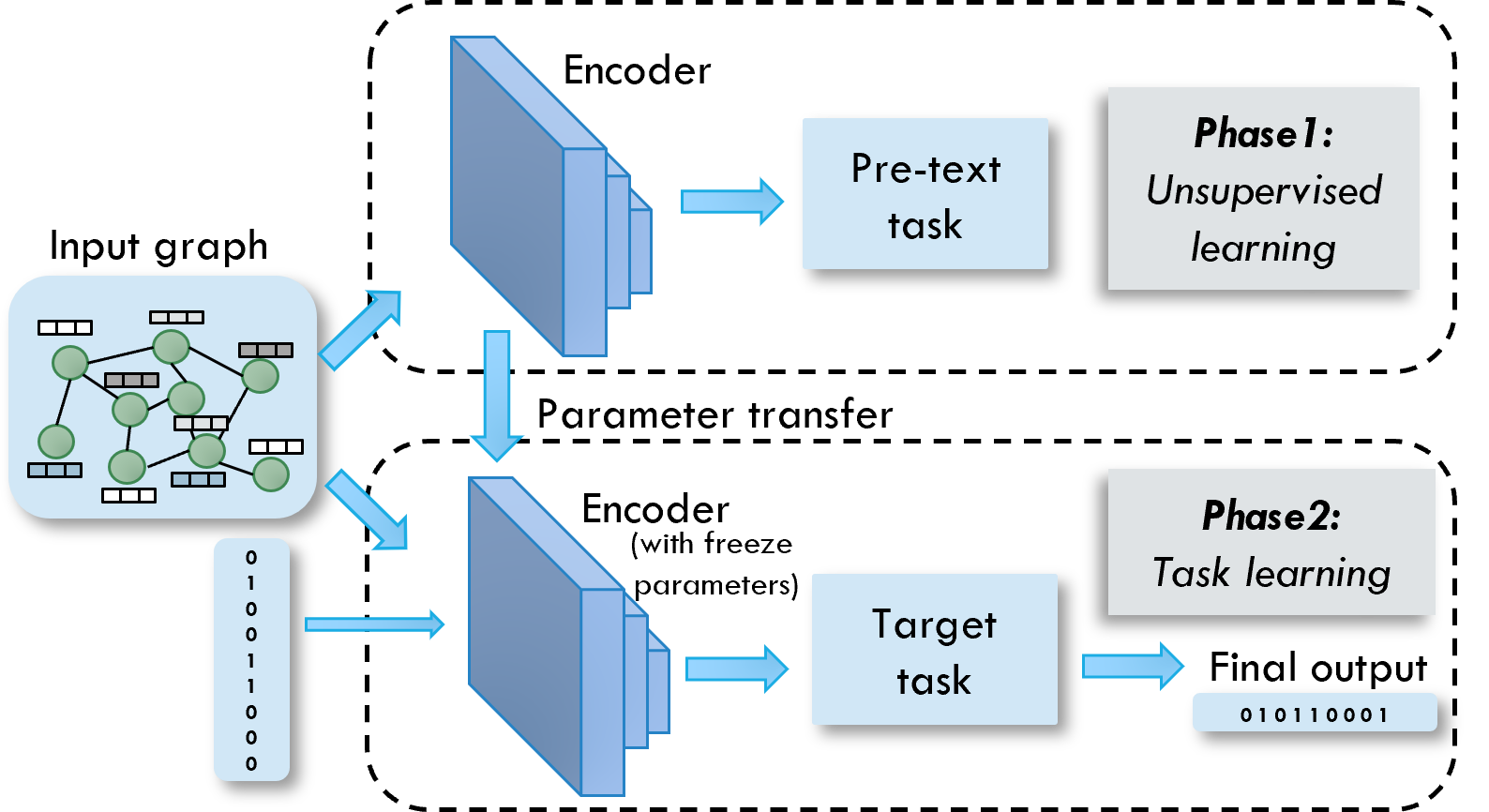}
        \caption{Unsupervised representation learning strategy.}
        \label{fig:subfig-c}
    \end{subfigure}
    \vspace{-1.5cm}
    \caption{The common training strategies for graph SSL.}
    \label{fig:SSLtraining}
\end{figure}

In this subsection, we delve into the different strategies used for training GNN following SSL. We report three prominent training strategies widely adopted in the literature, including pre-training with fine-tuning, joint training, and unsupervised representation learning. Each of these strategies holds its unique strengths to improve the adaptability and performance of GNN models. Figure \ref{fig:SSLtraining} provides a detailed visualization of each strategy, and Table \ref{tab:SSLstrategies} presents a comparative overview of these strategies, highlighting their primary goals, training phases, key benefits, and challenges.

\subsubsection{Pre-training and Fine-tuning (PF)}

The pre-training and fine-tuning strategy starts by pre-training the model on a given pretext task using a dataset that does not include labeled data \cite{zhao2024survey}. In this stage, the primary goal is to establish solid initial parameters for the model's encoder. Following this, the pre-trained encoder undergoes the fine-tuning stage. This stage integrates the encoder with a downstream task-specific decoder on labeled datasets to refine the pre-trained learned representations for the target task.

\subsubsection{Joint Training (JT)}
Joint training differs from the previous strategy of pre-training and fine-tuning by combining both stages into a single training process. In this approach, the model learns representations for both the pretext and downstream tasks simultaneously \cite{akkas2022jgcl}.
This strategy employs a combined loss function that fuses losses from these two tasks, usually balanced by a hyperparameter. This joint training is represented as a form of multi-task learning, where learning the pretext task aids in regularizing the learning of the downstream task. It is particularly useful to enhance generalization.

\subsubsection{Unsupervised Representation Training (URT)}
Unsupervised representation learning shares similarities with the pre-training and fine-tuning approach during its initial phase, in which the encoder is trained using objectives derived from SSL \cite{zhu2023unsupervised}. However, a key distinction in this strategy is that the encoder's parameters are locked in the post-training, and no additional adjustments are made during the application to downstream tasks. Training for both stages occurs using the same dataset, presenting a challenge for the encoder to effectively transition from purely unsupervised pre-training to performing supervised tasks. This method highlights the importance of robust SSL signals in building effective representations without relying on explicit, task-specific supervised training.

\begin{table}[H]
    \centering
    \caption{Comparison overview of the SSL training strategies for graph data.}
    \resizebox{\textwidth}{!}{
    \begin{tabular}{lllll}
        \hline
        \begin{tabular}[c]{@{}l@{}} \textbf{Aspect}  \end{tabular} & \begin{tabular}[c]{@{}l@{}} \textbf{Primary Goal}\end{tabular} & \begin{tabular}[c]{@{}l@{}}  \textbf{Training Phases} \end{tabular}  & \begin{tabular}[c]{@{}l@{}} \textbf{Key Benefits}  \end{tabular}& \begin{tabular}[c]{@{}l@{}} \textbf{Challenges}  \end{tabular}   \\ \hline
        
         \begin{tabular}[c]{@{}l@{}}Pre-training\\ with Fine-tuning \end{tabular} & \begin{tabular}[c]{@{}l@{}}  Establish strong initial\\ parameters through pre-\\training and then refine\\ these for specific downstream\\ tasks.  \end{tabular} & \begin{tabular}[c]{@{}l@{}}   Two distinct phases:\\ pre-training on unlabeled \\data followed by fine-\\tuning on labeled data.\end{tabular} & \begin{tabular}[c]{@{}l@{}}   	Allows for robust \\initial learning, \\reduces \\overfitting during \\fine-tuning.\end{tabular}& \begin{tabular}[c]{@{}l@{}}  Requires careful \\management of learning\\ rates and adaptation\\ during the fine-\\tuning phase. \end{tabular}  \\ \hline
         
          \begin{tabular}[c]{@{}l@{}}Joint Training\end{tabular} & \begin{tabular}[c]{@{}l@{}}  Simultaneously optimize for \\pretext and downstream \\tasks, enhancing generalization \\across tasks. \end{tabular} & \begin{tabular}[c]{@{}l@{}}  Single phase:\\ integrated phase where\\ pretext and downstream \\tasks are learned \\together. \end{tabular} & \begin{tabular}[c]{@{}l@{}}  Prevents overfitting\\ and improves model \\robustness by \\integrating learning \\tasks,  handles complex \\multitask scenarios. \end{tabular} & \begin{tabular}[c]{@{}l@{}}  Balancing the losses\\ from pretext and \\downstream tasks \\can be challenging \end{tabular} \\\hline
         
         \begin{tabular}[c]{@{}l@{}}  Unsupervised \\Representation\\ Training \end{tabular} & \begin{tabular}[c]{@{}l@{}} Build robust representations \\using SSL objectives \\without further adjustments\\ during downstream tasks.  \end{tabular} & \begin{tabular}[c]{@{}l@{}}  Two phases: \\ similar to pre-training\\ and fine-tuning, but\\ with locked parameters\\ for post pre-training. \end{tabular}  & \begin{tabular}[c]{@{}l@{}}   Minimizes the risk \\of model drift by \\using fixed \\representations. \end{tabular} & \begin{tabular}[c]{@{}l@{}}  The effectiveness \\ depends on the quality \\of the pre-training,\\ as no adjustments are \\made afterwards.  \end{tabular} \\\hline

    \end{tabular}}
    
    \label{tab:SSLstrategies}
\end{table}
\section{Self-supervised Learning in Graphs for Healthcare Applications}
\label{sec4}
SSL in graph-based applications is emerging as a novel approach in the healthcare field. This methodology is particularly well-suited to the complex healthcare data, which often involves complex networks of patient symptoms, disease correlations, and genetic data. SSL eliminates the requirement for significantly labeled datasets, a common constraint in medical data processing, through self-generating supervisory signals within the data. Its applications are very wide and range from enhancing disease prediction and diagnosis through pattern recognition in patient records to accelerating drug discovery by analyzing molecular structures and interactions. Incorporating SSL in graph-based healthcare applications holds significant promise for more insightful, accurate, and efficient medical analysis and patient care.

The following section explores various graph-based SSL approaches proposed for different healthcare applications. These include advanced predictive modeling and clinical insights, enhanced medical imaging and biomarker detection, and breakthroughs in drug discovery and molecular interaction analysis.

\newpage
\subsection{Advanced Predictive Modeling and Clinical Insights}

With the rapidly evolving healthcare landscape, advanced predictive modeling and clinical insights are presented as significant tools for forecasting outcomes and enhancing patient care. Leveraging the power of GNN and SSL, this section reviews methodologies that have improved the accuracy and efficacy of these predictive models. The integration of GNNs allows for a deeper understanding of complex and interconnected data structures inherent in patient records, while SSL facilitates the extraction of valuable insights from this data without the reliance on extensive labeled datasets. These technologies enable more precise clinical insights. 

Different approaches have been proposed, leveraging complex computational models to enhance the accuracy and efficiency of medical diagnoses and predictions. From enhancing predictive models that integrate electronic health records to innovating EEG analysis for seizure detection, each approach presents an effective methodology to handle specific challenges inherent to medical datasets \cite{xu2021predictive,tang2021self,yao2022self,ho2023self,lu2021self,xu2023seqcare,ruan2023msgcl}. These works demonstrate a commitment to improving model interpretability and robustness—essential qualities for gaining trust and facilitating adoption in clinical environments. Tang \textit{et al.} \cite{tang2021self} developed an approach for improved electroencephalographic (EEG) seizure analysis, including quantitative interpretability to measure and localize them. Works in \cite{lu2021self, yao2024self} present novel self-supervised graph learning frameworks for temporal health event prediction using electronic health records. The first work designed a network to detect disease complications effectively with enhanced personalized interpretability through a multilevel attention mechanism. On the other hand, Yao \textit{et al.} \cite{yao2024self} used the Kernel Subspace Augmentation to embed nodes into two different geometrically manifold views and then teach the model by maximizing mutual information between nodes and their neighbors on these views.  
In \cite{xie2024predicting}, Xie \textit{et al.} proposed a graph-based approach for predicting disease-gene associations, a critical task for understanding disease mechanisms and developing treatments. Their approach, named MiGCN (Self-Supervised Mutual Infomax Graph Convolution Network), utilizes an SSL strategy guided by external gene-gene and disease-disease collaborative graphs. The model eliminates noise within these graphs by maximizing mutual information between nodes and neighbors through a graphical mutual infomax layer.  

The studies in \cite{wen2023graph,sehanobish2021gaining,yu2024sparse,lu2024soft} successfully employed graph-based SSL methods to significantly enhance diagnostic accuracy and foster a deeper comprehension of complex diseases, including neurological disorders and COVID-19.  Wen \textit{et al.} \cite{wen2023graph} have developed an SSL framework tailored for brain network analysis. This method facilitates precise diagnostics in neurological disorders, such as autism spectrum disorder and bipolar disorder, by employing advanced graph representations for classification tasks. Moreover, Sehanobish \textit{et al.}'s \cite{sehanobish2021gaining} demonstrate that 
 integrating a transformer with a GAT efficiently processes complex datasets related to SARS-CoV-2 and enhances understanding of the virus's impact by capturing extensive contextual and local interactions within data. Table \ref{table:literature_summary1} provides an overview of recent research articles for advanced predictive modeling and Clinical Insights. 

The integration of SSL and GNNs in predictive modeling demonstrates the many strengths. This approach is adaptable to a wide range of healthcare tasks, from EHR analysis to molecular prediction. In addition, the graph SSL maximizes the use of unlabeled data and reduces the dependency on costly annotations. Many methods from our review have improved the models' interpretability, addressing critical requirements for clinical adoption.

\textcolor{black}{However, several challenges remain. Scalability is a major concern, as these approaches often require substantial computational resources to handle large graphs and datasets. Besides, data quality poses a critical challenge. Many methods depend on high-quality annotated datasets, and the lack of such data in real-world scenarios can restrict their performance and generalizability. Furthermore, the integration of multimodal data remains an open research area. Combining diverse data types, such as EHRs, medical imaging, and molecular data, presents a complex step toward creating comprehensive and robust healthcare solutions. }

\begin{table}[H]
\centering
\caption{Literature summary of graph-based SSL for advanced predictive modeling and Clinical Insights.}
\label{table:literature_summary1}
\resizebox{\textwidth}{!}{%
\begin{tabular}{@{}lcccccccccccccc@{}}
\toprule
\textbf{Reference} & \begin{tabular}[c]{@{}c@{}}

\textbf{Data} \end{tabular} & \begin{tabular}[c]{@{}c@{}}\textbf{Pretext}\\ \textbf{Task}\end{tabular} & \begin{tabular}[c]{@{}c@{}}\textbf{Target }\\ \textbf{Task} \end{tabular} & \multicolumn{3}{c}{\begin{tabular}[c]{@{}c@{}}\textbf{Learning }\\\textbf{Method}\end{tabular}} & \multicolumn{3}{c}{\begin{tabular}[c]{@{}c@{}}\textbf{Training}\\ \textbf{Method}\end{tabular}} & \multicolumn{5}{c}{\begin{tabular}[c]{@{}c@{}}\textbf{GNN }\\\textbf{Architecture}\end{tabular}} \\
\cmidrule(r){5-7} \cmidrule(lr){8-10} \cmidrule(lr){11-15}  
 & & & & \rotatebox[origin=c]{90}{\textbf{Contrastive}}  & \rotatebox[origin=c]{90}{\textbf{Predictive}} & \rotatebox[origin=c]{90}{\textbf{Generative}} & \rotatebox[origin=c]{90}{\textbf{PF}} &  \rotatebox[origin=c]{90}{JT} & \rotatebox[origin=c]{90}{\textbf{URT}}   & \rotatebox[origin=c]{90}{\textbf{GNN}} & \rotatebox[origin=c]{90}{\textbf{GCN}} & \rotatebox[origin=c]{90}{\textbf{GraphSage}} & \rotatebox[origin=c]{90}{\textbf{GAT}} & \rotatebox[origin=c]{90}{\textbf{GAE}}\\
\midrule

\begin{tabular}[c]{@{}c@{}} Yu \textit{et al.} \\ \cite{yu2024sparse} \end{tabular} & \begin{tabular}[c]{@{}c@{}}  HMDAD \\Disbiome \end{tabular}   & \begin{tabular}[c]{p{4cm}}   Learn informative representations using different data augmentation and multi-kernel similarity computation  \end{tabular}   & \begin{tabular}[c]{p{4cm}}  Prediction of microbe-disease associations\end{tabular} & \checkmark  &   &  &   & \checkmark &  &  & \checkmark &  & \\ \hline

\begin{tabular}[c]{@{}c@{}} Lu \textit{et al.} \\ \cite{lu2024soft} \end{tabular} & \begin{tabular}[c]{@{}c@{}} Brain\\ disease\\ datasets \end{tabular}   & \begin{tabular}[c]{p{4cm}} To learn more useful information using domain adaptive learning  \end{tabular}   & \begin{tabular}[c]{p{4cm}} Clustering brain functional connectivity data.  \end{tabular} &   & \checkmark  &  & \checkmark  &  &  & \checkmark &  &  & \\ \hline

\begin{tabular}[c]{@{}c@{}} Xu \textit{et al.} \\ \cite{xu2021predictive} \end{tabular} & \begin{tabular}[c]{@{}c@{}} Electronic\\ Health\\ Records \end{tabular}   & \begin{tabular}[c]{p{4cm}} Enhance the quality and completeness of the medical knowledge graph \end{tabular}   &\begin{tabular}[c]{p{4cm}} Predictive modeling of clinical events \end{tabular} &   & \checkmark  & \checkmark & \checkmark  &  &  &  & & \checkmark & \\ \hline

\begin{tabular}[c]{@{}c@{}} Tang \textit{et al.}\\ \cite{tang2021self} \end{tabular} & \begin{tabular}[c]{@{}c@{}} EEG\\ signals  \end{tabular} & \begin{tabular}[c]{p{4cm}} Predict the preprocessed EEG signals for the next time period \end{tabular} & \begin{tabular}[c]{p{4cm}} Seizure detection and classification  \end{tabular} & & \checkmark &  & \checkmark & &  &  & \checkmark & & & \\ \hline

\begin{tabular}[c]{@{}c@{}} Yao \textit{et al.} \\ \cite{yao2022self} \end{tabular} & \begin{tabular}[c]{@{}c@{}} Electronic\\health \\records \end{tabular} &   \begin{tabular}[c]{p{4cm}} Kernel subspace augmentation \end{tabular} & \begin{tabular}[c]{p{4cm}}  Medication outcome prediction  \end{tabular} & \checkmark  & &  &  & & \checkmark &\checkmark  & & & & \\ \midrule

\begin{tabular}[c]{@{}c@{}} Ho \textit{et al.} \\\cite{ho2023self} \end{tabular} & \begin{tabular}[c]{@{}c@{}} EEG\\ signals \end{tabular} & \begin{tabular}[c]{p{4cm}} Generate multiple types of EEG graphs to model normal brain activity patterns \end{tabular} & \begin{tabular}[c]{p{4cm}}   Detection and localization of seizure activities within the EEG data\end{tabular} & \checkmark   & & \checkmark &  & &  &  \checkmark & & & & \\ \hline

\begin{tabular}[c]{@{}c@{}}Lu \textit{\textit{et al.}} \\\cite{lu2021self}\end{tabular} & \begin{tabular}[c]{@{}c@{}} Electronic\\ health\\ records\end{tabular} & \begin{tabular}[c]{p{4cm}} Hierarchy enhanced historical  prediction \end{tabular} &\begin{tabular}[c]{p{4cm}} Temporal health event prediction \end{tabular} &  & & \checkmark &\checkmark  & &  & \checkmark &  & & & \\ \hline

\begin{tabular}[c]{@{}c@{}}  Xu \\\textit{et al.} \\\cite{xu2023seqcare}  \end{tabular}& \begin{tabular}[c]{@{}c@{}} Patient EHR data \end{tabular} & \begin{tabular}[c]{p{4cm}} Learn a bias-reduced feature space from the structure of the medical knowledge graph \end{tabular} &  \begin{tabular}[c]{p{4cm}} Diagnosis prediction  \end{tabular}  & \checkmark  & &  &  & \checkmark & &  & & &  & \\
\hline

\begin{tabular}[c]{@{}c@{}} Yao \textit{et al.} \\ \cite{yao2024self} \end{tabular} & \begin{tabular}[c]{@{}c@{}}  Electronic\\ health\\ records\end{tabular} & \begin{tabular}[c]{p{4cm}} Contrasting nodes and graph representations on two geometrically different manifold views  \end{tabular} &\begin{tabular}[c]{p{4cm}} Medication outcome prediction   \end{tabular} & \checkmark   & &  & \checkmark  & &  & \checkmark & & & & \\ \hline

\begin{tabular}[c]{@{}c@{}}  Xie \textit{et al.} \\ \cite{xie2024predicting} \end{tabular} & \begin{tabular}[c]{@{}c@{}} OMIM database \end{tabular} & \begin{tabular}[c]{p{4cm}} Maximize the mutual information between graph nodes and their neighbors in the \end{tabular} & \begin{tabular}[c]{p{4cm}} Predict disease-gene associations  \end{tabular} & \checkmark   & &  &\checkmark  & &  &  & \checkmark & & & \\ \hline

\begin{tabular}[c]{@{}c@{}} Wen \textit{\textit{et al.}}\\\cite{wen2023graph} \end{tabular} &\begin{tabular}[c]{@{}c@{}} Brain\\ network\\ data\end{tabular} &  \begin{tabular}[c]{p{4cm}} Mask graph autoencoder Signal representation learning \end{tabular}&\begin{tabular}[c]{p{4cm}} Brain network analysis\end{tabular} & \checkmark  & &  & \checkmark & & &  & \checkmark &   & & \\ \hline

\begin{tabular}[c]{@{}c@{}}  Sehanobish\\ \textit{et al. } \\\cite{sehanobish2021gaining} \end{tabular} & \begin{tabular}[c]{@{}c@{}} Bronchial \\epithelial\\ cell \end{tabular} &  \begin{tabular}[c]{p{4cm}} Learn node representations of the nodes local graph topology  \end{tabular} & \begin{tabular}[c]{p{4cm}} Predict the disease state \end{tabular} & \checkmark  & &  & \checkmark & & &  & & &\checkmark  & \\
\hline

\begin{tabular}[c]{@{}c@{}} Ruan \textit{et al. } \\\cite{ruan2023msgcl} \end{tabular} & \begin{tabular}[c]{@{}c@{}} miRNA \end{tabular} &  \begin{tabular}[c]{p{4cm}} Maximize the consistency between a label-free learner view and a known-association-based anchor view \end{tabular} & \begin{tabular}[c]{p{4cm}} Predict miRNA–disease  \end{tabular} & \checkmark  & &  & \checkmark & & &  & \checkmark & &  & \\
\hline

\begin{tabular}[c]{@{}c@{}} Jung \textit{et al. } \\ \cite{jung2024cancergate} \end{tabular} & \begin{tabular}[c]{@{}c@{}} 15 types of\\ cancer from \\the cancer \\genome atlas  \end{tabular} &  \begin{tabular}[c]{p{4cm}}  Capture meaningful representations  by reconstructing the input graph  \end{tabular} & \begin{tabular}[c]{p{4cm}} Predict cancer-driver genes  \end{tabular} &   & &  \checkmark &  & & \checkmark &  &  & &  & \checkmark \\
\hline

\end{tabular} }
\end{table}

\subsection{Improved Medical Imaging and Biomarker Detection} 
The evolution of AI has led to significant advancements in medical imaging and biomarker detection, transforming and enhancing diagnostic processes across healthcare systems. GNNs and SSL have revolutionized precision medicine, where complex imaging data and elusive biomarkers can be analyzed with remarkable accuracy and efficiency.
This section explores cutting-edge methodologies that make use of these technologies to refine the detection and interpretation of medical images and biomarkers. Integrating GNNs into medical imaging allows for a nuanced understanding of spatial and structural relationships within medical data. Similarly, SSL approaches facilitate extracting meaningful patterns from vast unlabeled datasets, enabling the identification of biomarkers without the extensive need for manual annotation. These developments accelerate the diagnostic process and enhance the predictive power of medical assessments, leading to more personalized patient care.

Different works have been proposed for fMRI image analysis to enhance learning and classification in scenarios with limited labeled data \cite{peng2022gate,wang2022contrastive,choi2024joint}. 
Peng \textit{et al.} \cite{peng2022gate} explored different new graph augmentation strategies that address the dynamic functional connectivities in fMRI data. The proposed model follows a two-step learning process for GCNs. 

In addition, various methods based on graph-SSL have been proposed for the diagnosis and management of specific medical conditions, such as fundus diseases and COVID-19. In \cite{ibrahim2022multi}, Ibrahim \textit{et al.} designed an SSL-based approach for breast cancer detection. They transformed the mammogram images into highly correlated multi-graphs to get rich structural relations and high-level texture features. Sun \textit{et al.} \cite{sun2021context} propose a novel approach, particularly for COVID-19 imaging datasets. The proposed approach uses a two-level learning objective: one at the regional anatomical level and another at the patient level. This method allows the handling of arbitrarily sized images in full resolution, which is a significant advantage over traditional methods that require fixed-size inputs. Ozen \textit{et al.} \cite{ozen2021self} address the challenges of region of interest (ROIs) retrieval in large and complex histopathology images. They follow the contrastive learning method to enable the effective handling of arbitrarily shaped ROIs by representing them as graphs and using spatial proximity for graph construction. Lin \textit{et al.} \cite{lin2021multi} present two novel multi-label classification networks, named MCG-Net and MCGS-Net, for an accurate diagnosis of fundus diseases. Fundus image classification is challenging due to its multi-label nature, as one image can contain multiple diseases. The MCG-Net is based on a GCN that captures relevant information from multi-label fundus images. 

Besides, different works based on graph SSL have been proposed to detect Parkinson's disease \cite{guo2022self,endo2022gaitforemer,guo2022tree,guo2022contrastive}. This use has enhanced spatial-temporal representations by capturing the nuances of human movement over time, which is essential for detecting subtle motor symptoms associated with Parkinson's disease.

For different medical tasks, including segmentation, classification, and object detection, MH Nguyen \textit{et al.} \cite{mh2024lvm} proposed an SSL approach that uses a second-order graph-matching formulation. This method captures images' global and local structural features by integrating advanced pair-wise image similarity metrics and constructing a combinatorial graph-matching objective. 

In \cite{wang2021hierarchical}, Wang \textit{et al.} address the challenge of conventional approaches for prostate cancer identification due to the complex nature of tissue phenotypes. Based on a DL framework using hierarchical graph-based representations, this approach explores multi-scale topological structures of whole slide images in an integrative context and improves progression prediction accuracy for prostate cancer patients.

These reviewed studies highlight the significance of SSL as an innovative approach for enhancing data analytics in various medical fields, as illustrated in Table \ref{table:literature_summary2}. The transition from conventional supervised learning to SSL is important due to the privacy concerns, high costs, and effort required for labeling large datasets. SSL's ability to utilize the abundant unlabeled data, especially images in healthcare, allows for training models that achieve high accuracy without extensive labeled datasets. Specific reviews on medical imaging demonstrate SSL's effectiveness in overcoming data annotation challenges, underscoring its potential to transform healthcare diagnostics and treatment strategies.

\begin{table}[H]
\centering
\caption{Literature summary of graph-based SSL approaches for improved medical imaging and biomarker detection. }
\label{table:literature_summary2}
\resizebox{\textwidth}{!}{
\begin{tabular}{@{}lcccccccccccccc@{}} 
\toprule
\textbf{Reference} & \begin{tabular}[c]{@{}l@{}} \textbf{Data}\end{tabular} & \begin{tabular}[c]{@{}l@{}}\textbf{Pretext}\\ \textbf{Task}\end{tabular} & \begin{tabular}[c]{@{}l@{}}\textbf{Target}\\ \textbf{Task}\end{tabular} & \multicolumn{3}{c}{\begin{tabular}[c]{@{}l@{}}\textbf{Learning }\\\textbf{Method}\end{tabular}} & \multicolumn{3}{c}{\begin{tabular}[c]{@{}l@{}}\textbf{Training}\\ \textbf{Method}\end{tabular}} & \multicolumn{5}{c}{\begin{tabular}[c]{@{}l@{}}\textbf{GNN }\\\textbf{Architecture}\end{tabular}} \\
\cmidrule(r){5-7} \cmidrule(lr){8-10} \cmidrule(lr){11-15}
& & & & \rotatebox[origin=c]{90}{\textbf{Contrastive}} & \rotatebox[origin=c]{90}{\textbf{Predictive}} & \rotatebox[origin=c]{90}{\textbf{Generative}} & \rotatebox[origin=c]{90}{\textbf{PF}} & \rotatebox[origin=c]{90}{\textbf{JT}} & \rotatebox[origin=c]{90}{\textbf{URT}} & \rotatebox[origin=c]{90}{\textbf{GNN}} & \rotatebox[origin=c]{90}{\textbf{GCN}} & \rotatebox[origin=c]{90}{\textbf{GraphSage}} & \rotatebox[origin=c]{90}{\textbf{GAT}} & \rotatebox[origin=c]{90}{\textbf{GAE}} \\
\midrule

\begin{tabular}[c]{@{}l@{}}Peng \textit{et al.}\\ \cite{peng2022gate}\end{tabular} & \begin{tabular}[c]{p{2cm}}fMRI images\end{tabular} & \begin{tabular}[c]{p{4cm}}Learn embeddings from augmented views\end{tabular} & \begin{tabular}[c]{p{4cm}} Disease classification\end{tabular} & \checkmark & & & \checkmark & & &\checkmark & & & & \\
\hline

\begin{tabular}[c]{@{}l@{}}Wang \textit{et al.}\\ \cite{wang2022contrastive}\end{tabular} & \begin{tabular}[c]{p{2cm}} fMRI images\end{tabular} & \begin{tabular}[c]{p{4cm}}Learn representations on functional connectivity graphs\end{tabular}&\begin{tabular}[c]{p{4cm}} Classify patients based on their functional connectivity patterns. \end{tabular}& \checkmark & & & \checkmark & & & & & & & \\
\bottomrule

\begin{tabular}[c]{@{}l@{}}Choi \textit{et al.}\\ \cite{choi2024joint}\end{tabular} &\begin{tabular}[c]{p{2cm}} fMRI data \end{tabular} & \begin{tabular}[c]{p{4cm}}Reconstruct dynamic graphs \end{tabular} & \begin{tabular}[c]{p{4cm}} Predicting phenotypes and psychiatric diagnoses\end{tabular}  & & & \checkmark & & \checkmark && & & & &\checkmark \\
\bottomrule

\begin{tabular}[c]{@{}l@{}}Ibrahim \textit{et al.}\\ \cite{ibrahim2022multi}\end{tabular} & \begin{tabular}[c]{@{}c@{}}Mammogram\\images\end{tabular} & \begin{tabular}[c]{p{4cm}}Learn robust node embeddings that capture the essential features and relationships within the mammogram segments\end{tabular} & \begin{tabular}[c]{p{4cm}}The classification of mammogram image segments\end{tabular} & \checkmark & & & \checkmark & & & & \checkmark & & & \\
\bottomrule

\begin{tabular}[c]{@{}l@{}}Sun \textit{et al.}\\ \cite{sun2021context}\end{tabular} & \begin{tabular}[c]{p{2cm}} CT images\end{tabular} & \begin{tabular}[c]{p{4cm}}Learn contextual relationships between different regions of the anatomy\end{tabular} & \begin{tabular}[c]{p{4cm}} Disease detection from medical imaging\end{tabular} & \checkmark & & & & & & & \checkmark & & & \\
\hline

\begin{tabular}[c]{@{}l@{}}Ozen \textit{et al.}\\ \cite{ozen2021self}\end{tabular} & \begin{tabular}[c]{p{2cm}}Whole Slide Images\end{tabular} & \begin{tabular}[c]{p{4cm}} Learning from the structure and content of the data\end{tabular} & \begin{tabular}[c]{p{4cm}} Region of interest representation learning\end{tabular} & \checkmark & & & & & \checkmark & & \checkmark & & & \\
\hline

\begin{tabular}[c]{@{}l@{}}Lin \textit{et al.}\\ \cite{lin2021multi}\end{tabular} & \begin{tabular}[c]{p{2cm}} Fundus images\end{tabular} & \begin{tabular}[c]{p{4cm}} Predict the type of geometric transformation\end{tabular} & \begin{tabular}[c]{p{4cm}}Fundus diseases diagnosis\end{tabular} & \checkmark & & & \checkmark & & & & \checkmark & & & \\
\hline

\begin{tabular}[c]{@{}l@{}}Guo \textit{et al.}\\ \cite{guo2022self}\end{tabular} & \begin{tabular}[c]{p{2cm}} Human Skeleton Sequences\end{tabular} & \begin{tabular}[c]{p{4cm}}Learn robust feature representation from spatial and temporal knowledge\end{tabular} &  \begin{tabular}[c]{p{4cm}} Parkinson disease motor symptoms assessment\end{tabular} & \checkmark & & & \checkmark & & & \checkmark & & & & \\
\hline

\begin{tabular}[c]{@{}l@{}}Guo \textit{et al.}\\ \cite{guo2022tree}\end{tabular} & \begin{tabular}[c]{p{2cm}} Video dataset of PD patients performing hand movements  \end{tabular} &\begin{tabular}[c]{p{4cm}} Distinguish between different augmented versions of the same hand movement sequences \end{tabular}&\begin{tabular}[c]{p{4cm}} Classify hand movement sequences  \end{tabular} & \checkmark & & \checkmark& & & & \checkmark& & & & \\
\bottomrule

\begin{tabular}[c]{@{}l@{}}Guo \textit{et al.}\\ \cite{guo2022contrastive}\end{tabular} &\begin{tabular}[c]{p{2cm}} Clinical video dataset \end{tabular} & \begin{tabular}[c]{p{4cm}} Learning representations by clustering features of each class \end{tabular} &\begin{tabular}[c]{p{4cm}} Classify the severity of bradykinesia \end{tabular}  &\checkmark & & & & \checkmark& & & \checkmark& & & \\
\bottomrule

\begin{tabular}[c]{@{}l@{}}MH Nguyen \\ \textit{et al.}\cite{mh2024lvm}\end{tabular} & \begin{tabular}[c]{p{2cm}} Medical imaging data (CT scans, MRI, X-rays, and ultrasound images)\end{tabular} & Graph matching & \begin{tabular}[c]{p{4cm}} Image segmentation and classification, object detection, domain generalization\end{tabular} & \checkmark & & & \checkmark & & & & \checkmark & & & \\
\hline

\begin{tabular}[c]{@{}l@{}} Wang\\ \textit{et al.}\cite{wang2021hierarchical}\end{tabular} &\begin{tabular}[c]{p{2cm}} UCLA prostate biopsy, Cedars-Sinai, and TCGA-PRAD dataset\end{tabular} & \begin{tabular}[c]{p{4cm}} Learn useful feature representations from the cell graphs
\end{tabular}  &  \begin{tabular}[c]{p{4cm}} Accurate prediction of prostate disease \end{tabular}  & \checkmark & & & \checkmark & & & & & & \checkmark & \\
\bottomrule

\end{tabular}}
\end{table}

\begin{table}[H]
\centering
\caption{Literature summary of graph-based SSL approaches for improved medical imaging and biomarker detection (continued)}
\resizebox{\textwidth}{!}{
\begin{tabular}{@{}lcccccccccccccc@{}} 
\toprule
\textbf{Reference} &  \textbf{Data} & \begin{tabular}[c]{@{}l@{}}\textbf{Pretext}\\ \textbf{Task}\end{tabular} & \begin{tabular}[c]{@{}l@{}}\textbf{Target}\\ \textbf{Task}\end{tabular} & \multicolumn{3}{c}{\begin{tabular}[c]{@{}l@{}}\textbf{Learning }\\\textbf{Method}\end{tabular}} & \multicolumn{3}{c}{\begin{tabular}[c]{@{}l@{}}\textbf{Training}\\ \textbf{Method}\end{tabular}} & \multicolumn{5}{c}{\begin{tabular}[c]{@{}l@{}}\textbf{GNN }\\\textbf{Architecture}\end{tabular}} \\
\cmidrule(r){5-7} \cmidrule(lr){8-10} \cmidrule(lr){11-15}
& & & & \rotatebox[origin=c]{90}{\textbf{Contrastive}} & \rotatebox[origin=c]{90}{\textbf{Predictive}} & \rotatebox[origin=c]{90}{\textbf{Generative}} & \rotatebox[origin=c]{90}{\textbf{PF}} & \rotatebox[origin=c]{90}{\textbf{JT}} & \rotatebox[origin=c]{90}{\textbf{URT}} & \rotatebox[origin=c]{90}{\textbf{GNN}} & \rotatebox[origin=c]{90}{\textbf{GCN}} & \rotatebox[origin=c]{90}{\textbf{GraphSage}} & \rotatebox[origin=c]{90}{\textbf{GAT}} & \rotatebox[origin=c]{90}{\textbf{GAE}} \\
\midrule

 \begin{tabular}[c]{@{}l@{}} Nguyen \textit{et al.}\\ \cite{nguyen2023companion} \end{tabular} &\begin{tabular}[c]{p{2cm}} 1.3 million medical images \end{tabular} & \begin{tabular}[c]{p{4cm}} Combines local and global image similarity metrics with structural constraints 
\end{tabular}  &  \begin{tabular}[c]{p{4cm}}  Segmentation, classification, and object detection \end{tabular}  & \checkmark & & & \checkmark & & & \checkmark & & & & \\
\bottomrule

\begin{tabular}[c]{@{}l@{}} Aryal \textit{et al. } \\\cite{aryal2024context} \end{tabular} & \begin{tabular}[c]{p{2cm}} PANDA dataset \end{tabular} &  \begin{tabular}[c]{p{4cm}} learn from relationships within the graph  \end{tabular} & \begin{tabular}[c]{p{4cm}} Cancer grading and diagnosis  \end{tabular} & \checkmark  & &  & \checkmark & & &  & \checkmark & &  & \\
\hline

\end{tabular}}
\end{table}

\subsection{Innovations in Drug Discovery and Molecular Interaction Networks}

With the rapidly evolving field of pharmaceuticals, searching for novel and efficient treatments is always necessary. Conventional drug discovery methods are often slow and have high costs and low success rates. With the advent of GNNs and SSL, computational approaches are setting new paradigms in identifying and developing therapeutic agents. This section reviews recent advancements that leverage these technologies to revolutionize drug discovery and analyze complex molecular interaction networks. 

Integrating GNNs with SSL has expedited the drug discovery process and enhanced the accuracy and efficacy of the drugs identified. For example, Zhang \textit{et al.} \cite{zhang2023antiviraldl} developed an innovative framework utilizing GNNs and SSL for accelerating antiviral drug development. They constructed a comprehensive virus-drug association dataset by integrating the Drugvirus2 database with FDA-approved antiviral drugs. Based on this data, a bipartite graph is created. To address the challenges of data sparsity, they employed contrastive learning techniques by training a Light Graph Convolutional Network (LightGCN). This approach improved the quality of the embeddings, which were important for predicting virus-drug associations by calculating the inner product between virus and drug embeddings. This method enhances prediction accuracy and streamlines the drug development process.

Building on leveraging SSL with GNNs, Rong \textit{et al.} \cite{rong2020self} develop a self-supervised graph transformer framework, GROVER, for large-scale molecular data analysis. GROVER addresses significant challenges in the field, including the scarcity of labeled molecular data and the poor generalization capabilities of models to new molecules. By integrating message-passing networks with Transformer architecture, GROVER provides a more expressive model that can efficiently learn from large-scale unlabeled molecular datasets.

Similarly, Zhao \textit{et al.} \cite{zhao2021csgnn} introduces the CSGNN, which incorporates a mix-hop neighborhood aggregator and a contrastive SSL task to capture high-order dependencies in molecular interaction networks.

Recognizing the importance of evaluating the quality of molecular graph embeddings, Wang \textit{et al.} \cite{wang2024evaluating} introduce the \textit{Molecular Graph Representation Evaluation} (MOLGRAPHEVAL), a comprehensive evaluation framework for assessing the quality of molecular graph embeddings learned through Graph-based SSL. MOLGRAPHEVAL is designed to benchmark Graph SSL methods against a suite of probing tasks categorized into three groups: generic graph properties, molecular substructures, and embedding space properties.

Further advancing molecular representation learning, Zang \textit{et al.} \cite{zang2023hierarchical} proposed a novel pretraining framework called Hierarchical Molecular Graph SSL (HiMol) for an enhanced prediction. The framework includes two main components: a Hierarchical Molecular GNN (HMGNN) and a Multi-level Self-supervised Pre-training (MSP). The HMGNN encodes motif structures and extracts hierarchical molecular representations of node-motif-graph, while the MSP develops multi-level generative and predictive tasks as self-supervised signals for the model.

In line with these advancements, Li \textit{et al.} \cite{li2021effective} introduced a novel SSL framework, MPG, for learning expressive molecular representations and enhancing drug discovery efforts. MPG employs a self-supervised pretraining strategy at both the node and graph levels, utilizing 11 million unlabeled molecules. This pretraining enables MolGNet to capture valuable chemical insights and produce interpretable representations. 

These studies are summarised in Table \ref{table:literature_summary3}. From the table, we can conclude how the integration of GNNs with SSL in drug discovery transforms the development of therapeutic agents and makes the process more efficient and accurate than traditional approaches.

\begin{table}[h!]
\centering
\caption{Literature summary of graph-based SSL approaches for drug discovery and molecular interaction networks.}
\label{table:literature_summary3}
\resizebox{\textwidth}{!}{%
\begin{tabular}{@{}lcccccccccccccc@{}}
\toprule
\textbf{Reference} &  \textbf{Data} & \begin{tabular}[c]{@{}l@{}}\textbf{Pretext}\\ \textbf{Task}\end{tabular} & \begin{tabular}[c]{@{}l@{}}\textbf{Target}\\ \textbf{Task}\end{tabular} & \multicolumn{3}{c}{\begin{tabular}[c]{@{}l@{}}\textbf{Learning }\\\textbf{Method}\end{tabular}} & \multicolumn{3}{c}{\begin{tabular}[c]{@{}l@{}}\textbf{Training}\\ \textbf{Method}\end{tabular}} & \multicolumn{5}{c}{\begin{tabular}[c]{@{}l@{}}\textbf{GNN }\\\textbf{Architecture}\end{tabular}} \\
\cmidrule(r){5-7} \cmidrule(lr){8-10} \cmidrule(lr){11-15}
& & & & \rotatebox[origin=c]{90}{\textbf{Contrastive}} & \rotatebox[origin=c]{90}{\textbf{Predictive}} & \rotatebox[origin=c]{90}{\textbf{Generative}} & \rotatebox[origin=c]{90}{\textbf{PF}} & \rotatebox[origin=c]{90}{\textbf{JT}} & \rotatebox[origin=c]{90}{\textbf{URT}} & \rotatebox[origin=c]{90}{\textbf{GNN}} & \rotatebox[origin=c]{90}{\textbf{GCN}} & \rotatebox[origin=c]{90}{\textbf{GraphSage}} & \rotatebox[origin=c]{90}{\textbf{GAT}} & \rotatebox[origin=c]{90}{\textbf{GAE}} \\
\midrule

\begin{tabular}[c]{@{}l@{}}Zhang \textit{et al.}\\ \cite{zhang2023antiviraldl}\end{tabular} & \begin{tabular}[c]{p{2cm}}   DrugVirus.info2 Database and FDA-approved Virus-Drug Associations \end{tabular} & \begin{tabular}[c]{p{4cm}} 
  Minimize the distance between augmented views of the same node and maximize the distance between different nodes \end{tabular} & \begin{tabular}[c]{p{4cm}}  Predicting virus-drug associations  \end{tabular} & \checkmark & & & &\checkmark & & & \checkmark & & & \\
\bottomrule

\begin{tabular}[c]{@{}l@{}}Rong \textit{et al.} \\ \cite{rong2020self}\end{tabular} & \begin{tabular}[c]{p{2cm}} 10 million unlabeled molecules\end{tabular} & \begin{tabular}[c]{p{4cm}} Learn from the structural and semantic information of molecules\end{tabular} & \begin{tabular}[c]{p{4cm}}  Molecular property prediction\end{tabular} & & & & \checkmark & \checkmark & & \checkmark & & & & \\
\bottomrule

\begin{tabular}[c]{@{}l@{}}Zhao \textit{et al.} \\ \cite{zhao2021csgnn}\end{tabular} & \begin{tabular}[c]{p{2cm}} Molecular interaction network\end{tabular} & \begin{tabular}[c]{p{4cm}} Maximize mutual information between node-level and graph-level representations\end{tabular} & \begin{tabular}[c]{p{4cm}} Molecular interaction prediction\end{tabular} & \checkmark & & & & \checkmark & & \checkmark & & & & \\
\bottomrule

\begin{tabular}[c]{@{}l@{}}Wang \textit{et al.} \\ \cite{wang2024evaluating} \end{tabular} & \begin{tabular}[c]{p{2cm}} Molecular dataset\end{tabular} & \begin{tabular}[c]{p{4cm}} Predict proxy objectives\end{tabular} &  \begin{tabular}[c]{p{4cm}} Prediction of Biochemical properties of molecules\end{tabular} & \checkmark &\checkmark &\checkmark & \checkmark &  & &  \checkmark & \checkmark & \  & \checkmark & \  \\
\bottomrule

\begin{tabular}[c]{@{}l@{}}Zang \textit{et al.} \\ \cite{zang2023hierarchical} \end{tabular} & \begin{tabular}[c]{p{2cm}} 6 different molecular datasets \end{tabular} & \begin{tabular}[c]{p{4cm}}  Reconstruct parts of the molecular graph based on its structure; predict molecules features \end{tabular} & \begin{tabular}[c]{p{4cm}} Molecular property prediction \end{tabular} &  & \checkmark & \checkmark & \checkmark &  & & \checkmark  &  &   &  &   \\
\bottomrule

\begin{tabular}[c]{@{}l@{}}  Li \textit{et al.} \\ \cite{li2021effective} \end{tabular} & \begin{tabular}[c]{p{2cm}}  11 million unlabeled molecules \end{tabular} & \begin{tabular}[c]{p{4cm}}  Learn global representations of molecules and capture structure and properties \end{tabular} & \begin{tabular}[c]{p{4cm}} Predict how effectively a drug binds to its target. \end{tabular} &  &  \checkmark & & \checkmark &  & & \checkmark &  &   &  &   \\
\bottomrule

\begin{tabular}[c]{@{}l@{}}  Jin \textit{et al.} \\ \cite{jin2024sadr} \end{tabular} & \begin{tabular}[c]{p{2cm}}  DeepDR DTI CDataset \end{tabular} & \begin{tabular}[c]{p{4cm}} Learn feature representations of  drugs and diseases nodes trough data augmentation \end{tabular} & \begin{tabular}[c]{p{4cm}} Predict potential associations between drugs and diseases  \end{tabular} & \checkmark  &  & & \checkmark &  & &  & \checkmark &   &  &   \\
\bottomrule

 \begin{tabular}[c]{@{}l@{}} Wang \textit{et al.}\\ \cite{wang2024hierarchical} \end{tabular} &\begin{tabular}[c]{p{2cm}}   DrugBank\end{tabular} &\begin{tabular}[c]{p{4cm}} Learn low-dimensional representations for drugs and diseases using meta-path aggregation
\end{tabular}  &  \begin{tabular}[c]{p{4cm}}  Predict latent associations for drugs and diseases \end{tabular}  & \checkmark & & &  & & \checkmark &  & \checkmark & & & \\
\bottomrule

 \begin{tabular}[c]{@{}l@{}} Liu \textit{et al.}\\ \cite{liu2024developing} \end{tabular} &\begin{tabular}[c]{p{2cm}}  LncRNADisease and MNDR \end{tabular} & \begin{tabular}[c]{p{4cm}} Learning through masking parts of the lncRNA-disease network 
\end{tabular}  & \begin{tabular}[c]{p{4cm}}  Identification of potential drugs targeting lncRNA\end{tabular}  &  & & \checkmark & \checkmark & &  &  &  & & & \checkmark \\
\bottomrule

\end{tabular}}
\end{table}

\subsection{Discussion and Derived insights}

\begin{figure}[h]
    \centering
    \begin{subfigure}[b]{0.4\textwidth}
        \centering
        \includegraphics[width=\textwidth]{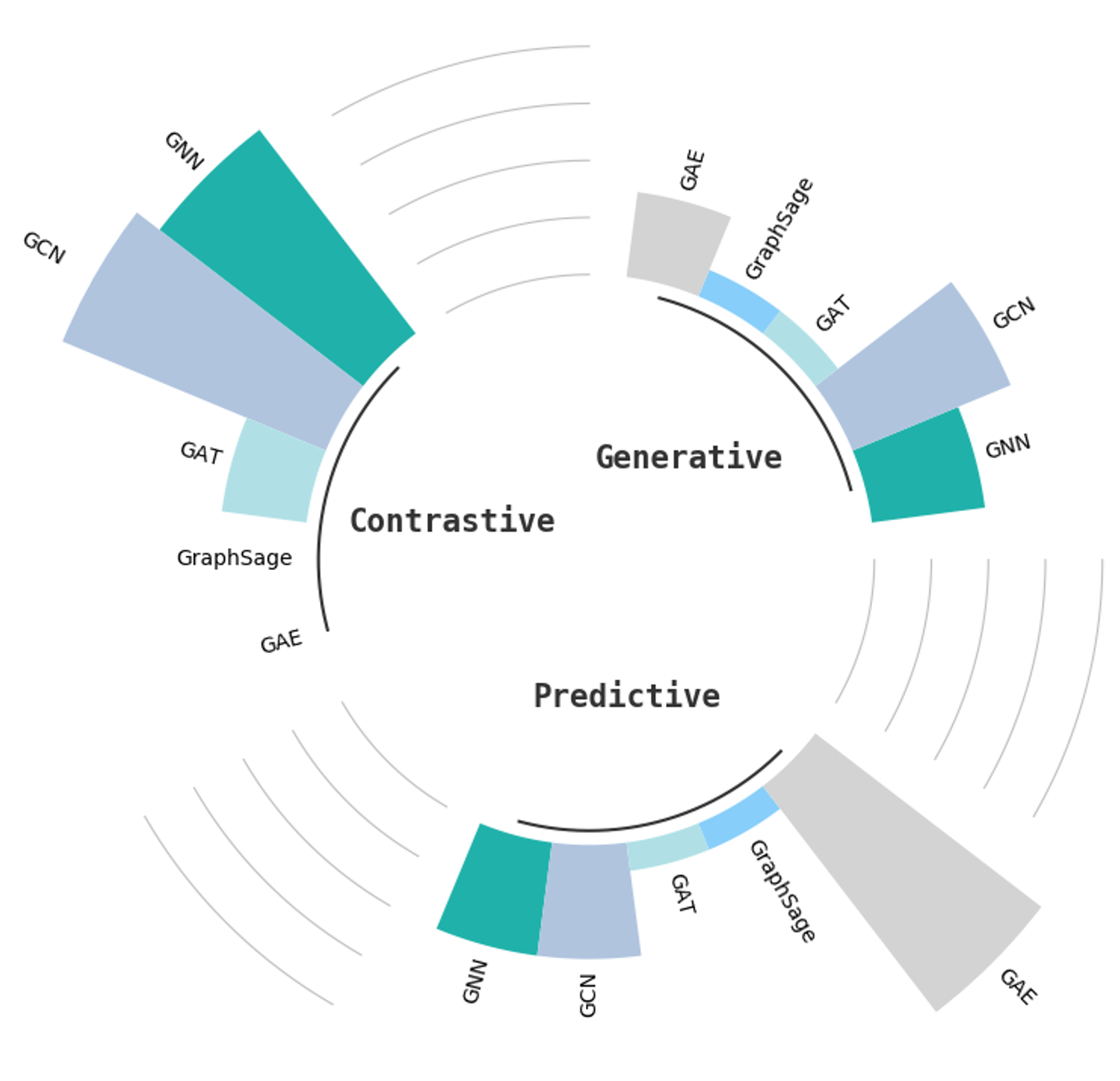}
        \caption{The prevalence of used network for various graph SSL approaches.}
        \label{fig:sub1}
    \end{subfigure}
    \hfill
    \begin{subfigure}[b]{0.4\textwidth}
        \centering
        \includegraphics[width=\textwidth]{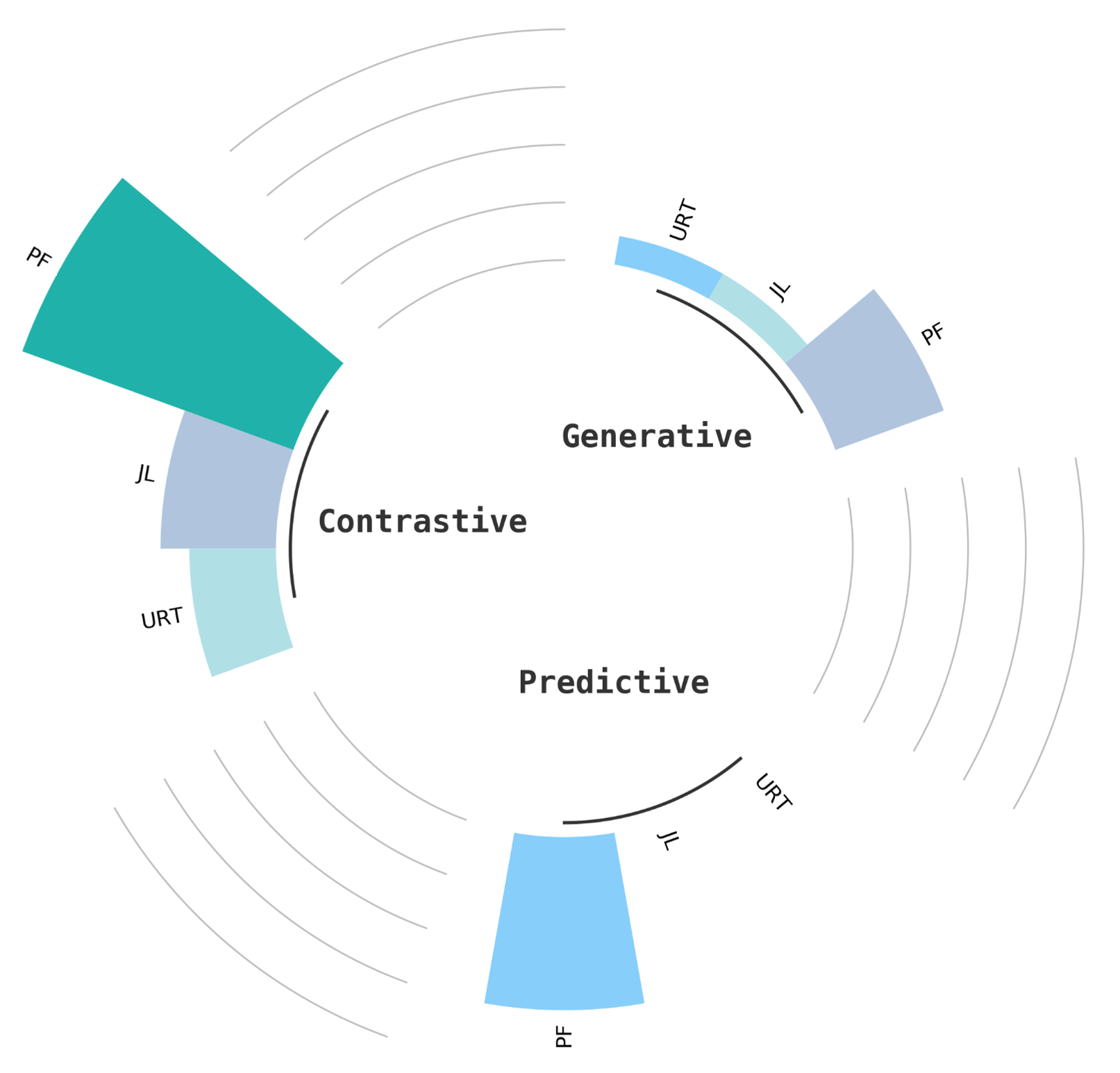}
        \caption{The prevalence of training strategy for various graph SSL approaches.}
        \label{fig:sub2}
    \end{subfigure}
    \caption{Comparative overview of the reviewed works categorized by the self-supervised learning method}
    \label{fig:comp}
\end{figure}

Figure \ref{fig:comp} provides a comparative overview of the reviewed works categorized by the SSL method. 
Subfigure (a) displays the different networks used in three SSL methods: contrastive, generative, and predictive. Subfigure (b) presents the various training strategies combined with the SSL methods.

It is evident that contrastive SSL is the most commonly used method. Its high adoption is due to its ability to effectively learn discriminative features by contrasting positive and negative pairs. The commonly used neural networks in contrastive SSL include simple GNN, GCN, and GAT, with GCN being the most prevalent due to its effectiveness in leveraging local neighborhood information. Additionally, the training of this contrastive SSL method is distributed across fine-tuning and pertaining, joint training, and unsupervised representation training.
\\
In contrast, the generative SSL approaches primarily use GCN, along with GNN and GAE. These methods focus on generating realistic graph data or reconstructing graphs to help understand the graph structure comprehensively. They often involve fine-tuning and pre-training the models before fine-tuning them for specific tasks. Predictive SSL methods often utilize URT, as generative models benefit from learning comprehensive data representations. The
Predictive SSL involves mainly GAE. These methods predict missing node attributes or future node states, aiding in node classification and link prediction tasks. Usually, this method follows the pretraining and fine-tuning training strategy.

From the detailed review of various studies using graph SSL for healthcare applications, we can draw several key conclusions about the benefits and advancements enabled by these applications.

\begin{itemize}
    \item SSL in graph-based applications is highly effective in environments where there is a shortage of labeled data, which is common in the healthcare industry due to the cost, privacy concerns, and difficulties in obtaining complete annotations. This technique creates its own supervisory signals from the data and allows for the efficient utilization of abundant unlabeled data.

    \item The applications of graph-based SSL for healthcare are vast and significant. They include improving disease diagnosis by recognizing patterns in patient records and accelerating drug discovery by analyzing molecular structures and interactions. This demonstrates the versatility of graph-based SSL in addressing various crucial healthcare tasks.

    \item Several studies, such as those employing RSML-GCN for drug repositioning or GATE for fMRI analysis, have demonstrated that SSL methods generally achieve excellent performance, surpassing traditional supervised learning models. They excel in handling complex tasks such as predicting drug-disease interactions and diagnosing diseases using medical imaging.
    
    \item SSL is particularly useful for processing the complex, hierarchical, and multi-modal data structures commonly found in healthcare, such as electronic health records, genetic information, and biomedical images. Techniques like GCNs are utilized to effectively model this data, thus improving the learning of intricate patterns and relationships.

    \item  The widespread use of SSL contrastive methods highlights their robustness and versatility in learning effective representations from graph data. While SSL generative methods are less common, they provide valuable insights into the graph generation and reconstruction process. SSL predictive methods bridge the gap between unsupervised and supervised learning by using pre-trained models for downstream tasks, thus improving performance on specific predictions.

    \item SSL frameworks such as GROVER \cite{rong2020self} demonstrate superior generalization capabilities compared to conventional methods, effectively adapting to new data. This is particularly important in healthcare, where emerging diseases and novel drug compounds frequently appear.

    \item Some SSL approaches incorporate mechanisms such as multi-level attention to provide insights into the model's decision-making process, enhancing interpretability and aiding in personalized medicine. This aspect is significant for clinical acceptance and customizing treatments to individual needs.

    \item The integration of advanced neural network architectures, such as combining Transformers with GNNs, provides powerful methods to address the specific challenges in analyzing healthcare data. These integrations enable the effective capture of local and global data dependencies.

    \item SSL methods offer scalability advantages by leveraging unlabeled data and enabling training on large-scale datasets, which is increasingly important due to the expanding volumes of data generated in healthcare.
    
\end{itemize}

\section{Datasets and Evaluation Metrics}
\label{sec5}
This section provides an overview of the datasets and evaluation metrics commonly used in research and applications GNNs and SSL in healthcare. The availability and quality of datasets play an essential role in the development and validation of models, while evaluation metrics confirm their effectiveness and reliability for real-world applications.

\subsection{Publicly Available Datasets}
A longstanding problem in ML for healthcare is the lack of public datasets, which thwarts the replicability of results. While most of the studies conduct their evaluations on private datasets only which prevents the comparability of existing works, we found a reasonable number of studies that reported performance on public datasets. 

Table \ref{tab:pubdatasets} provides an overview of the most commonly used datasets for supervised learning with graphs along with a download link, in an effort to promote the use of public benchmark data in future studies. Each dataset offers unique opportunities for researchers to explore various aspects of human health and disease. In the following, we categorize the datasets and provide a brief description of their features and relevance.
\newpage
\textbf{Clinical Data:}

\begin{itemize}
    \item  MIMIC-III: This dataset encompasses de-identified health information for over 112,000 critical care unit patients (2001-2012). It includes vital signs, medications, laboratory results, clinical notes, and more.
    \item MIMIC-IV: An expansion of MIMIC-III, containing de-identified electronic health records for 524,000 patients admitted between 2008 and 2019.
    \item  TCGA-PRAD: This standardized dataset offers comprehensive clinical data (biospecimens, DNA methylation, proteomics) for prostate adenocarcinoma patients. It provides a rich resource for cancer research.
\end{itemize}
\textbf{Imaging Data:}
\begin{itemize}
    \item REST-meta-MDD: This dataset comprises resting-state fMRI data from over 2,400 individuals, including patients with major depressive disorder and healthy controls.

    \item  CBIS-DDSM: This dataset contains 2,620 mammogram images with corresponding labels for regions of interest, allowing researchers to develop computer-aided diagnosis tools for breast cancer.
\end{itemize}
\textbf{Drug Discovery and Infectious Diseases:}

\begin{itemize}
    \item DrugVirus.info 2: This integrative portal facilitates exploration of antiviral drugs and their combinations. It includes information on 231 drugs and 153 viruses, aiding researchers in identifying potential treatments for emerging viral threats.

    \item  Cdataset: This extensive resource encompasses FDA-approved medications, drugs in clinical trials, and discontinued drugs. It includes drug similarity networks and disease similarity networks, valuable for drug repositioning research.
\end{itemize} 
\textbf{Genomics and Microbiome:}

\begin{itemize}
    \item ADHD200: This collaborative dataset provides structural and functional MRI data alongside phenotypic information from over 900 children and adolescents with ADHD and healthy controls.

    \item HMDAD: This database links microbes in the human body to various diseases. It currently contains information on 39 diseases and 292 microbes.

    \item Disbiome: This resource allows researchers to explore how the composition of microbes changes in various diseases.
\end{itemize}
\textbf{Molecular Data:}
\begin{itemize}

    \item Geom: This dataset features 37 million molecular conformations with energy and statistical weight annotations for over 450,000 molecules.

    \item DrugBank: This comprehensive resource combines detailed drug data with drug target and drug action information, valuable for in silico drug discovery and related research areas.
\end{itemize} 
\textbf{Genetic and miRNA-Disease Associations:}
\begin{itemize}
    \item OMIM: This freely available and regularly updated resource provides comprehensive information on human genes and genetic phenotypes.

    \item  HMDD: This database curates experimentally validated evidence for human microRNA (miRNA) and disease associations. It currently holds information on over 53,000 miRNA-disease associations.
 
\end{itemize}

\begin{table}[H]
\centering
\caption{Publicly Available Graph-Based Datasets.}
\begin{adjustbox}{max width=\textwidth}
\label{tab:pubdatasets}
\begin{tabular}{c c c c c}   
\hline
\centering \textbf{Name[citation]} &  \centering \textbf{Year} &  \centering \textbf{Number of cases} &  \centering \textbf{Field} &  \textbf{Download Link}  \\ \hline
MIMIC-III \cite{d3} & 2015 & 40,000 & Miscellaneous  & 
\href{https://physionet.org/content/mimiciii/1.4/}{physionet.org/content/mimiciii/}  \\ \hline
MIMIC-IV \cite{d1} & 2023 & 60,000 & Miscellaneous &  \href{https://mimic.mit.edu}{mimic.mit.edu}  \\ \hline
TCGA-PRAD \cite{d2} & 2012 & 500 & Prostate cancer & \href{https://portal.gdc.cancer.gov/projects/TCGA-PRAD}{gdc.cancer.gov/TCGA-PRAD}  \\ \hline
REST-meta-MDD \cite{d4} & 2017 & 2,428 & Major depressive disorder & \href{https://rfmri.org/REST-meta-MDD}{rfmri.org/REST-meta-MDD}  \\ \hline
CBIS-DDSM \cite{d5} & 2017 & 1,566 & Breast cancer & \href{https://www.kaggle.com/datasets/awsaf49/cbis-ddsm-breast-cancer-image-dataset}{cbis-ddsm-breast-cancer-dataset}  \\ \hline
DrugVirus.info 2 \cite{d6} & 2022 & - & Viral diseases & \href{https://drugvirus.info/}{drugvirus.info}  \\ \hline
TUdataset \cite{d7} & 2020 & - & Miscellaneous & \href{https://chrsmrrs.github.io/datasets/docs/datasets/}{chrsmrrs.github.io}  \\ \hline
ADHD200 \cite{d8} & 2016 & 947 & ADHD & \href{http://preprocessed-connectomes-project.org/adhd200/}{preprocessed-connectomes-project.org}  \\ \hline
HMDAD \cite{d9} & 2016 &  - & Miscellaneous & \href{http://www.cuilab.cn/hmdad}{www.cuilab.cn/hmdad}  \\ \hline
Disbiome \cite{d10} & 2018 & - & Miscellaneous & \href{https://disbiome.ugent.be/home}{disbiome.ugent.be/home}  \\ \hline
\begin{tabular}[c]{@{}l@{}} AD${_1}$ \& PTSD1${_1}$ \&\\ ADHD${_1}$ \& ASD${_1}$ \cite{d11} \end{tabular} & 2018 & 1,124 & Brain-based disorders & \href{https://github.com/xinyuzhao/identification-of-brain-based-disorders/tree/master/data}{github.com/xinyuzhao/} \\ \hline
\begin{tabular}[c]{@{}l@{}} AD${_2}$ \& PTSD1${_2}$ \& \\ ADHD${_2}$ \& ASD${_2}$ \cite{d12} \end{tabular} & 2019 & 2,137 & Brain-based disorders & \href{https://github.com/pradlanka/malini}{github.com/pradlanka/malini} \\ \hline
Cdataset \cite{d13} & 2014 & - & - & \href{https://github.com/bioinfomaticsCSU/MBiRW/tree/master/Datasets/CDatasets}{github.com/bioinfomaticsCSU} \\ \hline
GEOM \cite{d14} & 2022 & - &  - & \href{https://github.com/learningmatter-mit/geom}{github.com/learningmatter-mit/geom} \\ \hline
DrugBank \cite{d15} & 2006 & - &  - & \href{https://go.drugbank.com/}{go.drugbank.com} \\ \hline
OMIM \cite{d16} & 1994 & - & Genetic disorders & \href{https://www.ncbi.nlm.nih.gov/omim/}{www.ncbi.nlm.nih.gov/omim} \\ \hline
HMDD \cite{d17} & 2007 & - & miRNA related-diseases & \href{http://www.cuilab.cn/hmdd}{www.cuilab.cn/hmdd} \\ \hline
\end{tabular}
 \end{adjustbox}
\end{table}

\subsection{Proprietary and Confidential Datasets}
While public datasets provide valuable insights, private datasets offer unique opportunities to explore more specific and sensitive medical conditions. These datasets, though not publicly accessible, have been instrumental in advancing SSL techniques. Table \ref{tab:prdatasets} outlines a comprehensive list of all datasets that were applied. 

\begin{table}[H]
\centering
\caption{ Private Graph-Based Datasets (NGN: No Given Name).}
\label{tab:prdatasets}
\begin{adjustbox}{max width=\textwidth}
\begin{tabular}{c p{4.5in} c}
\hline
\centering \textbf{Name [ref]} &  \centering \textbf{Description} &  \textbf{Contact email} \\ \hline
UCLA \cite{di1} & \RaggedRight{A prostate biopsy dataset containing 20,229 slides from prostate needle biopsies of 830 patients pre- or post-diagnosis} & \url{jiayunli@g.ucla.edu}  \\ \hline
Cedars-Sinai \cite{di2} & \RaggedRight{A dataset consisting of 30 slides from prostatectomies of 30 patients, which were annotated with coarse contour annotations} & \url{Arkadiusz.Gerytch@cshs.org}  \\ \hline
NGN \cite{ozen2021self} & \RaggedRight{A breast histopathology dataset comprising 78 digital whole-slide images obtained by scanning tissue specimens from 63 patients which are categorized into four classes: benign, atypia, in situ carcinoma, and invasive carcinoma} & \url{saksoy@cs.bilkent.edu.tr}  \\ \hline
NGN \cite{nguyen2023companion} & \RaggedRight{A collection of 85,965 electronic medical records from 31 animal hospitals transformed to a knowledge graph} & \url{ojlee@catholic.ac.kr}  \\ \hline
US FDA \cite{zhang2023antiviraldl} & \RaggedRight{A manually collected FDA-approved drugs for the treatment of viral infectious diseases up until October 2022, constituting a dataset of 142 virus-drug associations between 111 drugs and 16 viruses} & \url{liguangdi@csu.edu.cn}  \\ \hline
GATH \cite{xu2023seqcare} & \RaggedRight{A real-world EHR dataset from a Grade A tertiary hospital in China including patient information from long-term follow-up over a period of 5 years} & \url{xuyx@stu.pku.edu.cn}  \\ \hline

\end{tabular}
 \end{adjustbox}
\end{table}

\subsection{Evaluation Metrics}
\label{sec6}

Graph SSL models are assessed using different metrics depending on the task. For graph classification, common metrics include ROC-AUC and accuracy, which evaluate the model's ability to classify graphs into different classes. On the other hand, for graph regression tasks, metrics like Mean Absolute Error (MAE) are used to gauge the model's accuracy in predicting continuous values associated with graphs.

In certain studies, these metrics are determined by performing k-fold cross-validation. This involves splitting the dataset into k folds, training the model on k-1 folds, and evaluating it on the remaining fold. The process is repeated k times, and the average performance across all folds is reported as the final result. For instance, Zhang \textit{et al.} \cite{zhang2023antiviraldl} adopted this approach with k equal to 5.

When evaluating multi-label diagnosis prediction models, previous studies \cite{xu2023seqcare} have used various performance measures, such as micro-averaged and macro-averaged areas under the ROC curve (AUROC) and areas under the precision-recall curve (AUPRC).

Beyond metrics like micro-AUROC and macro-AUPRC, other evaluation tools are employed for specific tasks. For prediction tasks, the Dice score \cite{dice} measures the overlap between predicted and actual outcomes. In segmentation tasks, the Concordance index (C-index) \cite{wang2021hierarchical} assesses how well the model ranks patients based on their predicted risk, especially when dealing with censored data.

To futher assess the developed models' capacity to cluster various classes, visual evaluation metrics were utilized. The research in \cite{li2021effective} employed the Uniform Manifold Approximation and Projection (UMAP), a dimensionality reduction technique, to project the high-dimensional representations extracted from the final layer of the pre-trained MolGNet into a 2D space. This enabled the visualization of both valid and invalid molecule representations, facilitating a qualitative evaluation of the model's clustering performance.

\section{Challenges and Limitations}
\label{sec7}
Creating effective graph SSL approaches for healthcare applications has several challenges and limitations. These difficulties may be divided into five primary categories: computational efficiency and scalability, generalizability of models, handling incomplete and irregular data, ethical and legal considerations, methodological innovations, and theoretical foundations. We will discuss these categories in more depth in the following sections.

\subsection{Heterogeneous Node Types and Relations}

A significant limitation of current graph-based SSL methods in healthcare lies in their assumption of homogeneous graph structures. This means that nodes and edges are often treated as identical entities, disregarding the inherent diversity and complexity of real-world healthcare data. For instance, in a medical knowledge graph, entities like genes, proteins, drugs, and diseases may have distinct properties and relationships.

When faced with such heterogeneous data, traditional graph neural networks may struggle to effectively capture the underlying patterns and dependencies. This can lead to suboptimal performance and limited generalization capabilities. As a result, the models may fail to accurately predict disease outcomes, identify drug targets, or recommend personalized treatment plans.

\subsection{Scalability and Computational Efficiency}

One of the biggest challenges related to the adoption of graph SSL in healthcare is related to scalability and computational efficiency.
Healthcare datasets can occasionally be found in extremely large and intricate forms, with millions of linked data points. 
Such large-scale graphs demand an enormous amount of computational power to process. The incremental learning strategies can be employed in this situation to update models dynamically as new data becomes available \cite{van2022three}. This approach will mitigate the need to reprocess the entire dataset, therefore enhancing computational efficiency.

\subsection{Generalizability of Models}

Generally, models learned from specific demographic data do not perform well on other groups, which leads to biased medical predictions and treatments.
Additionally, healthcare data vary significantly across different institutions and populations, making it difficult for models to generalize well across different settings.
To handle such a challenge, graph SSL models have to be tested across diverse datasets to ensure they generalize well. Besides, leveraging techniques like domain adaptation aids in developing more flexible models.

\subsection{Handling of Irregular, Noisy and Incomplete Data}
Missing data is common for healthcare data, where some records have missing or incomplete values. This restriction makes it more difficult to train reliable models.
Conversely, data collected in non-uniform time intervals, which is common in clinical settings, pose challenges for standard graph modeling techniques.
Incomplete or missing data can be handled using probabilistic models \cite{zhu2024review} and imputation techniques \cite{thomas2021systematic}. Graph-based SSL methods can be also sensitive to noise, which is common in healthcare datasets.
 
\subsection{Methodological Innovations}
Advanced methodological innovations are needed in graph SSL, especially related to pretext tasks for complex graphs and graph augmentation strategies.
A limited set of graph-specific pretext tasks addresses the complexities of different kinds of graphs like temporal dynamics and relational data \cite{ding2022data,zhao2022graph}.
Unlike image data, the augmentation techniques for graphs are limited. The basic graph augmentation techniques involve simple manipulations like adding or removing nodes and edges. There is a need for more sophisticated and empirically validated augmentation strategies that align with the complex, non-Euclidean nature of graph data. 

\subsection{Ethical and Legal Issues}
Data privacy is one of the foremost challenges in adopting graph-based SSL in healthcare. Developing such models often requires access to detailed patient data, which raises concerns about privacy and data protection.
Employing recent technologies like differential privacy \cite{ficek2021differential} or federated learning \cite{atitallah2023fedmicro} to protect patient data is presented as a promising solution.

Moreover, there is a pressing demand for high levels of model interpretability to understand the output in healthcare. This can be challenging with complex graph-based models that are often black boxes.
Developing methods for model explainability and ensuring that outputs are interpretable to practitioners \cite{stiglic2020interpretability} are in need.

\section{Future Directions and Trends}
\label{sec8}
SSL is advancing rapidly in the healthcare industry. In this section, we highlight a number of significant issues that might reflect future developments and trends in this study area.

\subsection{Potential of Self-supervised Learning in Emerging Healthcare Domains}

SSL is poised to revolutionize emerging healthcare fields, such as personalized medicine, continuous health tracking, and digital health aids. By harnessing large volumes of unlabeled health data, SSL can help create more precise and customized patient care choices, enhance disease progression prediction models, and reinforce remote health monitoring technologies.

\subsection{Integration with Other AI Techniques}

Combining SSL with additional AI methods such as reinforcement learning, transfer learning, and federated learning can significantly enhance the effectiveness and performance of healthcare models. For example, integrating SSL with reinforcement learning can improve treatment protocols in real time. Likewise, pairing SSL with federated learning allows for the development of models that uphold patient confidentiality while learning from diverse health data sources. 

A promising avenue for future research lies in the integration of large language models (LLMs) with graph-based SSL methods. LLMs have emerged as a powerful tool, demonstrating exceptional performance in the healthcare domain \cite{cg1,cg2,cg3}. With their ability to process and generate human-quality text, they can enhance the interpretability and clinical utility of graph-based models. By leveraging the strengths of both paradigms, more powerful and insightful healthcare solutions can be developed. For instance, LLMs can be employed to generate comprehensive and understandable explanations for model predictions, facilitating trust and adoption among healthcare professionals. Additionally, LLMs can assist in knowledge discovery and hypothesis generation by analyzing large-scale medical literature and identifying novel relationships between entities. By combining the structural understanding of graph-based models with the semantic capabilities of LLMs, new insights can be unlocked to accelerate medical research.

\subsection{Addressing the Challenges of Big Data in Healthcare}

As healthcare data volumes continue to rapidly expand, SSL techniques need to adapt to effectively manage and analyze this big data. This challenge involves handling large datasets and extracting valuable insights. Future studies should prioritize the development of scalable SSL algorithms capable of processing and identifying significant patterns from extensive and complex datasets, potentially in real time.

\subsection{Integration of Multimodal Data}

Leveraging a self-supervised framework to integrate diverse multimodal data sources holds substantial promise in advancing diagnostic models within the healthcare domain. By fusing graph data with imaging, textual, and tabular data, we can create more comprehensive and nuanced representations, ultimately leading to improved accuracy and efficacy of these models in clinical decision-making and patient care.

A potential approach is to utilize a self-supervised learning framework to pre-train a multimodal model on a large corpus of unlabeled medical data. This pre-training phase can help the model learn to extract meaningful representations from diverse data modalities, such as medical images, clinical notes, and genomic data. Subsequently, the pre-trained model can be fine-tuned on specific downstream tasks, such as disease diagnosis or drug discovery.

By leveraging self-supervised learning, we can address several challenges in healthcare data analysis, including data scarcity, label noise, and the need for domain-specific knowledge. Additionally, this approach can enable the development of more robust and generalizable models that can adapt to new and emerging data sources.

\subsection{Data Privacy and Ethical Considerations}

Advancements in ML and DL are revolutionizing healthcare. They have transformed healthcare delivery, with electronic health records, medical imaging, and digital health services becoming increasingly prevalent. They promise a new era of efficiency, accessibility, and personalized care. However, they request seamless data collection and analysis for research, public health management, and addressing global health challenges which bring complex ethical challenges. 

First, to ensure patient trust and well-being, data privacy and informed consent are paramount. This includes safeguarding the confidentiality and security of patients' personal health information (PHI) \cite{khan2016digital,ruotsalainen2020health}. Data breaches pose a grave threat, potentially causing immense damage to individuals' privacy, dignity, and financial well-being.

Ensuring robust data privacy necessitates a multifaceted approach, encompassing:
\begin{itemize}
    \item Protection from Unauthorized Access and Misuse: PHI must be shielded from unauthorized access and potential misuse by implementing stringent access controls and clear data governance protocols.
    \item Data Security Measures: Rigorous data security measures, such as encryption and intrusion detection systems, are crucial to safeguard PHI from cyberattacks and data breaches.
    \item Confidentiality: Maintaining the confidentiality of PHI is paramount. Healthcare providers and organizations have a fundamental ethical and legal obligation to keep patient information confidential.
    \item Responsible data sharing: Key measures need to be taken into consideration when sharing PHIs such as de-identifying the data, utilizing secure file transfer protocols (e.g., SFTP), conducting regular security audits to identify and addressing potential vulnerabilities.. \end{itemize}

Furthermore, patient autonomy and control over their data are essential. This entails ensuring patients are:
\begin{itemize}
    \item Informed: Patients have the right to be comprehensively informed about how their PHI will be used, who will have access to it, and the associated risks and benefits of utilizing digital health services.
    \item Empowered: Patients should be empowered to make informed decisions regarding the use of their PHI. This includes the right to grant, withhold, or withdraw consent for data collection and usage.
\end{itemize}

Finally, the integration of ethical principles into data management is imperative to preclude legal and ethical complications. Implementing robust data governance frameworks can help healthcare organizations comply with regulations and mitigate risks. To address these challenges and safeguard individual privacy, a significant number of regulatory frameworks have been established worldwide. These include, for instance:
\begin{itemize}
    \item General Data Protection Regulation (GDPR) in Europe \cite{p1}: This regulation outlines stringent data protection requirements for organizations processing personal data within the European Union.
    \item Health Insurance Portability and Accountability Act (HIPAA) in the USA \cite{p2}: HIPAA establishes national standards for protecting the privacy of individually identifiable health information.
    \item Personal Information Protection and Electronic Documents Act (PIPEDA) in Canada \cite{p3}: PIPEDA governs the collection, use, and disclosure of personal information in the course of commercial activities.
\end{itemize}

By adopting a comprehensive and ethically sound approach to data privacy, healthcare providers can ensure patient trust and navigate the evolving digital healthcare landscape responsibly.

\section{Conclusion}
\label{sec9}

The healthcare field has significantly benefited from the evolution of AI. In recent years, graph-based SSL has emerged as a powerful technique, demonstrating exceptional performance in handling complex and unlabeled datasets, especially in medical contexts.
In this comprehensive review, we investigate the state-of-the-art developments in graph learning algorithms in healthcare, particularly graph-based SSL, and explore its transformative applications across various healthcare sectors, including disease prediction, medical imaging, and drug discovery. We highlight potential challenges, limitations, and future research directions. 

It has been demonstrated that this approach is strong in managing complex interactions found in healthcare datasets, highlighting its advantages over conventional supervised learning techniques in different applications. Using graph-based SSL in healthcare settings points to a move toward more patient-specific, efficient, and data-driven methods. Graph-based SSL helps to provide more accurate diagnoses and customized treatment. With less dependence on huge labeled datasets, SSL provides new research prospects for examining disease causes and possible treatments. 

As graph-based SSL continues to evolve, its impact on healthcare is expected to grow and change both the medical research methodologies and clinical operations. While several challenges like data security, models interpretability, and seamlessly incorporating them into current healthcare practices persist, the potential advantages of graph-based SSL highlight its importance for continuous investment and development. Future research should address these challenges, improve graph-based SSL techniques, and validate their effectiveness in real-world healthcare settings.

\section*{Ethics Statement}
This study did not involve experiments with human or animal subjects.
\section*{Acknowledgement}
The authors would like to thank Prince Sultan University for their support.

 \bibliographystyle{elsarticle-num} 
 \bibliography{cas-refs}

\begin{thebibliography}{100}
\expandafter\ifx\csname url\endcsname\relax
  \def\url#1{\texttt{#1}}\fi
\expandafter\ifx\csname urlprefix\endcsname\relax\def\urlprefix{URL }\fi
\expandafter\ifx\csname href\endcsname\relax
  \def\href#1#2{#2} \def\path#1{#1}\fi

\bibitem{aung2021promise}
Y.~Y. Aung, D.~C. Wong, D.~S. Ting, The promise of artificial intelligence: a review of the opportunities and challenges of artificial intelligence in healthcare, British medical bulletin 139~(1) (2021) 4--15.

\bibitem{atitallah2024enhancing}
S.~B. Atitallah, M.~Driss, W.~Boulila, A.~Koubaa, Enhancing early alzheimer's disease detection through big data and ensemble few-shot learning, IEEE Journal of Biomedical and Health Informatics (2024).

\bibitem{calazans2024machine}
M.~A.~A. Calazans, F.~A. Ferreira, F.~A. Santos, F.~Madeiro, J.~B. Lima, Machine learning and graph signal processing applied to healthcare: A review, Bioengineering 11~(7) (2024) 671.

\bibitem{xia2021graph}
F.~Xia, K.~Sun, S.~Yu, A.~Aziz, L.~Wan, S.~Pan, H.~Liu, Graph learning: A survey, IEEE Transactions on Artificial Intelligence 2~(2) (2021) 109--127.

\bibitem{c1}
V.~Rani, M.~Kumar, A.~Gupta, M.~Sachdeva, A.~Mittal, K.~Kumar, Self-supervised learning for medical image analysis: a comprehensive review, Evolving Systems (2024) 1--27.

\bibitem{c2}
Z.~Liu, K.~Kainth, A.~Zhou, T.~W. Deyer, Z.~A. Fayad, H.~Greenspan, X.~Mei, A review of self-supervised, generative, and few-shot deep learning methods for data-limited magnetic resonance imaging segmentation, NMR in Biomedicine (2024) e5143.

\bibitem{c3}
B.~VanBerlo, J.~Hoey, A.~Wong, A survey of the impact of self-supervised pretraining for diagnostic tasks in medical x-ray, ct, mri, and ultrasound, BMC Medical Imaging 24~(1) (2024) 79.

\bibitem{pani2024examining}
K.~Pani, I.~Chawla, Examining the quality of learned representations in self-supervised medical image analysis: a comprehensive review and empirical study, Multimedia Tools and Applications (2024) 1--31.

\bibitem{huang2023self}
S.-C. Huang, A.~Pareek, M.~Jensen, M.~P. Lungren, S.~Yeung, A.~S. Chaudhari, Self-supervised learning for medical image classification: a systematic review and implementation guidelines, NPJ Digital Medicine 6~(1) (2023) 74.

\bibitem{krishnan2022self}
R.~Krishnan, P.~Rajpurkar, E.~J. Topol, Self-supervised learning in medicine and healthcare, Nature Biomedical Engineering 6~(12) (2022) 1346--1352.

\bibitem{shurrab2022self}
S.~Shurrab, R.~Duwairi, Self-supervised learning methods and applications in medical imaging analysis: A survey, PeerJ Computer Science 8 (2022) e1045.

\bibitem{chowdhury2021applying}
A.~Chowdhury, J.~Rosenthal, J.~Waring, R.~Umeton, Applying self-supervised learning to medicine: review of the state of the art and medical implementations, in: Informatics, Vol.~8, MDPI, 2021, p.~59.

\bibitem{lu2023disease}
H.~Lu, S.~Uddin, Disease prediction using graph machine learning based on electronic health data: A review of approaches and trends, in: Healthcare, Vol.~11, MDPI, 2023, p. 1031.

\bibitem{liu2022graph}
Y.~Liu, M.~Jin, S.~Pan, C.~Zhou, Y.~Zheng, F.~Xia, S.~Y. Philip, Graph self-supervised learning: A survey, IEEE transactions on knowledge and data engineering 35~(6) (2022) 5879--5900.

\bibitem{xie2022self}
Y.~Xie, Z.~Xu, J.~Zhang, Z.~Wang, S.~Ji, Self-supervised learning of graph neural networks: A unified review, IEEE transactions on pattern analysis and machine intelligence 45~(2) (2022) 2412--2429.

\bibitem{liu2021self}
X.~Liu, F.~Zhang, Z.~Hou, L.~Mian, Z.~Wang, J.~Zhang, J.~Tang, Self-supervised learning: Generative or contrastive, IEEE transactions on knowledge and data engineering 35~(1) (2021) 857--876.

\bibitem{jaiswal2020survey}
A.~Jaiswal, A.~R. Babu, M.~Z. Zadeh, D.~Banerjee, F.~Makedon, A survey on contrastive self-supervised learning, Technologies 9~(1) (2020) 2.

\bibitem{wang2023review}
W.-C. Wang, E.~Ahn, D.~Feng, J.~Kim, A review of predictive and contrastive self-supervised learning for medical images, Machine Intelligence Research 20~(4) (2023) 483--513.

\bibitem{gori2005new}
M.~Gori, G.~Monfardini, F.~Scarselli, A new model for learning in graph domains, in: Proceedings. 2005 IEEE International Joint Conference on Neural Networks, 2005., Vol.~2, IEEE, 2005, pp. 729--734.

\bibitem{paul2024systematic}
S.~G. Paul, A.~Saha, M.~Z. Hasan, S.~R.~H. Noori, A.~Moustafa, A systematic review of graph neural network in healthcare-based applications: recent advances, trends, and future directions, IEEE Access (2024).

\bibitem{wu2020comprehensive}
Z.~Wu, S.~Pan, F.~Chen, G.~Long, C.~Zhang, S.~Y. Philip, A comprehensive survey on graph neural networks, IEEE transactions on neural networks and learning systems 32~(1) (2020) 4--24.

\bibitem{kipf2016semi}
T.~N. Kipf, M.~Welling, Semi-supervised classification with graph convolutional networks, arXiv preprint arXiv:1609.02907 (2016).

\bibitem{hamilton2017inductive}
W.~Hamilton, Z.~Ying, J.~Leskovec, Inductive representation learning on large graphs, Advances in neural information processing systems 30 (2017).

\bibitem{velickovic2017graph}
P.~Velickovic, G.~Cucurull, A.~Casanova, A.~Romero, P.~Lio, Y.~Bengio, et~al., Graph attention networks, stat 1050~(20) (2017) 10--48550.

\bibitem{kipf2016variational}
T.~N. Kipf, M.~Welling, Variational graph auto-encoders, arXiv preprint arXiv:1611.07308 (2016).

\bibitem{sun2020graph}
M.~Sun, S.~Zhao, C.~Gilvary, O.~Elemento, J.~Zhou, F.~Wang, Graph convolutional networks for computational drug development and discovery, Briefings in bioinformatics 21~(3) (2020) 919--935.

\bibitem{jha2022prediction}
K.~Jha, S.~Saha, H.~Singh, Prediction of protein--protein interaction using graph neural networks, Scientific Reports 12~(1) (2022) 8360.

\bibitem{bian2021gatcda}
C.~Bian, X.-J. Lei, F.-X. Wu, Gatcda: predicting circrna-disease associations based on graph attention network, Cancers 13~(11) (2021) 2595.

\bibitem{ji2021predicting}
C.~Ji, Y.~Wang, J.~Ni, C.~Zheng, Y.~Su, Predicting mirna-disease associations based on heterogeneous graph attention networks, Frontiers in genetics 12 (2021) 727744.

\bibitem{nikolentzos2023synthetic}
G.~Nikolentzos, M.~Vazirgiannis, C.~Xypolopoulos, M.~Lingman, E.~G. Brandt, Synthetic electronic health records generated with variational graph autoencoders, npj Digital Medicine 6~(1) (2023) 83.

\bibitem{lu2021weighted}
H.~Lu, S.~Uddin, A weighted patient network-based framework for predicting chronic diseases using graph neural networks, Scientific reports 11~(1) (2021) 22607.

\bibitem{sun2020disease}
Z.~Sun, H.~Yin, H.~Chen, T.~Chen, L.~Cui, F.~Yang, Disease prediction via graph neural networks, IEEE Journal of Biomedical and Health Informatics 25~(3) (2020) 818--826.

\bibitem{zheng2022multi}
S.~Zheng, Z.~Zhu, Z.~Liu, Z.~Guo, Y.~Liu, Y.~Yang, Y.~Zhao, Multi-modal graph learning for disease prediction, IEEE Transactions on Medical Imaging 41~(9) (2022) 2207--2216.

\bibitem{zheng2024bpi}
K.~Zheng, S.~Yu, L.~Chen, L.~Dang, B.~Chen, Bpi-gnn: Interpretable brain network-based psychiatric diagnosis and subtyping, NeuroImage 292 (2024) 120594.

\bibitem{zheng2024ci}
K.~Zheng, S.~Yu, B.~Chen, Ci-gnn: A granger causality-inspired graph neural network for interpretable brain network-based psychiatric diagnosis, Neural Networks 172 (2024) 106147.

\bibitem{zhang2023graph}
L.~Zhang, Y.~Zhao, T.~Che, S.~Li, X.~Wang, Graph neural networks for image-guided disease diagnosis: A review, iRADIOLOGY 1~(2) (2023) 151--166.

\bibitem{lee2022chexgat}
Y.-W. Lee, S.-K. Huang, R.-F. Chang, Chexgat: A disease correlation-aware network for thorax disease diagnosis from chest x-ray images, Artificial Intelligence in Medicine 132 (2022) 102382.

\bibitem{bagwan2023precise}
F.~Bagwan, N.~Pise, A precise and timely graph-based approach to identify sars covid19 infection from medical imaging data using isocovnet, International Journal of Imaging Systems and Technology 33~(4) (2023) 1160--1176.

\bibitem{song2023covid}
K.~Song, H.~Park, J.~Lee, A.~Kim, J.~Jung, Covid-19 infection inference with graph neural networks, Scientific Reports 13~(1) (2023) 11469.

\bibitem{gaggion2022improving}
N.~Gaggion, L.~Mansilla, C.~Mosquera, D.~H. Milone, E.~Ferrante, Improving anatomical plausibility in medical image segmentation via hybrid graph neural networks: applications to chest x-ray analysis, IEEE Transactions on Medical Imaging 42~(2) (2022) 546--556.

\bibitem{kumar2022sars}
A.~Kumar, A.~R. Tripathi, S.~C. Satapathy, Y.-D. Zhang, Sars-net: Covid-19 detection from chest x-rays by combining graph convolutional network and convolutional neural network, Pattern Recognition 122 (2022) 108255.

\bibitem{gaudelet2021utilizing}
T.~Gaudelet, B.~Day, A.~R. Jamasb, J.~Soman, C.~Regep, G.~Liu, J.~B. Hayter, R.~Vickers, C.~Roberts, J.~Tang, et~al., Utilizing graph machine learning within drug discovery and development, Briefings in bioinformatics 22~(6) (2021) bbab159.

\bibitem{han2021reliable}
K.~Han, B.~Lakshminarayanan, J.~Liu, Reliable graph neural networks for drug discovery under distributional shift, arXiv preprint arXiv:2111.12951 (2021).

\bibitem{cheung2020graph}
M.~Cheung, J.~M. Moura, Graph neural networks for covid-19 drug discovery, in: 2020 IEEE International Conference on Big Data (Big Data), IEEE, 2020, pp. 5646--5648.

\bibitem{li2020learn}
P.~Li, J.~Wang, Y.~Qiao, H.~Chen, Y.~Yu, X.~Yao, P.~Gao, G.~Xie, S.~Song, Learn molecular representations from large-scale unlabeled molecules for drug discovery, arXiv preprint arXiv:2012.11175 (2020).

\bibitem{bongini2021molecular}
P.~Bongini, M.~Bianchini, F.~Scarselli, Molecular generative graph neural networks for drug discovery, Neurocomputing 450 (2021) 242--252.

\bibitem{wang2024accurate}
Y.~Wang, Z.~Yang, Q.~Yao, Accurate and interpretable drug-drug interaction prediction enabled by knowledge subgraph learning, Communications Medicine 4~(1) (2024) 59.

\bibitem{luo2024prediction}
H.~Luo, C.~Zhu, J.~Wang, G.~Zhang, J.~Luo, C.~Yan, Prediction of drug--disease associations based on reinforcement symmetric metric learning and graph convolution network, Frontiers in Pharmacology 15 (2024) 1337764.

\bibitem{rani2023self}
V.~Rani, S.~T. Nabi, M.~Kumar, A.~Mittal, K.~Kumar, Self-supervised learning: A succinct review, Archives of Computational Methods in Engineering 30~(4) (2023) 2761--2775.

\bibitem{zhang2023dive}
C.~Zhang, H.~Zheng, Y.~Gu, Dive into the details of self-supervised learning for medical image analysis, Medical Image Analysis 89 (2023) 102879.

\bibitem{khoshraftar2024survey}
S.~Khoshraftar, A.~An, A survey on graph representation learning methods, ACM Transactions on Intelligent Systems and Technology 15~(1) (2024) 1--55.

\bibitem{wang2022graph}
Y.~Wang, W.~Jin, T.~Derr, Graph neural networks: Self-supervised learning, Graph Neural Networks: Foundations, Frontiers, and Applications (2022) 391--420.

\bibitem{wu2021self}
L.~Wu, H.~Lin, C.~Tan, Z.~Gao, S.~Z. Li, Self-supervised learning on graphs: Contrastive, generative, or predictive, IEEE Transactions on Knowledge and Data Engineering 35~(4) (2021) 4216--4235.

\bibitem{yu2022graph}
S.~Yu, H.~Huang, M.~N. Dao, F.~Xia, Graph augmentation learning, in: Companion Proceedings of the Web Conference 2022, 2022, pp. 1063--1072.

\bibitem{hu2019strategies}
W.~Hu, B.~Liu, J.~Gomes, M.~Zitnik, P.~Liang, V.~Pande, J.~Leskovec, Strategies for pre-training graph neural networks, arXiv preprint arXiv:1905.12265 (2019).

\bibitem{cui2023smg}
Y.~Cui, Z.~Wang, X.~Wang, Y.~Zhang, Y.~Zhang, T.~Pan, Z.~Zhang, S.~Li, Y.~Guo, T.~Akutsu, et~al., Smg: self-supervised masked graph learning for cancer gene identification, Briefings in Bioinformatics 24~(6) (2023) bbad406.

\bibitem{zhu2021graph}
Y.~Zhu, Y.~Xu, F.~Yu, Q.~Liu, S.~Wu, L.~Wang, Graph contrastive learning with adaptive augmentation, in: Proceedings of the Web Conference 2021, 2021, pp. 2069--2080.

\bibitem{ali2024features}
A.~Ali, J.~Li, Features based adaptive augmentation for graph contrastive learning, Digital Signal Processing 145 (2024) 104312.

\bibitem{jiao2020sub}
Y.~Jiao, Y.~Xiong, J.~Zhang, Y.~Zhang, T.~Zhang, Y.~Zhu, Sub-graph contrast for scalable self-supervised graph representation learning, in: 2020 IEEE international conference on data mining (ICDM), IEEE, 2020, pp. 222--231.

\bibitem{subramonian2021motif}
A.~Subramonian, Motif-driven contrastive learning of graph representations, in: Proceedings of the AAAI Conference on Artificial Intelligence, Vol.~35, 2021, pp. 15980--15981.

\bibitem{suresh2021adversarial}
S.~Suresh, P.~Li, C.~Hao, J.~Neville, Adversarial graph augmentation to improve graph contrastive learning, Advances in Neural Information Processing Systems 34 (2021) 15920--15933.

\bibitem{yu2024sparse}
S.~Yu, H.~Wang, M.~Hua, C.~Liang, Y.~Sun, Sparse graph cascade multi-kernel fusion contrastive learning for microbe--disease association prediction, Expert Systems with Applications 252 (2024) 124092.

\bibitem{ding2022data}
K.~Ding, Z.~Xu, H.~Tong, H.~Liu, Data augmentation for deep graph learning: A survey, ACM SIGKDD Explorations Newsletter 24~(2) (2022) 61--77.

\bibitem{jin2024sadr}
S.~Jin, Y.~Zhang, H.~Yu, M.~Lu, Sadr: self-supervised graph learning with adaptive denoising for drug repositioning, IEEE/ACM Transactions on Computational Biology and Bioinformatics (2024).

\bibitem{zeng2021contrastive}
J.~Zeng, P.~Xie, Contrastive self-supervised learning for graph classification, in: Proceedings of the AAAI conference on Artificial Intelligence, Vol.~35, 2021, pp. 10824--10832.

\bibitem{hinton2006reducing}
G.~E. Hinton, R.~R. Salakhutdinov, Reducing the dimensionality of data with neural networks, science 313~(5786) (2006) 504--507.

\bibitem{baldi2012autoencoders}
P.~Baldi, Autoencoders, unsupervised learning, and deep architectures, in: Proceedings of ICML workshop on unsupervised and transfer learning, JMLR Workshop and Conference Proceedings, 2012, pp. 37--49.

\bibitem{liu2024developing}
H.~Liu, X.~Fu, H.~Chen, J.~Shang, H.~Zhou, W.~Zhe, X.~Yao, Developing explainable models for lncrna-targeted drug discovery using graph autoencoders, Future Generation Computer Systems (2024).

\bibitem{liu2023self}
C.~Liu, S.~Wu, R.~Li, D.~Jiang, H.-S. Wong, Self-supervised graph completion for incomplete multi-view clustering, IEEE Transactions on Knowledge and Data Engineering (2023).

\bibitem{you2020does}
Y.~You, T.~Chen, Z.~Wang, Y.~Shen, When does self-supervision help graph convolutional networks?, in: international conference on machine learning, PMLR, 2020, pp. 10871--10880.

\bibitem{kim2024hypeboy}
S.~Kim, S.~Kang, F.~Bu, S.~Y. Lee, J.~Yoo, K.~Shin, Hypeboy: Generative self-supervised representation learning on hypergraphs, arXiv preprint arXiv:2404.00638 (2024).

\bibitem{zang2023hierarchical}
X.~Zang, X.~Zhao, B.~Tang, Hierarchical molecular graph self-supervised learning for property prediction, Communications Chemistry 6~(1) (2023) 34.

\bibitem{xu2021predictive}
X.~Xu, X.~Xu, Y.~Sun, X.~Liu, X.~Li, G.~Xie, F.~Wang, Predictive modeling of clinical events with mutual enhancement between longitudinal patient records and medical knowledge graph, in: 2021 IEEE International Conference on Data Mining (ICDM), IEEE, 2021, pp. 777--786.

\bibitem{tang2021self}
S.~Tang, J.~A. Dunnmon, K.~Saab, X.~Zhang, Q.~Huang, F.~Dubost, D.~L. Rubin, C.~Lee-Messer, Self-supervised graph neural networks for improved electroencephalographic seizure analysis, arXiv preprint arXiv:2104.08336 (2021).

\bibitem{zhao2024survey}
Z.~Zhao, Y.~Li, Y.~Zou, R.~Li, R.~Zhang, A survey on self-supervised pre-training of graph foundation models: A knowledge-based perspective, arXiv preprint arXiv:2403.16137 (2024).

\bibitem{akkas2022jgcl}
S.~Akkas, A.~Azad, Jgcl: Joint self-supervised and supervised graph contrastive learning, in: Companion Proceedings of the Web Conference 2022, 2022, pp. 1099--1105.

\bibitem{zhu2023unsupervised}
Y.~Zhu, Y.~Xu, F.~Yu, Q.~Liu, S.~Wu, Unsupervised graph representation learning with cluster-aware self-training and refining, ACM Transactions on Intelligent Systems and Technology 14~(5) (2023) 1--21.

\bibitem{yao2022self}
H.-R. Yao, N.~Cao, K.~Russell, D.-C. Chang, O.~Frieder, J.~T. Fineman, Self-supervised representation learning on electronic health records with graph kernel infomax, ACM Transactions on Computing for Healthcare (2022).

\bibitem{ho2023self}
T.~K.~K. Ho, N.~Armanfard, Self-supervised learning for anomalous channel detection in eeg graphs: Application to seizure analysis, in: Proceedings of the AAAI conference on artificial intelligence, Vol.~37, 2023, pp. 7866--7874.

\bibitem{lu2021self}
C.~Lu, C.~K. Reddy, Y.~Ning, Self-supervised graph learning with hyperbolic embedding for temporal health event prediction, IEEE Transactions on Cybernetics 53~(4) (2021) 2124--2136.

\bibitem{xu2023seqcare}
Y.~Xu, X.~Chu, K.~Yang, Z.~Wang, P.~Zou, H.~Ding, J.~Zhao, Y.~Wang, B.~Xie, Seqcare: Sequential training with external medical knowledge graph for diagnosis prediction in healthcare data, in: Proceedings of the ACM Web Conference 2023, 2023, pp. 2819--2830.

\bibitem{ruan2023msgcl}
X.~Ruan, C.~Jiang, P.~Lin, Y.~Lin, J.~Liu, S.~Huang, X.~Liu, Msgcl: inferring mirna--disease associations based on multi-view self-supervised graph structure contrastive learning, Briefings in Bioinformatics 24~(2) (2023) bbac623.

\bibitem{yao2024self}
H.-R. Yao, N.~Cao, K.~Russell, D.-C. Chang, O.~Frieder, J.~T. Fineman, Self-supervised representation learning on electronic health records with graph kernel infomax, ACM Transactions on Computing for Healthcare 5~(2) (2024) 1--28.

\bibitem{xie2024predicting}
J.~Xie, J.~Rao, J.~Xie, H.~Zhao, Y.~Yang, Predicting disease-gene associations through self-supervised mutual infomax graph convolution network, Computers in Biology and Medicine 170 (2024) 108048.

\bibitem{wen2023graph}
G.~Wen, P.~Cao, L.~Liu, J.~Yang, X.~Zhang, F.~Wang, O.~R. Zaiane, Graph self-supervised learning with application to brain networks analysis, IEEE Journal of Biomedical and Health Informatics (2023).

\bibitem{sehanobish2021gaining}
A.~Sehanobish, N.~Ravindra, D.~van Dijk, Gaining insight into sars-cov-2 infection and covid-19 severity using self-supervised edge features and graph neural networks, in: Proceedings of the AAAI Conference on Artificial Intelligence, Vol.~35, 2021, pp. 4864--4873.

\bibitem{lu2024soft}
H.~Lu, T.~Jin, H.~Wei, M.~Nappi, H.~Li, S.~Wan, Soft-orthogonal constrained dual-stream encoder with self-supervised clustering network for brain functional connectivity data, Expert Systems with Applications 244 (2024) 122898.

\bibitem{jung2024cancergate}
S.~Jung, S.~Wang, D.~Lee, Cancergate: Prediction of cancer-driver genes using graph attention autoencoders, Computers in Biology and Medicine (2024) 108568.

\bibitem{peng2022gate}
L.~Peng, N.~Wang, J.~Xu, X.~Zhu, X.~Li, Gate: Graph cca for temporal self-supervised learning for label-efficient fmri analysis, IEEE Transactions on Medical Imaging 42~(2) (2022) 391--402.

\bibitem{wang2022contrastive}
X.~Wang, L.~Yao, I.~Rekik, Y.~Zhang, Contrastive functional connectivity graph learning for population-based fmri classification, in: International Conference on Medical Image Computing and Computer-Assisted Intervention, Springer, 2022, pp. 221--230.

\bibitem{choi2024joint}
J.~Choi, H.~Lee, B.-H. Kim, J.~Lee, Joint-embedding masked autoencoder for self-supervised learning of dynamic functional connectivity from the human brain, arXiv preprint arXiv:2403.06432 (2024).

\bibitem{ibrahim2022multi}
M.~Ibrahim, S.~Henna, G.~Cullen, Multi-graph convolutional neural network for breast cancer multi-task classification, in: Irish Conference on Artificial Intelligence and Cognitive Science, Springer, 2022, pp. 40--54.

\bibitem{sun2021context}
L.~Sun, K.~Yu, K.~Batmanghelich, Context matters: Graph-based self-supervised representation learning for medical images, in: Proceedings of the AAAI Conference on Artificial Intelligence, Vol.~35, 2021, pp. 4874--4882.

\bibitem{ozen2021self}
Y.~Ozen, S.~Aksoy, K.~K{\"o}semehmeto{\u{g}}lu, S.~{\"O}nder, A.~{\"U}ner, Self-supervised learning with graph neural networks for region of interest retrieval in histopathology, in: 2020 25th International conference on pattern recognition (ICPR), IEEE, 2021, pp. 6329--6334.

\bibitem{lin2021multi}
J.~Lin, Q.~Cai, M.~Lin, Multi-label classification of fundus images with graph convolutional network and self-supervised learning, IEEE Signal Processing Letters 28 (2021) 454--458.

\bibitem{guo2022self}
R.~Guo, J.~Sun, C.~Zhang, X.~Qian, A self-supervised metric learning framework for the arising-from-chair assessment of parkinsonians with graph convolutional networks, IEEE Transactions on Circuits and Systems for Video Technology 32~(9) (2022) 6461--6471.

\bibitem{endo2022gaitforemer}
M.~Endo, K.~L. Poston, E.~V. Sullivan, L.~Fei-Fei, K.~M. Pohl, E.~Adeli, Gaitforemer: Self-supervised pre-training of transformers via human motion forecasting for few-shot gait impairment severity estimation, in: International Conference on Medical Image Computing and Computer-Assisted Intervention, Springer, 2022, pp. 130--139.

\bibitem{guo2022tree}
R.~Guo, H.~Li, C.~Zhang, X.~Qian, A tree-structure-guided graph convolutional network with contrastive learning for the assessment of parkinsonian hand movements, Medical Image Analysis 81 (2022) 102560.

\bibitem{guo2022contrastive}
R.~Guo, J.~Sun, C.~Zhang, X.~Qian, A contrastive graph convolutional network for toe-tapping assessment in parkinson’s disease, IEEE Transactions on Circuits and Systems for Video Technology 32~(12) (2022) 8864--8874.

\bibitem{mh2024lvm}
D.~MH~Nguyen, H.~Nguyen, N.~Diep, T.~N. Pham, T.~Cao, B.~Nguyen, P.~Swoboda, N.~Ho, S.~Albarqouni, P.~Xie, et~al., Lvm-med: Learning large-scale self-supervised vision models for medical imaging via second-order graph matching, Advances in Neural Information Processing Systems 36 (2024).

\bibitem{wang2021hierarchical}
Z.~Wang, J.~Li, Z.~Pan, W.~Li, A.~Sisk, H.~Ye, W.~Speier, C.~W. Arnold, Hierarchical graph pathomic network for progression free survival prediction, in: Medical Image Computing and Computer Assisted Intervention--MICCAI 2021: 24th International Conference, Strasbourg, France, September 27--October 1, 2021, Proceedings, Part VIII 24, Springer, 2021, pp. 227--237.

\bibitem{nguyen2023companion}
T.~S. Nguyen, S.~Lee, J.~Lee, L.~V. Nguyen, O.-J. Lee, et~al., Companion animal disease diagnostics based on literal-aware medical knowledge graph representation learning, IEEE Access (2023).

\bibitem{aryal2024context}
M.~Aryal, N.~Y. Soltani, Context-aware self-supervised learning of whole slide images, IEEE Transactions on Artificial Intelligence (2024).

\bibitem{zhang2023antiviraldl}
P.~Zhang, X.~Hu, G.~Li, L.~Deng, Antiviraldl: Computational antiviral drug repurposing using graph neural network and self-supervised learning, IEEE Journal of Biomedical and Health Informatics (2023).

\bibitem{rong2020self}
Y.~Rong, Y.~Bian, T.~Xu, W.~Xie, Y.~Wei, W.~Huang, J.~Huang, Self-supervised graph transformer on large-scale molecular data, Advances in neural information processing systems 33 (2020) 12559--12571.

\bibitem{zhao2021csgnn}
C.~Zhao, S.~Liu, F.~Huang, S.~Liu, W.~Zhang, Csgnn: Contrastive self-supervised graph neural network for molecular interaction prediction., in: IJCAI, 2021, pp. 3756--3763.

\bibitem{wang2024evaluating}
H.~Wang, J.~Kaddour, S.~Liu, J.~Tang, J.~Lasenby, Q.~Liu, Evaluating self-supervised learning for molecular graph embeddings, Advances in Neural Information Processing Systems 36 (2024).

\bibitem{li2021effective}
P.~Li, J.~Wang, Y.~Qiao, H.~Chen, Y.~Yu, X.~Yao, P.~Gao, G.~Xie, S.~Song, An effective self-supervised framework for learning expressive molecular global representations to drug discovery, Briefings in Bioinformatics 22~(6) (2021) bbab109.

\bibitem{wang2024hierarchical}
Y.~Wang, J.~Song, Q.~Dai, X.~Duan, Hierarchical negative sampling based graph contrastive learning approach for drug-disease association prediction, IEEE Journal of Biomedical and Health Informatics (2024).

\bibitem{d3}
A.~E. Johnson, T.~J. Pollard, L.~Shen, L.-w.~H. Lehman, M.~Feng, M.~Ghassemi, B.~Moody, P.~Szolovits, L.~Anthony~Celi, R.~G. Mark, Mimic-iii, a freely accessible critical care database, Scientific data 3~(1) (2016) 1--9.

\bibitem{d1}
A.~E. Johnson, L.~Bulgarelli, L.~Shen, A.~Gayles, A.~Shammout, S.~Horng, T.~J. Pollard, S.~Hao, B.~Moody, B.~Gow, et~al., Mimic-iv, a freely accessible electronic health record dataset, Scientific data 10~(1) (2023) 1.

\bibitem{d2}
J.~Liu, T.~Lichtenberg, K.~A. Hoadley, L.~M. Poisson, A.~J. Lazar, A.~D. Cherniack, A.~J. Kovatich, C.~C. Benz, D.~A. Levine, A.~V. Lee, et~al., An integrated tcga pan-cancer clinical data resource to drive high-quality survival outcome analytics, Cell 173~(2) (2018) 400--416.

\bibitem{d4}
C.-G. Yan, X.~Chen, L.~Li, F.~X. Castellanos, T.-J. Bai, Q.-J. Bo, J.~Cao, G.-M. Chen, N.-X. Chen, W.~Chen, et~al., Reduced default mode network functional connectivity in patients with recurrent major depressive disorder, Proceedings of the National Academy of Sciences 116~(18) (2019) 9078--9083.

\bibitem{d5}
R.~S. Lee, F.~Gimenez, A.~Hoogi, K.~K. Miyake, M.~Gorovoy, D.~L. Rubin, A curated mammography data set for use in computer-aided detection and diagnosis research, Scientific data 4~(1) (2017) 1--9.

\bibitem{d6}
A.~Ianevski, R.~M. Simonsen, V.~Myhre, T.~Tenson, V.~Oksenych, M.~Bj{\o}r{\aa}s, D.~E. Kainov, Drugvirus. info 2.0: an integrative data portal for broad-spectrum antivirals (bsa) and bsa-containing drug combinations (bccs), Nucleic acids research 50~(W1) (2022) W272--W275.

\bibitem{d7}
C.~Morris, N.~M. Kriege, F.~Bause, K.~Kersting, P.~Mutzel, M.~Neumann, Tudataset: A collection of benchmark datasets for learning with graphs, arXiv preprint arXiv:2007.08663 (2020).

\bibitem{d8}
P.~Bellec, C.~Chu, F.~Chouinard-Decorte, Y.~Benhajali, D.~S. Margulies, R.~C. Craddock, The neuro bureau adhd-200 preprocessed repository, Neuroimage 144 (2017) 275--286.

\bibitem{d9}
W.~Ma, L.~Zhang, P.~Zeng, C.~Huang, J.~Li, B.~Geng, J.~Yang, W.~Kong, X.~Zhou, Q.~Cui, An analysis of human microbe--disease associations, Briefings in bioinformatics 18~(1) (2017) 85--97.

\bibitem{d10}
Y.~Janssens, J.~Nielandt, A.~Bronselaer, N.~Debunne, F.~Verbeke, E.~Wynendaele, F.~Van~Immerseel, Y.-P. Vandewynckel, G.~De~Tr{\'e}, B.~De~Spiegeleer, Disbiome database: linking the microbiome to disease, BMC microbiology 18 (2018) 1--6.

\bibitem{d11}
X.~Zhao, D.~Rangaprakash, T.~S. Denney, J.~S. Katz, M.~N. Dretsch, G.~Deshpande, Identifying neuropsychiatric disorders using unsupervised clustering methods: data and code, Data in brief 22 (2019) 570--573.

\bibitem{d12}
P.~Lanka, D.~Rangaprakash, M.~N. Dretsch, J.~S. Katz, T.~S. Denney, G.~Deshpande, Supervised machine learning for diagnostic classification from large-scale neuroimaging datasets, Brain imaging and behavior 14 (2020) 2378--2416.

\bibitem{d13}
H.~Luo, J.~Wang, M.~Li, J.~Luo, X.~Peng, F.-X. Wu, Y.~Pan, Drug repositioning based on comprehensive similarity measures and bi-random walk algorithm, Bioinformatics 32~(17) (2016) 2664--2671.

\bibitem{d14}
S.~Axelrod, R.~Gomez-Bombarelli, Geom, energy-annotated molecular conformations for property prediction and molecular generation, Scientific Data 9~(1) (2022) 185.

\bibitem{d15}
D.~S. Wishart, C.~Knox, A.~C. Guo, D.~Cheng, S.~Shrivastava, D.~Tzur, B.~Gautam, M.~Hassanali, Drugbank: a knowledgebase for drugs, drug actions and drug targets, Nucleic acids research 36~(suppl\_1) (2008) D901--D906.

\bibitem{d16}
A.~Hamosh, A.~F. Scott, J.~S. Amberger, C.~A. Bocchini, V.~A. McKusick, Online mendelian inheritance in man (omim), a knowledgebase of human genes and genetic disorders, Nucleic acids research 33~(suppl\_1) (2005) D514--D517.

\bibitem{d17}
Z.~Huang, J.~Shi, Y.~Gao, C.~Cui, S.~Zhang, J.~Li, Y.~Zhou, Q.~Cui, Hmdd v3. 0: a database for experimentally supported human microrna--disease associations, Nucleic acids research 47~(D1) (2019) D1013--D1017.

\bibitem{di1}
J.~Li, W.~Li, A.~Sisk, H.~Ye, W.~D. Wallace, W.~Speier, C.~W. Arnold, A multi-resolution model for histopathology image classification and localization with multiple instance learning, Computers in biology and medicine 131 (2021) 104253.

\bibitem{di2}
N.~Ing, Z.~Ma, J.~Li, H.~Salemi, C.~Arnold, B.~S. Knudsen, A.~Gertych, Semantic segmentation for prostate cancer grading by convolutional neural networks, in: Medical Imaging 2018: Digital Pathology, Vol. 10581, SPIE, 2018, pp. 343--355.

\bibitem{dice}
A.~P. Zijdenbos, B.~M. Dawant, R.~A. Margolin, A.~C. Palmer, Morphometric analysis of white matter lesions in mr images: method and validation, IEEE transactions on medical imaging 13~(4) (1994) 716--724.

\bibitem{van2022three}
G.~M. Van~de Ven, T.~Tuytelaars, A.~S. Tolias, Three types of incremental learning, Nature Machine Intelligence 4~(12) (2022) 1185--1197.

\bibitem{zhu2024review}
Y.~Zhu, S.~Zhao, Y.~Zhang, C.~Zhang, J.~Wu, A review of statistical-based fault detection and diagnosis with probabilistic models, Symmetry 16~(4) (2024) 455.

\bibitem{thomas2021systematic}
T.~Thomas, E.~Rajabi, A systematic review of machine learning-based missing value imputation techniques, Data Technologies and Applications 55~(4) (2021) 558--585.

\bibitem{zhao2022graph}
T.~Zhao, W.~Jin, Y.~Liu, Y.~Wang, G.~Liu, S.~G{\"u}nnemann, N.~Shah, M.~Jiang, Graph data augmentation for graph machine learning: A survey, arXiv preprint arXiv:2202.08871 (2022).

\bibitem{ficek2021differential}
J.~Ficek, W.~Wang, H.~Chen, G.~Dagne, E.~Daley, Differential privacy in health research: A scoping review, Journal of the American Medical Informatics Association 28~(10) (2021) 2269--2276.

\bibitem{atitallah2023fedmicro}
S.~B. Atitallah, M.~Driss, H.~B. Ghezala, Fedmicro-ida: A federated learning and microservices-based framework for iot data analytics, Internet of Things 23 (2023) 100845.

\bibitem{stiglic2020interpretability}
G.~Stiglic, P.~Kocbek, N.~Fijacko, M.~Zitnik, K.~Verbert, L.~Cilar, Interpretability of machine learning-based prediction models in healthcare, Wiley Interdisciplinary Reviews: Data Mining and Knowledge Discovery 10~(5) (2020) e1379.

\bibitem{cg1}
M.~Cascella, J.~Montomoli, V.~Bellini, E.~Bignami, Evaluating the feasibility of chatgpt in healthcare: an analysis of multiple clinical and research scenarios, Journal of medical systems 47~(1) (2023) 33.

\bibitem{cg2}
D.~Restrepo, C.~Wu, C.~V{\'a}squez-Venegas, J.~Matos, J.~Gallifant, L.~A. Celi, D.~S. Bitterman, L.~F. Nakayama, Analyzing diversity in healthcare llm research: A scientometric perspective, arXiv preprint arXiv:2406.13152 (2024).

\bibitem{cg3}
Z.~A. Nazi, W.~Peng, Large language models in healthcare and medical domain: A review, in: Informatics, Vol.~11, MDPI, 2024, p.~57.

\bibitem{khan2016digital}
S.~Khan, A.~Hoque, et~al., Digital health data: a comprehensive review of privacy and security risks and some recommendations, Computer Science Journal of Moldova 71~(2) (2016) 273--292.

\bibitem{ruotsalainen2020health}
P.~Ruotsalainen, B.~Blobel, Health information systems in the digital health ecosystem—problems and solutions for ethics, trust and privacy, International journal of environmental research and public health 17~(9) (2020) 3006.

\bibitem{p1}
M.~Budryt{\.e}, General data protection regulation (gdpr) in european union: From proposal to implementation, Ph.D. thesis (2021).

\bibitem{p2}
P.~F. Edemekong, P.~Annamaraju, M.~J. Haydel, Health insurance portability and accountability act (2018).

\bibitem{p3}
D.~Jaar, P.~E. Zeller, Canadian privacy law: The personal information protection and electronic documents act (pipeda), Int'l. In-House Counsel J. 2 (2008) 1135.

\end{thebibliography}

\end{document}